\documentclass[journal]{IEEEtran}
\usepackage{algorithm}
\usepackage{multirow}
\usepackage{array}
\usepackage{algpseudocode}
\usepackage{amsmath}
\usepackage{cite}
\usepackage{booktabs}
\usepackage{graphicx}
\usepackage{float}
\usepackage{threeparttable}
\usepackage{epstopdf}
\usepackage{xfrac}
\usepackage{makecell}
\usepackage{amssymb}
\usepackage{subfigure}
\usepackage{bm}
\usepackage{cases}
\usepackage{url}
\usepackage{xcolor}
\usepackage{makecell}
\usepackage[breaklinks, colorlinks, 
citecolor=blue,
linkcolor=red,
urlcolor=blue]{hyperref}
\usepackage{threeparttable}

\newcommand{\Rmnum}[1]{\expandafter\@slowromancap\romannumeral #1@}
\makeatother

\begin{document}
	
	\title{A Novel Immune Algorithm for Multiparty Multiobjective Optimization}
	
	\author{Kesheng~Chen, Wenjian~Luo,~\IEEEmembership{Senior Member,~IEEE}, Qi~Zhou, Yujiang~liu, Peilan~Xu,~\IEEEmembership{Member,~IEEE} Yuhui~Shi,~\IEEEmembership{Fellow,~IEEE}%
		\thanks{This study is supported by the National Natural Science Foundation of China (Grant No. U23B2058), Shenzhen Fundamental Research Program (Grant No. JCYJ20220818102414030), the Major Key Project of PCL (Grant No. PCL2022A03), Shenzhen Science and Technology Program (Grant No. ZDSYS20210623091809029), Guangdong Provincial Key Laboratory of Novel Security Intelligence Technologies (Grant No. 2022B1212010005). \textit{(Corresponding author: Wenjian Luo.)}}
		\thanks{Kesheng~Chen, Wenjian~Luo, Qi~Zhou and Yujiang~Liu are with Guangdong Provincial Key Laboratory of Novel Security Intelligence Technologies, School of Computer Science and Technology, Harbin Institute of Technology, Shenzhen 518055, China. Wenjian Luo is also with Peng Cheng Laboratory, Shenzhen 518000, China. (e-mail: 22s151138@stu.hit.edu.cn, luowenjian@hit.edu.cn, 22s051036@stu.hit.edu.cn, 23s151125@stu.hit.edu.cn). Peilan Xu is with School of Artificial Intelligence, Nanjing University of Information Science and Technology, Nanjing 210044, China. (e-mail:xpl@nuist.edu.cn). Yuhui Shi is with School of Computer Science and Engineering, Southern University of Science and Technology, Shenzhen 518055, China. (e-mail: shiyh@sustech.edu.cn)}%
		\thanks{The code for this study can be accessed at \url{https://github.com/MiLab-HITSZ/2023ChenMPIA}.}}
	
	\IEEEpeerreviewmaketitle
	\maketitle
	\begin{abstract}
		Traditional multiobjective optimization problems (MOPs) are insufficiently equipped for scenarios involving multiple decision makers (DMs), which are prevalent in many practical applications. These scenarios are categorized as multiparty multiobjective optimization problems (MPMOPs). For MPMOPs, the goal is to find a solution set that is as close to the Pareto front of each DM as much as possible. This poses challenges for evolutionary algorithms in terms of searching and selecting. To better solve MPMOPs, this paper proposes a novel approach called the multiparty immune algorithm (MPIA). The MPIA incorporates an inter-party guided crossover strategy based on the individual's non-dominated sorting ranks from different DM perspectives and an adaptive activation strategy based on the proposed multiparty cover metric (MCM). These strategies enable MPIA to activate suitable individuals for the next operations, maintain population diversity from different DM perspectives, and enhance the algorithm's search capability. To evaluate the performance of MPIA, we compare it with ordinary multiobjective evolutionary algorithms (MOEAs) and state-of-the-art multiparty multiobjective optimization evolutionary algorithms (MPMOEAs) by solving synthetic multiparty multiobjective problems and real-world biparty multiobjective unmanned aerial vehicle path planning (BPUAV-PP) problems involving multiple DMs. Experimental results demonstrate that MPIA outperforms other algorithms.
	\end{abstract}
	
	\maketitle
	\IEEEpeerreviewmaketitle
	
	\section{Introduction}
	Multiparty multiobjective optimization problems (MPMOPs) provide a more effective approach to describing multiobjective optimization problems (MOPs) involving multiple decision makers (DMs) and each DM is concerned about different conflicting objectives \cite{liuEvolutionaryApproachMultiparty2020,10.1007/978-3-030-78811-7_6,song2022multiobjective,chen2023MPUAV,chang2022multiparty,she2022evolutionary,chang2023biparty}. In MPMOPs, the goal is to find a solution set as close as possible to the true Pareto front (PF) constructed by the objectives of each DM.

	One practical application of MPMOPs is the unmanned aerial vehicle (UAV) path planning problems \cite{chen2023MPUAV}. UAV path planning problems in urban environments often require considerations of multiple objectives and involve multiple departments, such as efficiency-related departments and safety-related government regulation departments \cite{pangUAVPathOptimization2022a,chen2023MPUAV}. The efficiency-related DM focuses on minimizing UAV flying time, reducing energy consumption, and achieving high-quality mission completion (e.g., capturing clear remote sensing photos and collecting extensive sensor data) and other efficiency-related considerations. On the other hand, the safety-related DM aims to minimize external impacts caused by UAVs, such as minimizing risks to people and vehicles in urban areas, minimizing property damage, reducing noise emissions from UAVs, and other safety-related concerns.

	In such an MPMOP, we need to find a Pareto set of path solutions that simultaneously consider the objectives of both efficiency-related DM and safety-related government DM. Therefore, when solving MPMOPs, the algorithm is usually required to effectively balance the convergence and diversity of the population from multiple DM perspectives. Due to the inability of ordinary MOEAs to effectively identifying the more valuable solutions and the strategy of maintaining population diversity across different DMs, ordinary multiobjective evolutionary algorithms (MOEAs) cannot be directly and effectively used to solve MPMOPs.

	To efficiently solve MPMOPs, such as biparty multiobjective UAV path planning problems, some multiparty multiobjective evolutionary algorithms (MPMOEAs)\cite{liuEvolutionaryApproachMultiparty2020,10.1007/978-3-030-78811-7_6,song2022multiobjective,chen2023MPUAV,chang2022multiparty,she2022evolutionary,chang2023biparty} have been proposed. Some algorithms utilize the non-dominated sorting method from multiple decision maker perspectives to improve the convergence ability of the algorithm toward different decision makers (DMs). For example, the first MPMOEA, OptMPNDS\cite{liuEvolutionaryApproachMultiparty2020}, employs the multiparty non-dominated sorting (MPNDS) operator, which uses the ordinary Pareto sorting ranks of different DMs to calculate the multiparty ranks. Additionally, the MPNDS2 operator, which uses two round ordinary Pareto sort and its corresponding OptMPNDS2 algorithm\cite{10.1007/978-3-030-78811-7_6} have been proposed. Following these work, some researchers have designed algorithms with stronger search capabilities using advanced search operators or other evolutionary algorithm frameworks, such as OptMPNDS3\cite{she2022evolutionary}, MOEA-D/MP\cite{chang2023biparty}, BPNNIA\cite{chen2023MPUAV}, BPHEIA\cite{chen2023MPUAV}, BPAIMA\cite{chen2023MPUAV}.

	However, existing MPMOEAs still face certain limitations when addressing MPMOPs. These limitations can be summarized into two main issues: search direction and population diversity in the presence of multiple DMs.

	The first issue concerns the search direction about convergence. MPMOPs require that the final solution set be as close as possible to the true Pareto front constructed by the objectives of each DM, which is more stringent than traditional MOPs. Furthermore, at least two or more decision makers, each typically having two or more objectives, results in a larger objective space, which reduces the efficiency of traditional search methods in finding individuals that meet the requirements. This issue poses a challenge in determining the appropriate search direction. Even though existing MPMOEAs\cite{liuEvolutionaryApproachMultiparty2020,10.1007/978-3-030-78811-7_6,song2022multiobjective,chen2023MPUAV,chang2022multiparty,she2022evolutionary,chang2023biparty} have been improved over traditional MOEAs, their performance gains mostly come from specialized sorting techniques for MPMOPs or advanced search methods previously used in MOEAs. These works still do not exploit certain inherent properties in MPMOPs, such as convergence information and diversity information of the population from different DM perspectives. These works only consider relevant indicators in the function space consisting of all objectives.
	
	To leverage this characteristic, this paper proposes an adaptive operator that combines a novel operator called inter-party guided crossover with multiple ordinary operators. Inspired by the guide mechanisms\cite{guo2023knee,yu2022correlation,ardeh2021genetic} or transfer learning mechanisms \cite{zuo2023process,chen2023evolutionary,guo2022knowledge,li2022reducing,jiang2020individual} in dynamic multiobjective optimization problems (DMOPs), we believe the presence of multiple decision makers allows for the effective application of such an inter-party guided strategy to guide MPMOEAs in their search processes. During the solving process, the population's non-dominated sorting ranks are assigned based on the perspectives of different decision makers. This information about ranks could be effectively utilized to guide search direction. This strategy enables the algorithm to intelligently incorporate information from different decision makers when selecting crossover individuals, thus enhancing its searchability.

	
	The second issue relates to population diversity for different DMs. In MPMOPs, some individuals are at the PF or objective space boundary of a single DM, which may play important roles in maintaining population diversity for some DMs and revealing search direction. Current MPMOEAs, whether using selection operators (in genetic-based algorithms) or activation operators (in immune-based algorithms), do not adequately consider these potentially valuable individuals. Experimental results of this paper show that the ordinary selection operators or activation operators used in existing MPMOEAs have limitations in solving MPMOPs. For example, the fixed-size activation strategy used in MPMOEAs such as BPNNIA\cite{chen2023MPUAV}, BPHEIA \cite{chen2023MPUAV}, and BPAIMA\cite{chen2023MPUAV}, when combined with existing multiparty non-dominated ranking methods, often results in a lack of diversity for some DMs, which could constrain the algorithm's performance.

	In order to address this issue, we utilize the multiparty cover metric (MCM) to measure the diversity of the solution set across multiple DM’s perspectives. Furthermore, we have designed an adaptive activation strategy that adjusts the activation size based on MCM. This strategy aims to improve the overall performance of the immune algorithm in solving MPMOPs. By considering the multiparty cover ratio, our method enhances the diversity of the solution set and effectively improves the performance of the MPMOEA in tackling complex multiparty multiobjective optimization problems.

	Building upon these strategies, we present a new multiparty multiobjective evolutionary algorithm called the multiparty immune algorithm (MPIA) specifically designed for MPMOPs. To assess the effectiveness of MPIA, comprehensive experiments are conducted on both multiparty multiobjective benchmarks \cite{liuEvolutionaryApproachMultiparty2020} and biparty multiobjective UAV path planning problems\cite{chen2023MPUAV}.

	Therefore, the main work of this paper is as follows: 
	\begin{enumerate}
		\item In order to address the challenge of efficiently searching for solutions that meet the requirements of MPMOPs, this paper fully exploits the inherent properties of MPMOPs and proposes the inter-party guided strategy that effectively utilizes population information in the objective spaces from different decision makers (DMs). This strategy allows for the intelligent selection of crossover individuals, ultimately enhancing the algorithm's effectiveness for MPMOPs.
		
		\item In order to address the issue of loss of population diversity for some DMs in solving MPMOPs, 
		this paper proposes an adaptive activation strategy based on the multiparty cover metric (MCM). This strategy guarantees that the activation set has good diversity for all DMs, thus enhancing the algorithm's performance in solving MPMOPs.
		
		\item Building upon the above strategies, a novel multiparty multiobjective immune algorithm (MPIA) is introduced. Its performance is compared with ordinary multiobjective optimization algorithms and state-of-the-art multiparty multiobjective optimization algorithms, including OptMPNDS, OptMPNDS2, BPNNIA, BPHEIA, and BPAIMA. Experiments are conducted and analyzed on synthetic multiparty multiobjective problems and real-world biparty multiobjective UAV path planning problems.
	\end{enumerate}

	The structure of this paper is as follows: Section \ref{sec:backgrounds} provides an overview of MPMOPs and biparty multiobjective UAV path planning problems. This is followed by Section \ref{sec:mpia}, which presents a novel multiparty immune algorithm designed for MPMOPs. Section \ref{sc:set} outlines the metrics used and the parameter settings of the algorithms. Next, Section \ref{sc:expmpmop} details the experiments conducted on synthetic multiparty multiobjective problems, and Section \ref{sc:expuav} expands on the experiments involving biparty multiobjective UAV path planning problems. Finally, we conclude and summarize the paper in Section \ref{sec:con}.
	\section{Backgrounds and Related Work} 
	\label{sec:backgrounds}
	This section briefly reviews the multiparty multiobjective optimization problems and the biparty multiobjective UAV path planning problem.
	\subsection{Multiparty Multiobjective Optimization Problems} 
	A multiobjective optimization problem (MOP)\cite{debFastElitistMultiobjective2002a} is considered a problem with multiple conflicting objectives. Without loss of generality, a minimized MOP can be defined as follows.
	\begin{equation}
		\begin{aligned}
			& \min F(\mathbf{x})=(f_{1}(\bold{x}),f_2(\bold{x}),\dots,f_m(\bold{x})),\\
			&	s.t. \begin{cases}
				g_i(\bold{x}) \leq 0, &  i = 1,\dots,n_g \\
				h_j(\bold{x}) = 0   , &  j = 1,\dots,n_h \\
				\bold{x} \in [x_{min},x_{max}]^d,& 
			\end{cases} \\
		\end{aligned}
	\end{equation}
	where $g_i(\bold{x})$ denotes the $i$-th inequality constraint on $\bold{x}$, $h_j(\bold{x})$ denotes the $j$-th equality constraint on $\bold{x}$, $n_g$ and $n_h$ correspond to the numbers of inequality constraints and equality constraints, respectively. $\bold{x}=(x_1,x_2,\dots,x_d)$ is an $d$-dimensional decision variable. The parameter $m$ denotes the number of objectives, $f_i(\bold{x})$ is the $i$-th objective function.

	For two decision vectors $\bold{x}$ and $\bold{y}$ satisfy the following two conditions, it can be said that $\bold{x}$ Pareto dominates $\bold{y}$, which is denoted as $\bold{x} \prec \bold{y} $.
	\begin{enumerate}
		\item none of the objectives of $\bold{x}$ are greater than the objectives of $\bold{y}$;
		\item there exists at least one objective of $\bold{x}$ that is less than the corresponding objective of $\bold{y}$.
	\end{enumerate}

	Based on the definition of Pareto dominance, the solution $\bold{x}^P$ is Pareto optimal if and only if no other solutions Pareto dominate $\bold{x}^P$. In addition, all Pareto optimal solutions compose the Pareto optimal set (PS), and the corresponding objective vectors set is called the Pareto front (PF).

	In MOPs, there is only one DM and the DM is concerned with all the objectives. However, different objectives may be considered by different departments in many real-world applications. For example, UAV urban path planning\cite{chen2023MPUAV}, power flow optimization\cite{she2022evolutionary}, and airport slot scheduling may involve multiple departments or companies. MPMOPs can better describe such problems\cite{liuEvolutionaryApproachMultiparty2020}, where multiple decision makers (DMs) are involved, each DM represents a department, and at least one DM cares about multiple objectives.

	Suppose the objective set of the $k$-th DM is $M_k$, which represents the set of objectives of interest to the $k$-th DM, which is at least one objective and the objective values of individual $\bold{x}$ at the $k$-th DM can be denoted as $F_{k}(\bold{x}) =\{f_i(\bold{x})|f_i(\bold{x}) \in M_k\}$. An MPMOP that minimizes all objectives can be described as follows\cite{liuEvolutionaryApproachMultiparty2020}.
	\begin{equation}
		\begin{aligned}
			& \min \bold{E(x)}=(\bold{F_{\text{1}}(x)},\bold{F_\text{2}(x)},\dots,\bold{F_\text{K}(x)}),\\
			&	where \begin{cases}
				\bold{F_{\text{1}}(x)}=\{f_{11}(\bold{x}),f_{12}(\bold{x}),\dots,f_{1m_{1}}(\bold{x})\}\\
				\bold{F_{\text{2}}(x)}=\{f_{21}(\bold{x}),f_{22}(\bold{x}),\dots,f_{2m_{2}}(\bold{x})\}\\
				\ \ \ \ \vdots\\
				\bold{F_{\text{K}}(x)}=\{f_{K1}(\bold{x}),f_{K2}(\bold{x}),\dots,f_{Km_{K}}(\bold{x})\},\\
			\end{cases} \\
			&	\ \ \ s.t. \begin{cases}
				g_i(\bold{x}) \leq 0, &  i = 1,\dots,n_g \\
				h_j(\bold{x}) = 0   , &  j = 1,\dots,n_h \\
				\bold{x} \in [x_{min},x_{max}]^d,& 
			\end{cases} \\
		\end{aligned}
	\end{equation}
	where $\bold{E(x)}$ is the set of objective vectors from all the $K$ DMs in the MPMOP. The parameter $m_k$ denotes the number of objectives considered for the $k$-th DM.

	In some cases, a multiparty multiobjective optimization problem can be thought of as a grouping of objective functions (not in the strict sense of grouping, as the objective functions of interest to different decision makers can overlap) for the MOP or MaMOP problem combined with information about the decision makers present in the real-world application.

	We use $\bold{x} \prec_k \bold{y}$ to denote that the decision vector $\bold{x}$ Pareto dominates of the decision vector $\bold{y}$ in the objective space $F_{k}(\bold{x})$ of the $k$-th DM\cite{liuEvolutionaryApproachMultiparty2020}. The objective of solving an MPMOP is to search for solutions that are as close to the Pareto front in the objective space of each DM as possible. In addition, $\bold{x}^{MP}$ is called the multiparty non-dominated Pareto solution, if there does not exist a decision vector $\bold{z}$ such that\cite{chang2022multiparty}:
	\begin{enumerate}
		\item for at least one party $i$, $\bold{z} \prec_i \bold{x}^{MP}$;
		\item for any other party $j$, $\bold{z} \prec_j \bold{x}^{MP}$, or $\bold{z}$ and $\bold{x}^{MP}$ are non-dominated for each other.
	\end{enumerate}

	The main difference between MPMOPs and MOPs is that they introduce multiple DMs, which means that the PS of MPMOPs must be their own from the perspective of multiple DMs. MOPs overlook this widely present requirement in real-world applications. Therefore, if we directly use optimization algorithms designed for multiobjective optimization problems (MOPs) or many-objective optimization problems (MaOPs), they will not effectively differentiate between $\bold{x}^P$ and $\bold{x}^{MP}$, resulting in the solution set that cannot be simultaneously considered by multiple decision makers (DMs), and the guarantees of convergence and diversity cannot be provided\cite{liuEvolutionaryApproachMultiparty2020,10.1007/978-3-030-78811-7_6,song2022multiobjective,chen2023MPUAV,chang2022multiparty,she2022evolutionary,chang2023biparty}.

    It is noteworthy that the core motivation of MPMOPs lies in collaborative decision making by different DMs using a completely consistent population of solutions. Unlike in multitask problems (MTPs)\cite{wei2021review} or multiobjective multi-task problems (MOMTPs)\cite{qiao2023self}, where different tasks can have their populations that may not reside in the same decision variable space (e.g., different numbers of decision variables), in MPMOPs, the set of candidate solutions considered must be identical from all DMs' perspectives. Based on this shared set of candidate solutions, optimization is carried out according to the different objectives of the DMs.
	\subsection{Biparty Multiobjective UAV Path Planning Problem}
	UAVs have gained widespread usage in various urban operations, including aerial photography\cite{li2021real}, unmanned deliveries\cite{lv2021digital}, urban traffic monitoring\cite{yang2022aoi}, and IoT data collection\cite{huang2019differential}. Considering these applications, some universal biparty multiobjective UAV path planning problems (BP-UAVPPs) which involve the objectives for efficiency-related DM, the objectives for safety-related DM, and the constraints in UAV path planning are constructed\cite{chen2023MPUAV}.

	Here, a brief introduction is given. The problems consist of two decision makers, the efficiency decision maker and the safety decision maker. The efficiency decision maker considers objectives regarding path length, flight energy consumption, path height changes, and mission hover point distance and the safety decision maker considers the objectives on fatal risks, property risks, and noise pollution. These objectives are computed based on the given maps (in this paper, two maps called MAP-A and MAP-B are considered) and the given path $(\bold{x})$ of the UAV. For a given path consisting of $n$+1 trajectory points, where the $i$-th discrete trajectory point can be expressed as $\vec{p_i}=(x_i,y_i,z_i)$, the decision vector $\bold{x}$ corresponding to the given path can be expressed as $\bold{x} =(x_0,y_0,z_0,x_1,y_1,z_1,\dots,x_{n},y_{n},z_{n})$. Based on the path trajectory points of the UAV, we can easily calculate the objective values for efficiency decision maker and safety decision maker. The detailed formulas and methods for calculating these objectives are given in Appendix S--B \textit{``Details about Biparty Multiobjective UAV Path Planning Problems''}.

	Based on the above-mentioned different decision makers with different objectives, we further formulate the problem. The efficiency objectives include the minimization of flight path length $f_{\textup{length}}$, the minimization of flight energy consumption $f_{\textup{fuel}}$, the minimization of path height changes $f_{\textup{height}}$ and minimization of the total sum of mission hover point distance $f_{\textup{distance}}$. The safety objectives include the minimization of fatal risks $f_{\textup{fatal}}$, the minimization of property risk $f_{\textup{eco}}$ and the minimization of noise pollution $f_{\textup{noise}}$. Without loss of generality, the model for all cases can be represented generically as follows\cite{chen2023MPUAV}.
	\begin{equation}
		\label{eqt:case2}
		\begin{aligned}
			& \min\  (\bold{F}_{\text{eff}}(\textbf{x}),\bold{F}_{\text{safe}}(\textbf{x})),\\
			& \bold{F}_{\text{eff}}=(f_{\textup{length}}(\textbf{x}),f_{\textup{distance}}(\textbf{x}),\dots),\\
			& \bold{F}_{\text{safe}}=(f_{\textup{fatal}}(\textbf{x}),f_{\textup{eco}}(\textbf{x}),\dots),\\
			& \bold{x} =(x_0,y_0,z_0,x_1,y_1,z_1,\dots,x_{n},y_{n},z_{n}), \\
			&	s.t. \begin{cases}
				H_{min} \leq z_i \leq H_{max}, &  i = 1,\dots,n ;\\
				|\alpha_i| \leq \alpha_{max} , &  i = 1,\dots,n-1 ;\\
				|\beta_i | \leq \beta_{max} , &  i = 1,\dots,n-1 .\\
			\end{cases} \\
		\end{aligned}
	\end{equation}
	where $\bold{F}_{\text{eff}}$ is the set of objectives for the efficiency DM, and $\bold{F}_{\text{safe}}$ is the set of objectives for the safety DM. Here, we construct twelve cases for the experiment. The details of all cases will be given in Section \ref{sc:expuav}.
	\section{Proposed Algorithm}
	\label{sec:mpia}
	This section introduces the motivation behind the proposed strategies and an overview of the multiparty immune algorithm (MPIA). Subsequently, the details of the multiparty non-dominated sorting operation, adaptive activation operation, clone operation, and adaptive operator are presented.
	\subsection{Motivations and Overview}
	\subsubsection{Motivation of Inter-party Guided Crossover}
	MPMOPs typically involve two or more decision makers, each with two or more objectives. At the same time, MPMOPs require that the solution set is as close to the corresponding true Pareto front of each DM as possible, which is more stringent than traditional MOPs. It poses a challenge in determining the appropriate search direction and obtaining suitable individuals.
	Existing MPMOEA algorithms, such as OptMPNDS\cite{liuEvolutionaryApproachMultiparty2020}, OptMPNDS2, MOEA-D/MP\cite{chang2022multiparty,chang2023biparty}, OptMPNDS3\cite{she2022evolutionary}, and BPAIMA\cite{chen2023MPUAV}, have been proposed in recent years. 

	However, these improvements are mainly attributed to the use of multiparty non-dominated sorting methods designed for MPMOPs or more advanced search methods that have been used in ordinary MOEAs. These studies do not fully utilize the guiding information inherent in the objective space of multiple DMs and can improve the search direction to the Pareto front of some DMs. Reasonably utilizing this guiding information can effectively enhance the search capability of the algorithms.

	Such an idea has been widely applied in dynamic multiobjective optimization problems (DMOPs) and is referred to as guide mechanisms\cite{guo2023knee,yu2022correlation,ardeh2021genetic} or transfer learning mechanisms\cite{zuo2023process,chen2023evolutionary,guo2022knowledge,li2022reducing,jiang2020individual}. For instance, Jiang \emph{et al.}\cite{jiang2020individual} proposed an individual-based transfer learning algorithm for DMOPs, utilizing historical search information to guide population re-initialization. Drawing inspiration from such work, we believe that multiparty multiobjective optimization algorithms could leverage the convergence information of individuals across different DMs. Individuals close to at least one DM's PF can guide the population to generate individuals approaching both DMs’ PF.

\begin{figure}[H]
		\centering  
		\includegraphics[scale=0.83,trim=288 230 250 140,clip]{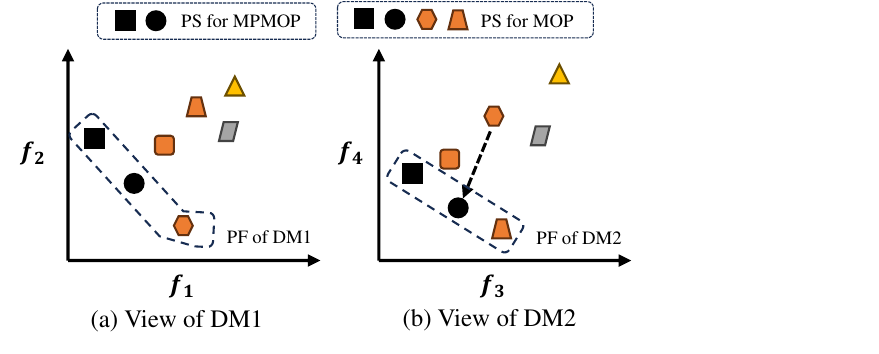}    
		\caption{An example of using inter-party guided crossover. The black circular entities are located on the PF of both DMs. The hexagonal entities are regarded as individuals being guided, while the black circular entities serve as guiding individuals, enabling crossover with the expectation of generating offspring that approach all PF of different DMs.}
		\label{fig:AA}
	\end{figure}

	Therefore, we have designed a crossover operator based on the inter-party guided crossover strategy. This strategy intelligently selects guiding individuals for the guided individual based on the individual's non-dominated ranks from different DM perspectives. We give an example in Fig. \ref{fig:AA} for better understanding. Suppose we consider an MPMOP with two decision makers, i.e.,  DM1 and DM2, and each considering two different decision objectives, totaling four objectives. The population under DM1's perspective is shown in Fig. \ref{fig:AA} (a), and the population under DM2's perspective is shown in Fig. \ref{fig:AA} (b). The hexagonal individuals will serve as the guided individuals, while the circular individuals will serve as the guiding individuals. This strategy aims to search for new individuals located at the current PF of both DMs. The figure illustrates the fundamental concept of the proposed inter-party guide crossover. 
	The details will be given in Section \ref{sec:aos}.
	
	Additionally, using this approach in solving MPMOPs has more intuitive advantages than using it in solving MOPs. In MOPs involving single DM, guiding information is typically provided by a select few individuals near the front. Thus, the population converges towards these guiding individuals. Without proper strategic control, the population may experience potential local convergence and lack of diversity. However, in multiparty multiobjective optimization problems, guidance is required only from the perspective of one DM, leading to a significant increase in the number of individuals capable of offering guiding information. As a result, in multiparty multiobjective optimization problems, most of the population could consist of individuals capable of providing guiding information based on this strategy.
	\subsubsection{Motivation of Adaptive Activation}

\begin{figure}[h]
		\centering  
		\includegraphics[scale=0.88,trim=310 240 250 100,clip]{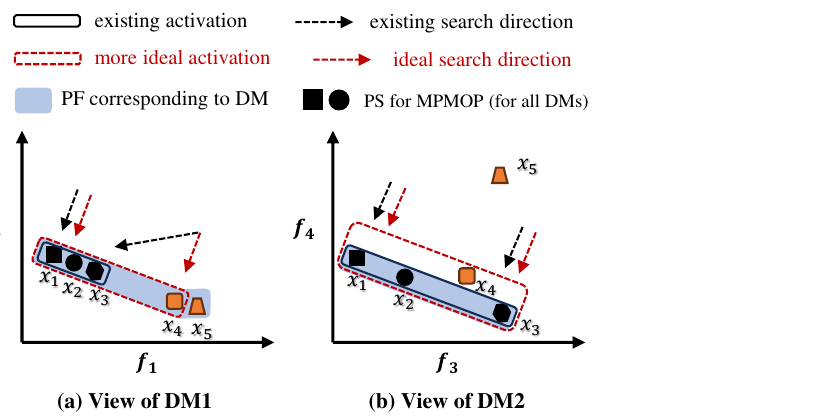}  
		\caption{Inappropriate activations cause the population to lose diversity in the DM1 perspective. The existing method will select $\{x_1,x_2,x_3\}$ as the activation set. However, adding $x_4$ enables the activation set to have higher population diversity for both DMs.}
		\label{AA}
	\end{figure}

	In MOPs, some individuals near the boundary (at least one objective of the achieves minimum value for the whole PS or the corresponding DM) of a single DM must be individuals at the Pareto non-dominated set (because only one DM in ordinary MOPs). However, in MPMOPs, this situation is not always true. Suppose we get the five solutions $\{x_1,x_2,x_3,x_4,x_5\}$ and the information corresponding to solutions is shown in Table \ref{tab:SMALLCASE2}. The individuals $\{x_1,x_2,x_3\}$ marked black in Fig. \ref{AA} constitute the multiparty common PS. The other solutions are non-common PS, but they may still be individuals at the Pareto front or boundary of a single DM. This is because the algorithm uses multiparty non-dominated sorting operations such as the MPNDS operator or the MPNDS2 operator, which makes it possible that some individuals at the boundary of objective space may not be at the multiparty Pareto set.

	\begin{table}[]
        \setlength{\tabcolsep}{0.32cm}
		\renewcommand\arraystretch{1.5}
		\centering
		\caption{The situation of the solutions from different perspectives}
		\scalebox{0.98}{
		\begin{threeparttable}
			\begin{tabular}{cccccc}
				\toprule[1.5pt]
				Solution & Layers\tnote{*} & \makecell[c]{PS for\\ DM1} & \makecell[c]{PS for\\ DM2} &\makecell[c]{PS for \\ MOP} & \makecell[c]{PS for \\ MPMOP} \\
				\midrule
				$X_1$ & (1,1) &\checkmark & \checkmark & \checkmark &\checkmark\\
				$X_2$ & (1,1) &\checkmark & \checkmark & \checkmark &\checkmark\\
				$X_3$ & (1,1) &\checkmark & \checkmark & \checkmark &\checkmark\\
				$X_4$ & (1,2) &\checkmark & $\times$ & \checkmark & $\times$ \\
				$X_5$ & (1,3) &\checkmark & $\times$ & \checkmark & $\times$ \\
				\bottomrule[1.5pt]
			\end{tabular}
			\begin{tablenotes}
				\footnotesize
				\item[*] Non-dominated layers from different DM perspectives
			\end{tablenotes}
		\end{threeparttable}}
		\label{tab:SMALLCASE2}    
	\end{table}

	These individuals, such as $\{x_4,x_5\}$ in Fig. \ref{AA} and table \ref{tab:SMALLCASE2}, which are located only in the PF of the corresponding DMs from the perspective of partial DMs are the main contributors to the population diversity for these DMs and play important roles in guiding search. Methods that solely consider the common frontier individuals (PS for MPMOP) have to some extent compromised the long-term diversity of the entire population.
	
    Unfortunately, existing MPMOEAs directly select a fixed number of individuals from a common Pareto front. As a result, these algorithms fail to maintain diversity among offspring across different DM objective spaces. They also struggle to effectively utilize individuals located on the Pareto front (PF) in certain DM objective spaces to produce offspring, despite the valuable information they contain. 
	
    To better understand this issue, we present a comparison of the real performance of the existing multiparty immune algorithms such as BPAIMA, etc., and our proposed MPIA. Fig. \ref{Asize20_P1} illustrates that when using the existing multiparty immune algorithm to activate individuals, although most of these individuals are located on the common PF of both DMs, they fail to represent the population information of DM1. In contrast, Fig. \ref{Asize50_P1} presents the more ideal activated individuals (selected from the population to produce offspring), where the activated individuals cover the fronts of different DMs. In appendix S--C \textit{``Analysis and Visualization of the Impact of Activation Size''}, we provide a more detailed analysis of the activation size.

    \begin{figure}[h]
    \centering  
    \includegraphics[scale=0.49,trim=30 300 60 310,clip]{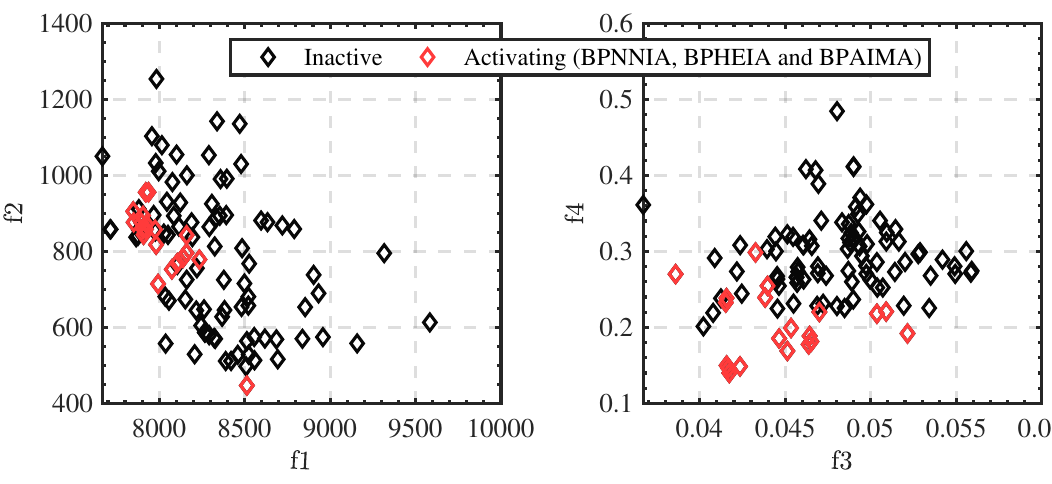}  
    \caption{The population distribution in solving Case 1 using existing multiparty immune algorithms (such as BPNNIA, BPHEIA and BPAIMA) is shown when the number of activations is set to 20. The Case 1 problem is given in Table \ref{table3}.}
    \label{Asize20_P1}
    \end{figure}
    
    \begin{figure}[h]
    \centering  
    \includegraphics[scale=0.5,trim=30 300 60 300,clip]{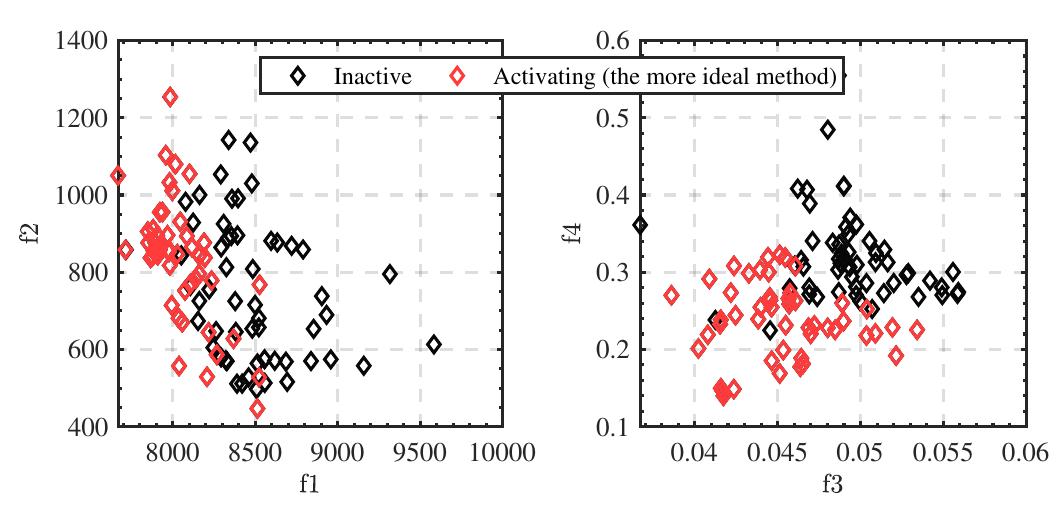}     
    \caption{The population distribution in solving Case 1 using MPIA is shown when the number of activations is adaptive. The Case 1 problem is given in Table \ref{table3}.}
    \label{Asize50_P1}
    \end{figure}

	If the adaptive strategy for selecting the appropriate number of activated individuals can ensure diversity from multiple perspectives, it will effectively enhance the performance of immune algorithms when addressing MPMOPs. Based on this motivation, we have proposed the multiparty cover metric (MCM) to measure the diversity quality of the selected activated individuals, thus further enabling the adaptive activation strategy in algorithm design for solving MPMOPs. The details will be given in Section \ref{sec:aa}.

	\subsubsection{Overview of Proposed Algorithm}
\begin{figure*}[]
		\centering  
		\includegraphics[scale=0.55,trim=40 0 0 0,clip]{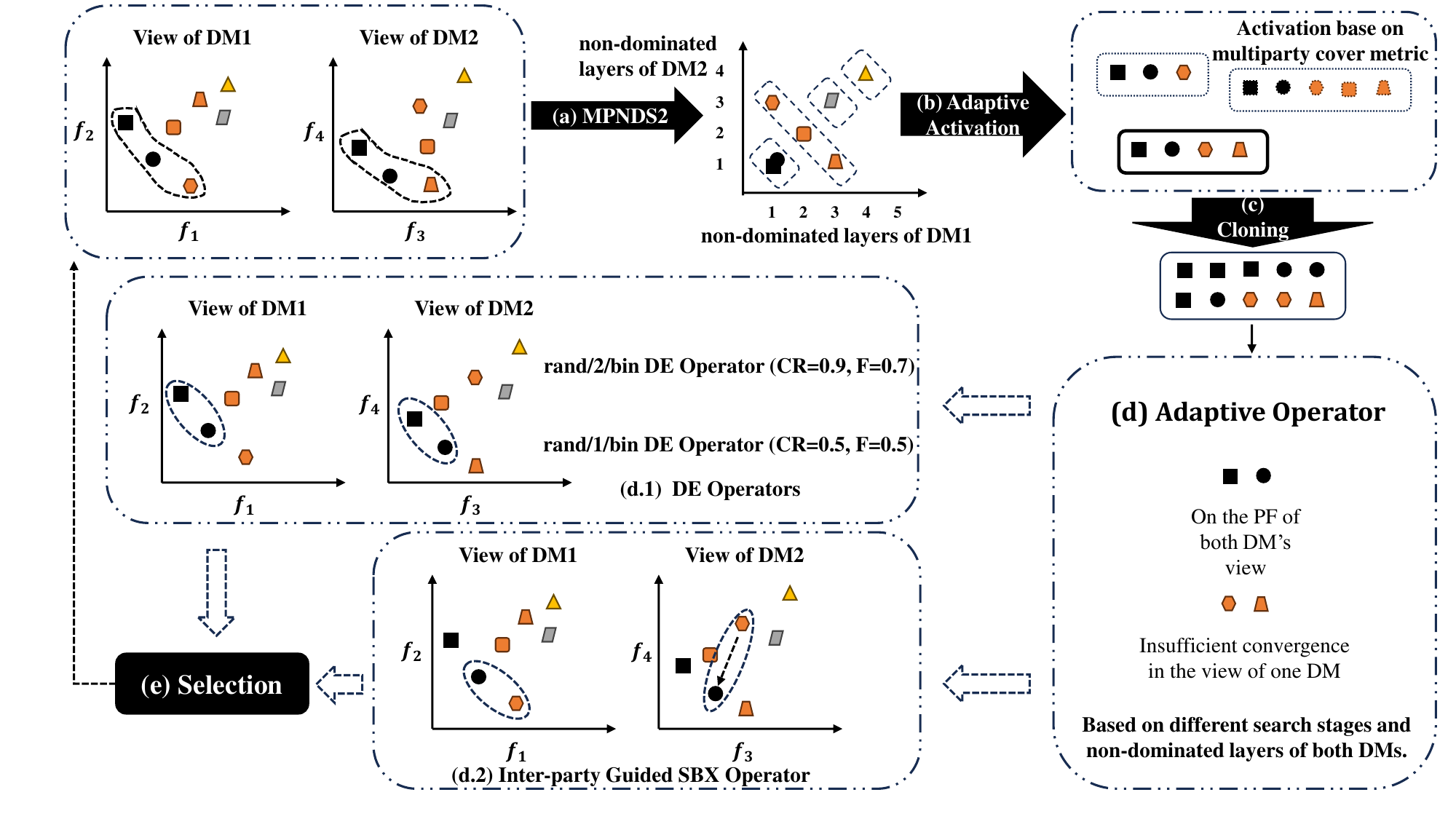}     
		\caption{Overall framework of the MPIA. Here, for better understanding, we assume that there are two parties and each party has two objectives.}
		\label{overview of the algorithm}
	\end{figure*}

	Based on the above motivations, this paper designed a novel immune optimization algorithm for MPMOPs. Fig. \ref{overview of the algorithm} shows the main process of one iteration in MPIA, as shown in Alg. \ref{alg:MPIA}. The first subfigure describes population distributions in the objective space of different DMs' view. The operation (a), i.e., MPNDS2 (Alg. \ref{alg:MPNDSort}) sorts the population to get the multiparty multiobjective non-dominated ranks. The operation (b), i.e., adaptive activation (Alg. \ref{alg:Active}) is used to select the good individuals for both DMs as the parent individuals to produce offspring. The operation (c), i.e., cloning operation (Alg. \ref{alg:Clone}), uses the non-dominated sorting rank information and crowding distance information\cite{debFastElitistMultiobjective2002a} to allocate replication resources. The operation (d), i.e., Alg. \ref{alg:AdaptiveOperator} is the adaptive operation based on the multiple DE operator and inter-party guided crossover operator. The operation (e), i.e., Alg. \ref{alg:es} uses multiparty non-dominated sorting and crowding distance to select the population into the next iteration.
	\begin{algorithm*}[]
		\caption{MPIA}
		\label{alg:MPIA}
		\textbf{Input:} $nC$ (the clone size, and also the population size) \\
		\textbf{Output:} $MPS$ (the multiparty Pareto optimal solutions)
		\begin{algorithmic}[1]
			\State $t = 0, \mathcal{P}_0 = \text{Initialization}(nC)$  \hfill{\textcolor{gray!90!}{\textit{\# Initialize population $\mathcal{P}_0$ using a uniform distribution}}};
			\State $E_0 = \text{Evaluation}(\mathcal{P}_0)$ 
            \hfill{\textcolor{gray!90!}{\textit{\# Calculate the objective values for different DMs $E_0$ of the population $\mathcal{P}_0$.}}};
			\State $\mathcal{P}_{0} = \text{MPNDS2}(\mathcal{P}_{0},\mathcal{F}_0)$
            \hfill{\textcolor{gray!90!}{\textit{\# Sort the population $\mathcal{P}_{0}$ for MPMOPs using Alg. \ref{alg:MPNDSort}.}}};
			\While {$FE$ is not reached}
			\State $\mathcal{A} = \text{AdaptiveActivation}(\mathcal{P}_t)$ 
            \hfill{\textcolor{gray!90!}{\textit{\# Generate activation set $\mathcal{A}$ from population $\mathcal{P}_t$ using Alg. \ref{alg:Active}.}}};
			\State $\mathcal{C} = \text{Clone}(\mathcal{A}, nC)$
            \hfill{\textcolor{gray!90!}{\textit{\# Generate clone individual set $\mathcal{C}$ from activation set $\mathcal{A}$ using Alg. \ref{alg:Clone}.}}};            
			\State $\mathcal{O} = \text{AdaptiveOperator}(\mathcal{P}_t, \mathcal{A}, \mathcal{C}, t)$
            \hfill{\textcolor{gray!90!}{\textit{\# Generate offspring $\mathcal{O}$ from clone individual set $\mathcal{C}$ and $\mathcal{P}_t$ using Alg. \ref{alg:AdaptiveOperator}.}}};       
			\State $\mathcal{P}_{t}  = \mathcal{P}_t \cup \mathcal{O}$
            \hfill{\textcolor{gray!90!}{\textit{\# Combine population $\mathcal{P}_t$ and offspring $\mathcal{O}$}}};             
			\State $E_t = E_t \cup \text{Evaluation}(\mathcal{O})$ 
            \hfill{\textcolor{gray!90!}{\textit{\# Combine objective values of population and objective values of offspring.}}};   
			\State $\mathcal{B} = \text{OptMPNDS2}(\mathcal{P}_{t},E_t)$
            \hfill{\textcolor{gray!90!}{\textit{\# Sort the new population $\mathcal{P}_{t}$ using Alg. \ref{alg:MPNDSort}.}}};   
			\State $\mathcal{P}_{t+1} = \text{Selection}(\mathcal{B},nC)$ 
            \hfill{\textcolor{gray!90!}{\textit{\# Select individual set $\mathcal{P}_{t+1}$ from the sorted population $\mathcal{B}$ using Alg. \ref{alg:es}.}}};   
			\State $t = t+1$
			\EndWhile 
			\State Set $MPS$ as solutions which are Pareto optimal in all DMs in $\mathcal{P}_{t+1}$;
		\end{algorithmic}
	\end{algorithm*}
	\subsection{Multiparty Nondominated Sorting and Selection}
	The multiparty non-dominated sorting operator is designed for MPMOPs to select solutions that closely approximate the Pareto non-dominated front of multiple DMs. In the proposed algorithm, we employed the MPNDS2 operator\cite{10.1007/978-3-030-78811-7_6} to conduct multiparty non-dominated sorting. The MPNDS2 operator\cite{10.1007/978-3-030-78811-7_6}  utilizes two rounds of fast non-dominated sorting for multiparty non-dominated sorting. The first round involves non-dominated sorting from the perspective of each DM, as depicted in lines 2-4 of Alg. \ref{alg:MPNDSort}. The second round entails a non-dominated sort based on the number of non-dominated layers of each DM. The resulting number of layers determines the order of the multiparty non-dominated sort. 
	This approach ensures that individuals ranked at the top by multiple DMs also occupy top positions in the multiparty non-dominated sorting rank of individuals.

	\begin{algorithm}[]
		\caption{MPNDS2\cite{10.1007/978-3-030-78811-7_6}}
		\label{alg:MPNDSort}
		\textbf{Input:} $\mathcal{P}_t$ (the population in the $t$-th iteration), $E_t=(F_{1}=(f_{11},f_{12},\dots,f_{1m_1}),F_{2}=(f_{21},f_{22},\dots,f_{2m_2}),\dots,F_{K}=(f_{K1},f_{K2},\dots,f_{Km_K}))$ (the objective values of population for different DMs) \\
		\textbf{Output:} $\mathcal{P}_t$ (the sorted population)
		\begin{algorithmic}[1]
			\State $\mathcal{L} = \emptyset$;
			\For{$i \in \{1,\cdots ,K\}$}
			\State $\mathcal{P}_t, \mathcal{L}[:,i] = \text{NonDominatedSorting}(\mathcal{P}_t,F_{i})$;
			\EndFor
			\State $\mathcal{P}_t, \mathcal{LL} = \text{NonDominatedSorting}(\mathcal{P}_t, \mathcal{L})$;
		\end{algorithmic}
	\end{algorithm}
	
	For the selection operator, this is the same step as most multiparty multiobjective optimization evolutionary algorithms (MPMOEAs) such as OptMPNDS2\cite{10.1007/978-3-030-78811-7_6} and BPAIMA\cite{chen2023MPUAV}. First, we retain the solutions based on the number of layers of multiparty non-dominated sorting operator MPNDS2 layer by layer, as shown in lines 1-6 in Alg \ref{alg:es}. In a given layer, if the total number of retained solutions exceeds the number of maintained population, the crowding distance information filters out some of the crowded solutions in that layer, as shown in lines 7-9 in Alg \ref{alg:es}.
	
	\begin{algorithm}[]
		\caption{Selection}
		\label{alg:es}
		\textbf{Input:} $\mathcal{B}$ (the candidate population sorted by MPNDS2 operator), $nC$ (the number of cloning) \\
		\textbf{Output:} $\mathcal{P}$ (the selected population)
		\begin{algorithmic}[1]
			\State sort $\mathcal{B}$ by crowding distance for each rank.
			\State $\mathcal{P} = \emptyset, i = 1$;
			\While{$|\mathcal{P}| < nC$}
			\State Add the $i-th$ layer's individuals into $\mathcal{P}$.
			\State $i = i+1$;
			\EndWhile
			\While{$|\mathcal{P}| > nC$}
			\State Remove the least crowded individual in the $i$-th layer from $\mathcal{P}$.
			\EndWhile 
		\end{algorithmic}
	\end{algorithm}
	\subsection{Adaptive Activation}
	\label{sec:aa}
	\begin{figure*}[h]
		\centering  
		\includegraphics[scale=0.76,trim=100 180 100 145,clip]{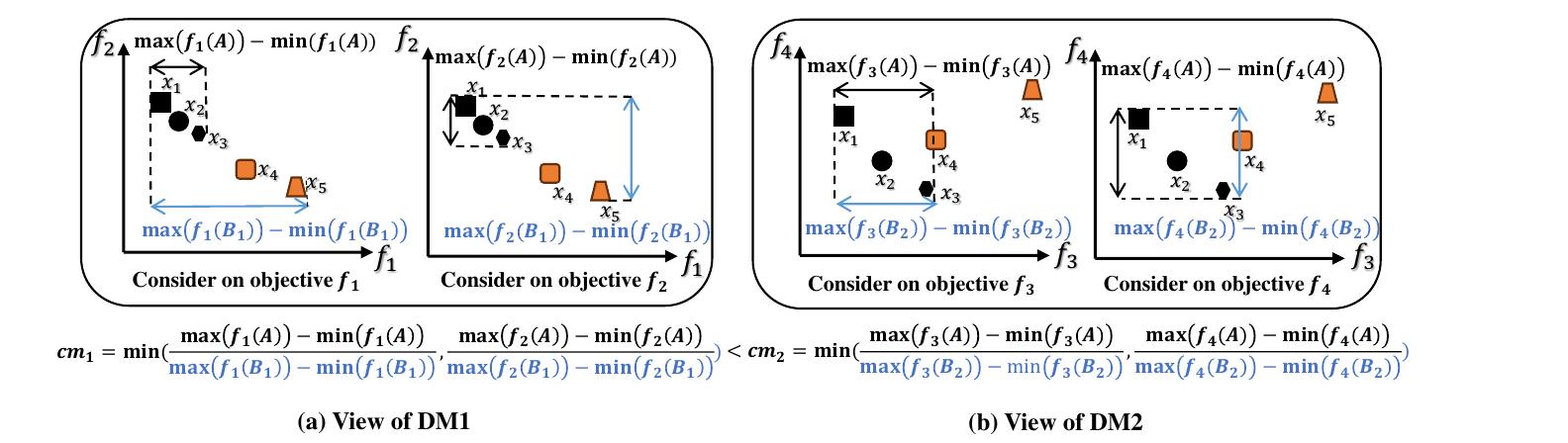}      
		\caption{The relevant details of calculating the cover metric $cm_k$ of the $k$-th DM for the given activation individual set $\mathcal{A} =\{x_1,x_2,x_3\}$. The details about the population are provided in Table \ref{tab:SMALLCASE2}. Where $\mathcal{B}_k$ is the set of frontier individuals for $k$-th DM. $M_k$ denotes the set of objectives for the $k$-th DM.  $M_1=\{f_1,f_2\},M_2=\{f_3,f_4\},\mathcal{B}_1=\{x_1,x_2,x_3,x_4,x_5\},\mathcal{B}_2=\{x_1,x_2,x_3\}$. The figure intuitively demonstrates why the cover metric $cm$ indicator can measure the cover metric of activation individual set $\mathcal{A}$ in different DM perspectives. The metric $cm_2$ is close to 1 and $cm_1$ is much smaller than $cm_2$, indicating that the given activation individual set $\mathcal{A}$ has sufficient diversity for DM2, but lacks diversity for DM1.}
		\label{BB}
	\end{figure*}
	
	The activation operator in immune algorithms is designed to select a certain percentage of high-quality individuals from the population to clone and generate improved individuals. Traditional immune algorithms, such as NNIA\cite{gongMultiobjectiveImmuneAlgorithm2008}, HEIA\cite{linHybridEvolutionaryImmune2015}, AIMA\cite{lin2018adaptive}, as well as multiparty immune algorithms like BPAIMA\cite{chen2023MPUAV}, commonly employ a fixed number of individuals from the Pareto front as active individuals in subsequent cloning and cross-mutate steps. However, this fixed-size activation strategy could result in a loss of diversity when addressing MPMOPs. To tackle this problem, a numerical value known as the multiparty cover metric (MCM) is used to evaluate the performance of different sets of active individuals. Before calculating the MCM, it is necessary to calculate the cover metric ($cm$) for the different decision makers. The cover metric ($cm_k$) for the $k$-th DM is calculated using the following formula.
	
	\begin{equation}
		cm_k = \min_{i \in M_k} \left(\frac{\max(f_i(\mathcal{A})) - \min(f_i(\mathcal{A}))}{\max(f_i(\mathcal{B}_k)) - \min(f_i(\mathcal{B}_k))}\right),
		\label{eq:cm}
	\end{equation}
	where the function $f_i(\cdot)$ represents a list of the $i$-th objective value components of the corresponding individual set. $\mathcal{A}$ represents the given activated individuals, and $\mathcal{B}_k$ is the set composed of some individuals located at the Pareto front of the $k$-th DM among all individuals. $M_k$ denotes the set of objectives for the $k$-th DM. Fig. \ref{BB} visually illustrates the concept of coverage metrics.

	After calculating the cover metrics $cm$ for all DMs, the smallest $cm$ is chosen as the indicator to measure the quality of the given activation individual set. This indicator is called the multiparty cover metric ($MCM$), and its calculation formula is as follows.
	
	\begin{equation}
		MCM = \min_{k=1,\dots, K} cm_k,
		\label{eq:CM}
	\end{equation}
	where $cm_k$ is the cover metric of the $k$-th DM, and $K$ is the number of decision makers. The $MCM$ measures the proportion of the Pareto front distribution range for a given activation individual set $\mathcal{A}$ from the worst-case DM's perspective. A higher value of $MCM$ indicates better diversity for all DMs in the given activation individual set $\mathcal{A}$.
	
	\begin{algorithm}[H]
		\caption{AdaptiveActivation}
		\label{alg:Active}
		\textbf{Input:} $\mathcal{P}_t$ (the sorted population in the $t$-th iteration)\\
		\textbf{Output:} $\mathcal{A}$ (the activated individual set)
		\begin{algorithmic}[1]
			\State $ActivateNumList = \{20,30,40,50,60,70\}$;
			\State Calculate the $B_k$ (PS for $k$-th DM, Refer to Fig. \ref{BB})
			\For{$i \in \{1,\cdots, |ActivateNumList|\}$ }
			\State $nA = ActivateNumList[i]$; // Set the $i$-th list 
			\Statex \ \ \ \ element as the size of activation individual set $\mathcal{A}$.
			\State  $\mathcal{A} = \mathcal{P}_t[1:nA,:] $; 
			\For{$k \in \{1,\cdots, K\}$}
			\State Calculate $cm_k$ according to Eq.  \ref{eq:cm} based on 
			\Statex  \ \ \ \ \ \ \ \ $\mathcal{A},\mathcal{B}_{k}$.;
			\EndFor
			\State Calculate $MCM$ according to Eq.  \ref{eq:CM} based on $cm_k$.;
			\If{$MCM > 0.99$}
			\State \textbf{break};
			\EndIf
			\EndFor
		\end{algorithmic}
	\end{algorithm}
	
	By computing the $MCM$ of a given activation individual set $\mathcal{A}$, we can quantify the diversity of $\mathcal{A}$ for different DMs. The algorithm aims to select an activated individual set $\mathcal{A}$ with a high value of the $MCM$ to maintain sufficient diversity of the offspring. This is easily achievable because as the size of the activation individual set $\mathcal{A}$ increases, the $MCM$ always monotonically increases. However, an excessively large activation individual set $\mathcal{A}$ can cause the algorithm to struggle to effectively distinguish more convergent individuals for MPMOPs, thus compromising the algorithm's convergence performance for MPMOPs. A threshold value (e.g., 0.99) is set to strike a balance. The minimum number of activations, which ensures $MCM$ is no smaller than the threshold value, is selected from a candidate activation number list $ActivateNumList$ (e.g., $\{20,30,40,50,60,70\}$). The pseudocode for this strategy is shown in Alg. \ref{alg:Active}.
	
	\subsection{Clone}
	The cloning operator is a vital component of the immune algorithm 	\cite{gongMultiobjectiveImmuneAlgorithm2008,li2008multiobjective,lin2018adaptive, IMADE, ADE-MOIA,chen2023MPUAV} that facilitates the allocation of computational resources. Assigning more replication resources to the activation individuals with higher quality, increases their opportunities to cross-mutate and generate offspring. In many immune algorithms, resource allocation is primarily determined by the crowding distance. However, considering our adaptive activation strategy, it is possible that some activated individuals may not be Pareto solutions for MPMOP. Therefore, maintaining population convergence becomes crucial. In order to achieve this, we utilize the crowding distance during the cloning process and incorporate a convergence metric to allocate computational resources effectively.
	
	The convergence metric ($p_i$) for the $i$-th activated individual is calculated using the following formula.
	
	\begin{equation}
		p_i = (\max_{j=1,\dots,nA} layer_j) - layer_i,
	\end{equation}
	where $p_i$ represents the convergence metric of the $i$-th activated individual, $nA$ is the number of activated individuals, and $layer_i$ denotes the multiparty non-dominated sorting layers of the $i$-th activated individual. The higher value of $p_i$ indicates better convergence. Based on $p_i$, the formula for determining the number of clones per individual is as follows:
	
	\begin{equation}
		CloneNumList[i]= \lceil nC \cdot \left(\frac{\textup{CD}(A_i) + p_i}{\sum_{j=1}^{nA} (\textup{CD}(A_j) + p_j)}\right) \rceil,
		\label{eq:ri}
	\end{equation}
	where $CloneNumList[i]$ represents the number of replicates for the $i$-th activated individual, $A_i$ represents the $i$-th activated individual, $\textup{CD}(\cdot)$ represents the corresponding crowding distance, $\lceil \cdot \rceil$ is a function that rounds up the value, and $nC$ represents the desired number of individuals to be cloned.
	
	\begin{algorithm}[]
		\caption{Clone}
		\label{alg:Clone}
		\textbf{Input:} $\mathcal{A}$ (the activation individual set), $nC$ (number of clones)\\ 
		\textbf{Output:} $\mathcal{C}$ (the clone individual set)
		\begin{algorithmic}[1]
			\State Calculate the cloning number list $CloneNumList$ of per individual based on {Eq. \ref{eq:ri}}
			\State $\mathcal{C} = \emptyset$;
			\For{$i \in \{1,\cdots ,|\mathcal{A}|\}$}
			\For{$j \in \{1,\cdots ,CloneNumList[i]\}$}
			\State $\mathcal{C} = \mathcal{C} \cup \mathcal{A}[i,:]$;
			\EndFor
			\EndFor
		\end{algorithmic}
	\end{algorithm}
	This method ensures that the MPIA algorithm considers diversity and sufficient convergence during cloning. When all activated individuals are on the first multiparty non-dominated sorting layer for MPMOPs, this method is equivalent to ordinary cloning methods. The pseudocode for this strategy is shown in Alg.  \ref{alg:Clone}.
	
	\subsection{Adaptive Operator}
	\label{sec:aos}
	This section introduces two existing differential evolution (DE) operators and the proposed inter-party guided crossover used in this algorithm. A set of rules has been designed to ensure the appropriate application of these operators during the algorithm process.
	
	Firstly, the first operator is employed for cross-mutation with a probability of $P_1$. The probability $P_1$ of using the first operator is controlled by the following formula, which is related to the number of iterations. This concept is also applied in AIMA\cite{lin2018adaptive} and BPAIMA\cite{chen2023MPUAV}:
	
	\begin{equation}
		P_1 = \frac{0.95}{1+exp(20*\frac{t}{T}-3.0)},
		\label{eq:p1}
	\end{equation}
	where $t$ represents the number of current generations and $T$ denotes the maximum number of generations. This control method enables the algorithm to search with a large step size in the early stage, accelerating the population's convergence and enhancing the search capability of the algorithm.

	\noindent \textbf{First Operator}: The rand/2/bin DE operator \cite{qin2008differential,lin2018adaptive} with a large step size primarily enables the algorithm to exhibit a strong search capability in the early stage.

	\begin{footnotesize}
		\begin{equation}
			\label{eq:op1}
			op1:O_{i,j}=\begin{cases}
				C_{i,j}+F_1*(A_{r_1,j}-A_{r_2,j}) &{\text{if}}\ rand<CR_1 \ or
				\\ \ \ \ \ \ \  +\ F_1*(A_{r_3,j}-A_{r_4,j}) & \ \ \ \ \ \ \ j==randj \\
				C_{i,j} &{\text{otherwise.}}
			\end{cases},
		\end{equation}
	\end{footnotesize}
	where $O$ represents the offspring individuals generated after the operation, $C$ represents the cloned individuals, and $A$ represents the activated individuals. The subscripts ${i,j}$ represent the $j$-th decision variable of the $i$-th corresponding individual. Index $r_1, r_2, r_3, r_4$ are the distinct integers randomly selected from $[1, 2, \dots, nA]$, $nA$ is the size of activated individuals. $CR_1$ means the crossover rate, which is 0.9. $F_1$ means the scaling factor, which is 0.7.
	
	The algorithm uses the second and third operators with a probability of $1-P_1$. When the three conditions ($Con_1$, $Con_2$, $Con_3$) used by the second operator (inter-party guided crossover) are held, the algorithm uses the second operator. Otherwise, the algorithm resorts to the third operator.

	The three conditions ($Con_1$, $Con_2$, $Con_3$) for the second operator are as follows:
	
	\begin{enumerate}
		\item[$Con_1$:] The cloned individual (guided individual) is not located at the multiparty non-dominated forefront after sorting with the MPNDS2 operator. This ensures that the resulting offspring tend to be positioned at the forefront of multiple decision makers.
		\item[$Con_2$:] The finding in the population (guiding individual) is closer or equally close to the forefront of a DM than the guided individual. Additionally, the multiparty non-dominated rank of the guiding individual is smaller than that of the guided individual. This enables the guiding individual to provide effective information to the guided individual.
		\item[$Con_3$:] For all clones of the same guided individuals who have not undergone one inter-party guided crossover operation, then use the guided crossover with probability $P_2$, which is 0.6. This allows an individual and copy to use multiple different crossover methods to maintain population diversity.
	\end{enumerate}
	
	\noindent\textbf{Second Operator}: The inter-party guided crossover operator is then performed using the simulated binary crossover\cite{deb1995simulated}, as shown in the following equation.
	
	\begin{footnotesize}
		\begin{equation}
			\label{eq:op2}
			op2:O_{i,j}=\begin{cases}
				0.5*[(1+\delta)*C_{i,j}+(1-\delta)*A_{guide,j}] &{\text{if}}\ rand<P_c \\
				C_{i,j} &{\text{otherwise.}}
			\end{cases},
		\end{equation}
	\end{footnotesize}
	where $O$ represents the offspring individuals generated after the operation, $C$ represents the cloned individuals, and $A$ represents the activated individuals. The subscripts ${i,j}$ represent the $j$-th decision variable of the $i$-th corresponding individual. Index ${guide}$ is the index of the guiding individual. $P_c$ means the crossover rate, which is the reciprocal of decision variables. 
	$\delta$ is calculated by the following formula\cite{deb1995simulated}.


	\noindent\textbf{Third Operator}: The commonly used rand/1/bin DE operator\cite{lin2018adaptive}, a conventional crossover operator that is used when individuals do not fulfill the conditions of the first and the second approaches. It is a basic crossover.

	\begin{footnotesize}
		\begin{equation}
			\label{eq:op3}
			op3:O_{i,j}=\begin{cases}
				C_{i,j}+F_2*(A_{r1,j}-A_{r2,j}) &{\text{if}}\ rand<CR_2\ or
				\\ \ \ \ \ \ \ \ \ \  & \ \ \ \ \ \ \ j==randj \\
				C_{i,j} &{\text{otherwise.}}
			\end{cases},
		\end{equation}
	\end{footnotesize}
	where $CR_2$ means the crossover rate, which is 0.5. $F_2$ means the scaling factor, which is 0.5.

	After all crossovers, the mutation operation is performed on the individual using the polynomial mutation\cite{debFastElitistMultiobjective2002a} (PM) operator. The pseudocode for this strategy is shown in Alg. \ref{alg:AdaptiveOperator}.

	\begin{algorithm}[]
		\caption{AdaptiveOperator}
		\label{alg:AdaptiveOperator}
		\textbf{Input:} $\mathcal{P}$ (the population set), $\mathcal{A}$ (the activation individual set), $\mathcal{C}$ (the clone individual set), $t$ (current iteration number) \\
		\textbf{Output:} $\mathcal{O}$ (the offspring set)
		\begin{algorithmic}[1]
			\For{$i \in \{1,\cdots ,|\mathcal{C}|\}$}
			\State $r_1,r_2,r_3,r_4 = \text{RandIndex()}$;
			\If{$\text{rand}() < P_1$} \qquad \qquad \qquad \qquad \ \ \ \ \ (Eq. \ref{eq:p1})
			\State $\mathcal{O}[i,:] = \text{Operation1}(\mathcal{C}[i,:],\mathcal{A}[r_1,:],$
			\Statex  \qquad \qquad \qquad \ \  $\mathcal{A}[r_2,:]$,$\mathcal{A}[r_3,:]$,$\mathcal{A}[r_4,:]);$ \qquad \ \ (Eq. \ref{eq:op1})
			\Else
			\State Find the guiding individual $A_{guide}$ for $\mathcal{C}[i,:]$ which
			\Statex \ \ \ \ \ \ \ \ \ meets condition $Con1$\&$Con2$ from population $\mathcal{P}$;
			\If{$ \textbf{exist}(A_{guide}) \ \textbf{and}\ \text{condition}\ Con_3 \ \text{hold} $}
			\State $\mathcal{O}[i,:] = \text{Operation2}(\mathcal{C}[i,:],A_{guide})$;\ \ (Eq. \ref{eq:op2})
			\Else
			\State $\mathcal{O}[i,:] = \text{Operation3}(\mathcal{C}[i,:],\mathcal{A}[r_1,:],\mathcal{A}[r_2,:])$;
			\Statex  \qquad \qquad \qquad \qquad \qquad \qquad \qquad \qquad \qquad \qquad (Eq. \ref{eq:op3})
			\EndIf
			\EndIf
			\State $\mathcal{O}[i,:] = \textup{PolynomialMutation}(\mathcal{O}[i,:])$;
			\EndFor
		\end{algorithmic}
		
	\end{algorithm}
	\section{Metric and Experimental settings}
	\label{sc:set}
	\subsection{Metric}
	For MPMOPs, Liu\emph{ et al.}\cite{liuEvolutionaryApproachMultiparty2020} used the multiparty inverted generational distance (MPIGD\cite{liuEvolutionaryApproachMultiparty2020}) to measure the performance of the algorithms. MPIGD is defined as
	\begin{equation}
		\text{MPIGD}(P^{MP},P)=\frac{\sum_{v \in P^{MP}} d(v,P)}{|P^{MP}|},
	\end{equation}
	\begin{equation}
		d(v,P) = \min_{s \in P}(\sum_{j=1}^{M}\sqrt{(v_{j1}-s_{j1})^2+\dots+(v_{jm_j}-s_{jm_j})^2}),
	\end{equation}	
	where $P^{MP}$ represents the true PF of the MPMOP and $P$ is the PF obtained by the algorithms. $d(v, P)$ represents the minimum distance between $v$ from $P$ and points from $P$. Respectively, $(v_{j1}, . . . , v_{jm})$ means the $m$ objectives of the $j$-th DM for solution $v$, and $(s_{j1}, . . . , s_{jm})$ means the same for solution $s$.
	
	However, there is often no definite true PF for real-world applications to calculate the MPIGD metric. To obtain the performance of different algorithms on MPMOPs, in this paper, the \textit{sumHV} metric is adopted, which is based on the hypervolume (HV) metric\cite{zitzler1999multiobjective,chen2023MPUAV}.
	
	For general MOPs, the HV metric is defined as the super volume constituted by the normalized solution set. Suppose $HV_i$ is the HV metric of the solution set on the set of objectives of the $i$-th DM, then \textit{sumHV} is defined as the sum of $HV_i$ of the solution set on all decision makers shown as follows.
	\begin{equation}
		\textit{sumHV}(P) = \sum_{i=1}^K HV_i(P).
	\end{equation}
	
	The \textit{sumHV} metric examines the performance of the solution set on different decision makers, and the performances of all DMs are chosen as the performance metric of the solution set.

	\subsection{Experimental Settings}
	The experimental comparison involves several algorithms, including the ordinary multiobjective algorithm NSGA-II\cite{debFastElitistMultiobjective2002a}, NSGA-III\cite{deb2013evolutionary} and existing multiparty multiobjective optimization algorithms such as OptMPNDS\cite{liuEvolutionaryApproachMultiparty2020} and OptMPNDS2\cite{10.1007/978-3-030-78811-7_6}. Additionally, multiparty multiobjective optimization algorithms framed in an immune algorithm context were also considered, such as BPNNIA\cite{chen2023MPUAV} that extends from NNIA\cite{gongMultiobjectiveImmuneAlgorithm2008}, BPHEIA\cite{chen2023MPUAV} that extends from HEIA\cite{linHybridEvolutionaryImmune2015}, and BPAIMA\cite{chen2023MPUAV} that extends from AIMA\cite{lin2018adaptive}.

    To comprehensively evaluate MPIA's performance relative to existing multiobjective optimization algorithms and other multiparty multiobjective optimization algorithms, this study designs experiments on synthetic multiparty multiobjective and biparty multiobjective UAV path planning problems. Each experiment is run independently 30 times for all the algorithms. 
 
    Table \ref{tab:parms} shows the parameter settings related to the experiment, NSGA-II, NSGA-III, and MPMOEAs primarily use the default settings in the PlatEMO platform \cite{tianPlatEMOMATLABPlatform2017} or the original paper. NSGA-II, OptMPNDS, OptMPNDS2, BPNNIA, and BPAIMA employ the conventional simulated binary crossover (SBX) operator, while MPIA utilizes the guided-based simulated binary crossover operator. All these algorithms also employ the polynomial mutation operator. BPHEIA uses the rand/1/bin DE operator-I, BPAIMA uses the rand/2/bin DE operator, rand/1/bin DE operator-I and rand/1/bin DE operator-II, while MPIA uses the rand/1/bin DE operator-I and rand/2/bin DE operator. 

    \begin{table}[htbp]
        \setlength{\tabcolsep}{0.1cm}
        \renewcommand\arraystretch{1.5}
        \centering
            \caption{Parameters of Algorithms for Experiments}
        \scalebox{0.9}{
        \begin{threeparttable}
        \begin{tabular}{ll}
            \toprule[1.5pt]
            Parameters & Value  \\
            \midrule
            Population size & 105 \\
            Maximum number of function evaluations (MPUAV-PP) & 1.4e5 \\
            Number of independent runs & 30 \\
            Maximum number of function evaluations (MPMOPs) & $1000*d*M$ \tnote{*}\\
            SBX crossover distribution index & 20 \\
            SBX crossover probability & 1.0 \\
            Guided-based crossover distribution index & 20 \\
            Guided-based crossover probability & 1.0 \\
            PM distribution index & 20 \\
            Decision variables (MPUAV-PP problems) & 88 \\
            Decision variables (MPMOPs) & 20 \\
            Activated num list (MPIA) & $\{ 20,30,40,50,60,70\}$ \\
            Rand/1/bin DE operator-I (BPHEIA, BPAIMA, MPIA) & ($CR = 0.5$, $F = 0.5$) \\
            Rand/1/bin DE operator-II (BPAIMA) & ($CR = 0.1$, $F = 0.5$) \\
            Rand/2/bin DE operator (BPAIMA, MPIA) & ($CR = 0.9$, $F = 0.7$) \\
            \bottomrule[1.5pt]
        \end{tabular}
        \begin{tablenotes}
	\footnotesize
	\item[*] $d$ means the number of the decisions variables, $M$ means the number of the objectives
	\end{tablenotes}
        \end{threeparttable}
        }
        \label{tab:parms}
    \end{table}

    \section{Experiments on Synthetic Multiparty Multiobjective Optimization Problems}
	\label{sc:expmpmop}
	\subsection{Benchmarks and Setup}
	The first public benchmark of MPMOPs with common PS is the MPMOP1 - MPMOP11\cite{liuEvolutionaryApproachMultiparty2020}. The benchmark MPMOPs were derived from the test functions of the CEC'2018 Competition on Dynamic Multiobjective Optimization\cite{jiang2018benchmark}. 
	
	In the synthetic multiparty multiobjective problem experiments. The maximum number of function evaluations was established as $1000*d*M$ for all problems, where $d$ represents the dimension of the decision variables and $M$ represents the number of DMs. Each test problem was run independently 30 times for all the algorithms.
	
	The goal of the corresponding MPMOP is to find the common PS of the MOPs group. The detailed objective function formulation with objective function grouping in the test problem, is given in Ref.\cite{liuEvolutionaryApproachMultiparty2020}.
	
	\subsection{Result and Analysis}
	Table \ref{tab:MPIGD_MPMOP} displays the MPIGD metric of various algorithms in MPMOP1 - MPMOP11.  The MPIA algorithm records the smallest MPIGD in five problems, while other algorithms best address the remaining problems. The results of the Wilcoxon rank sum test with a significance level of 0.05 are presented in the table's final row and the symbols ``$-$'', ``$+$'' and ``$\approx$'' indicate that the performance of the corresponding algorithm is respectively worse than, better than, and similar to that of the proposed MPIA.

	Experimental results reveal performance differences between MOEAs and MPMOEAs in solving MPMOPs. MOEAs, such as NSGA-II and NSGA-III, perform poorly, often failing to attain a common Pareto set (PS) for certain problems. In contrast, MPMOEAs achieve a common PS across all problems, with MPIA demonstrating the best performance in many cases. The primary reason for these experimental results is the differing strategies for ranking or retaining elite individuals between MOEAs and MPMOEAs. The search directions represented by the ranking or retention methods in MOEAs do not fully align with the actual search direction toward the true PS of MPMOPs.

	Compared to other MPMOEAs, MPIA demonstrates the greatest advantages in solving MPMOPs, achieving the best performance on the majority of the tested problems. However, there are instances where MPIA shows only average performance on certain MPMOPs. Therefore, we aim to provide insights regarding the experimental results and the characteristics of these problems to assist researchers in better addressing these issues in the future.
	
	We suspect that this average performance is primarily due to the relatively low search difficulty of some MPMOPs (which only include 10-30 decision variables, making them easier to solve compared to MPUAV-PP problems), indicating that most MPMOEAs can converge to satisfactory results. Specifically, problems such as MPMOP1, MPMOP4, MPMOP6, MPMOP10, and MPMOP11 exhibit similar orders of magnitude in MPIGD metrics across different MPMOEAs, indicating comparable performance among the algorithms on these problems.
	
	One potential explanation could be that all algorithms have not yet converged; however, this possibility is unlikely since similar results persist even when the number of iterations is increased. Another potential explanation is that all algorithms converge to solutions that are very close to the optimal results. This hypothesis is supported by the visualization results presented in Appendix S–A, titled “Visualization Results in Solving Synthetic Multiparty Multiobjective Optimization Problems.” When the populations are sufficiently close to the Pareto front, the use of inter-party guided crossover and adaptive activation strategies does not significantly enhance algorithm performance or yield better results. In fact, these strategies may slightly hinder convergence. Consequently, in this context, inter-party guided crossover and adaptive activation strategies do not offer substantial improvements.

	The performance differences are also due to variations in search frameworks. For example, genetic algorithm-based MPMOEAs like MPNDS and MPNDS2 have different performance compared to immune algorithm-based MPMOEAs like BPNNIA, BPHEIA, BPAIMA, and MPIA. However, when comparing MPIA with other immune algorithm-based MPMOEAs, or with versions of MPIA without inter-party guided crossover and adaptive activation strategies (as detailed in appendix S--E), MPIA shows performance improvements in most problems and significant improvements in a few (e.g., MPMOP2, MPMOP3, and MPMOP8). In solving more complex real-world problems like MPUAV-PP problems (with 88 decision variables), MPIA demonstrates a more noticeable advantage.

	Among them, MPMOP1 to MPMOP6 have two decision makers, while MPMOP7 to MPMOP11 have three decision makers. Each decision maker has two objectives among MPMOP1, MPMOP3, MPMOP7, and MPMOP8. Fig. \ref{fig:mpmop9} depicts the final solution sets obtained by MPIA when solving MPMOP9. Each figure's different subgraph corresponds to the objective space from different DM perspectives. Appendix S--A \textit{``Visualization Results in Solving Synthetic Multiparty Multiobjective Optimization Problems''} provides the corresponding results for all the other MPMOPs These experimental results demonstrate that the final solution sets obtained by MPIA are capable of approximating the true Pareto front and maintaining diversity.

	\begin{figure}[htbp]
		\centering
		\includegraphics[scale=0.3,trim=60 0 0 0,clip]{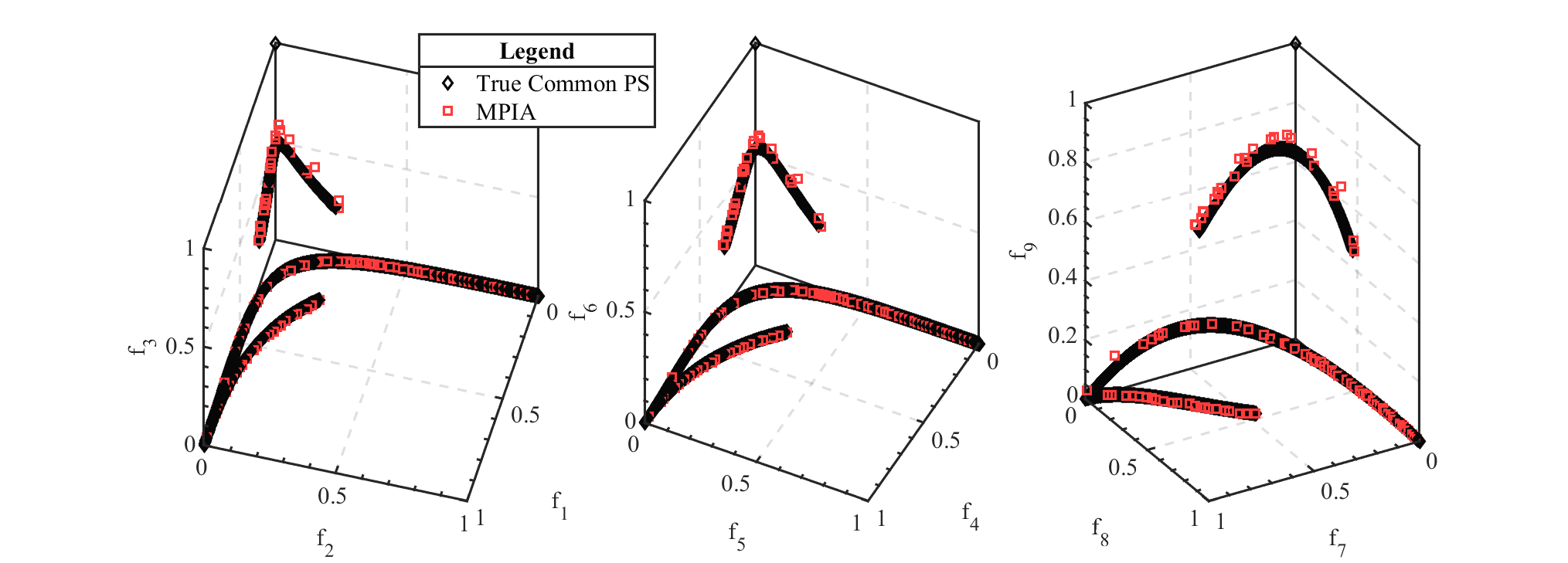}    
		\caption{Results of MPIA in solving MPMOP9}
		\label{fig:mpmop9}
	\end{figure}
	\begin{table*}[htbp]
        \setlength{\tabcolsep}{0.12cm}
		\centering
		\caption{MPIGD metric of different algorithms in MPMOP 1 - MPMOP 12}
		\scalebox{0.85}{
        \begin{tabular}{l|cccccccr}
        \toprule[1.5pt]
        \multicolumn{1}{c}{\multirow{2}[4]{*}{Problems}} & \multicolumn{8}{c}{Algorithms} \\
    \cmidrule{2-9}    \multicolumn{1}{c}{} & NSGA-II & NSGA-III & MPNDS & MPNDS2 & BPHEIA  & BPNNIA  & BPAIMA  & \multicolumn{1}{c}{MPIA} \\
        \midrule
        \midrule
            MPMOP1 & 9.26E-03(3.7E-03) & \textbackslash{} & 1.59E-05(4.8E-06) & \textbf{1.43E-05(4.6E-06)} & 4.15E-05(1.4E-05) & 3.96E-05(1.5E-05) & 6.19E-05(1.6E-05) & \multicolumn{1}{c}{2.89E-05(5.8E-06)} \\
    MPMOP2 & 3.90E-02(4.1E-02) & 3.75E-01(1.7E-01) & 1.28E-02(3.6E-02) & 1.52E-02(3.8E-02) & 3.59E-02(5.5E-02) & 3.28E-02(5.7E-02) & 3.08E-05(1.4E-05) & \multicolumn{1}{c}{\textbf{2.37E-05(8.8E-06)}} \\
    MPMOP3 & 2.25E-01(9.1E-02) & 2.13E-01(4.7E-02) & 2.27E-01(8.9E-02) & 2.10E-01(9.6E-02) & 1.82E-01(9.6E-02) & 2.06E-01(1.3E-01) & 7.86E-02(2.6E-02) & \multicolumn{1}{c}{\textbf{6.84E-02(2.3E-02)}} \\
    MPMOP4 & 1.40E+00(8.0E-01) & 1.17E+00(1.4E+00) & 4.88E-02(6.7E-03) & 5.18E-02(8.5E-03) & 5.85E-02(1.6E-02) & 5.96E-02(1.8E-02) & \textbf{3.03E-02(3.4E-03)} & \multicolumn{1}{c}{3.25E-02(3.1E-03)} \\
    MPMOP5 & 3.56E-01(8.9E-02) & \textbackslash{} & 4.00E-02(3.8E-03) & 4.03E-02(4.1E-03) & 4.31E-02(5.5E-03) & 4.34E-02(5.8E-03) & 3.58E-02(5.7E-03) & \multicolumn{1}{c}{\textbf{3.49E-02(6.9E-03)}} \\
    MPMOP6 & 1.99E+00(1.4E+00) & 2.30E-01(5.9E-02) & 1.50E-02(8.0E-04) & 1.49E-02(1.1E-03) & 1.46E-02(2.1E-03) & \textbf{1.41E-02(2.3E-03)} & 1.81E-02(2.7E-03) & \multicolumn{1}{c}{1.85E-02(1.9E-03)} \\
    MPMOP7 & \textbackslash{} & \textbackslash{} & \textbf{5.42E-06(2.3E-06)} & 6.38E-06(5.0E-06) & 6.52E-05(1.4E-05) & 6.49E-05(1.5E-05) & 2.80E-05(6.9E-06) & \multicolumn{1}{c}{1.88E-05(4.7E-06)} \\
    MPMOP8 & 1.56E-01(1.0E-01) & 6.69E-01(6.1E-01) & 1.54E-01(1.4E-01) & 3.60E-03(2.0E-02) & 7.54E-02(7.9E-02) & 8.02E-02(8.5E-02) & 1.71E-05(7.5E-06) & \multicolumn{1}{c}{\textbf{7.83E-06(3.1E-06)}} \\
    MPMOP9 & 2.00E+00(9.0E-01) & 1.91E+00(1.3E+00) & 7.04E-02(9.8E-03) & 7.28E-02(9.5E-03) & 8.16E-02(1.6E-02) & 7.87E-02(1.5E-02) & 4.41E-02(2.7E-03) & \multicolumn{1}{c}{\textbf{4.40E-02(3.4E-03)}} \\
    MPMOP10 & 6.68E-01(2.5E-01) & \textbackslash{} & 3.88E+00(2.9E+00) & \textbf{3.25E-02(2.1E-03)} & 3.54E-02(2.4E-03) & 3.57E-02(2.6E-03) & 3.58E-02(4.3E-03) & \multicolumn{1}{c}{3.62E-02(3.5E-03)} \\
    MPMOP11 & 3.58E+00(2.9E+00) & 8.72E-01(1.1E+00) & 1.70E-02(7.0E-04) & \textbf{1.67E-02(8.2E-04)} & 1.77E-02(6.5E-04) & 1.75E-02(1.4E-03) & 1.92E-02(7.6E-04) & \multicolumn{1}{c}{1.91E-02(1.2E-03)} \\
				\midrule
				\textbf{($+/\approx/-$)} & 0/0/11 & 0/0/11 & 4/0/7 & 5/0/6 & 2/1/8 & 2/1/8 & 1/6/4 &  \\
				\bottomrule[1.5pt]
			\end{tabular}%
			
		}
		\label{tab:MPIGD_MPMOP}%
	\end{table*}%

	\section{Experiments on Biparty Multiobjective UAV Path Planning Problems}
	\label{sc:expuav}
	\subsection{Problems and Setup}
	In the preceding section, we discussed the objectives relating to both efficiency and safety, as well as constraints associated with UAV performance limits that will be used in this paper's experiments across all application scenarios.

    In the biparty multiobjective UAV path planning (MPUAV-PP) problem experiments, all cases consist of 88 decision variables, with the number of evaluations limited to 140,000. The reference points for calculating HV are the negative ideal points formed by the worst objective function values obtained from the merged solution set of all algorithms involved in the experiment.
	

	In Ref.\cite{chen2023MPUAV}, considering different DMs choosing different conflicting objectives in Map-A to form six biparty biobjective optimization problems. To better test the algorithm's performance, we consider one more map (MAP-B). By considering different maps with different combinations of optimization objectives, the six cases in Ref.\cite{chen2023MPUAV} are further extended to twelve cases in this paper. The detailed objective combinations are displayed in table \ref{table3}. Twelve cases are adopted here to test the algorithms thoroughly. More detailed information about these problems can be found in the supplementary document.
	
	
	
	\begin{table}[htbp]
		\centering
		\caption{Case design}
		\scalebox{0.85}{
			\begin{tabular}{lllc}
				\toprule[1.5pt]
				Problems &  Efficiency DM Objectives   & Safety DM Objectives & Map Type      \\ 
				\cline{1-4}
				Case 1   &  $\textbf{F}_{\text{eff}}=(f_{\textup{length}},f_{\textup{distance}})$  & $\textbf{F}_{\text{safe}}=(f_{\textup{fatal}},f_{\textup{eco}})$   & MAP-A	\\
				Case 2   &  $\textbf{F}_{\text{eff}}=(f_{\textup{length}}+f_{\textup{height}},f_{\textup{distance}})$  & $\textbf{F}_{\text{safe}}=(f_{\textup{fatal}},f_{\textup{eco}})$  & MAP-A    \\
				Case 3  &   $\textbf{F}_{\text{eff}}=(f_{\textup{fuel}},f_{\textup{distance}})$  & $\textbf{F}_{\text{safe}}=(f_{\textup{fatal}},f_{\textup{eco}})$  & MAP-A   \\
				Case 4  &   $\textbf{F}_{\text{eff}}=(f_{\textup{length}},f_{\textup{distance}})$  & $\textbf{F}_{\text{safe}}=(f_{\textup{fatal}},f_{\textup{noise}})$  & MAP-A    \\
				Case 5  &  $\textbf{F}_{\text{eff}}=(f_{\textup{length}}+f_{\textup{height}},f_{\textup{distance}})$ & 
				$\textbf{F}_{\text{safe}}=(f_{\textup{fatal}},f_{\textup{noise}})$& MAP-A \\
				Case 6  &  $\textbf{F}_{\text{eff}}=(f_{\textup{fuel}},f_{\textup{distance}})$  & $\textbf{F}_{\text{safe}}=(f_{\textup{fatal}},f_{\textup{noise}})$ & MAP-A \\
				
				Case 7   &  $\textbf{F}_{\text{eff}}=(f_{\textup{length}},f_{\textup{distance}})$  & $\textbf{F}_{\text{safe}}=(f_{\textup{fatal}},f_{\textup{eco}})$  & MAP-B 	\\
				Case 8   &  $\textbf{F}_{\text{eff}}=(f_{\textup{length}}+f_{\textup{height}},f_{\textup{distance}})$  & $\textbf{F}_{\text{safe}}=(f_{\textup{fatal}},f_{\textup{eco}})$   & MAP-B   \\
				Case 9  &   $\textbf{F}_{\text{eff}}=(f_{\textup{fuel}},f_{\textup{distance}})$  & $\textbf{F}_{\text{safe}}=(f_{\textup{fatal}},f_{\textup{eco}})$ & MAP-B    \\
				Case 10  &   $\textbf{F}_{\text{eff}}=(f_{\textup{length}},f_{\textup{distance}})$  & $\textbf{F}_{\text{safe}}=(f_{\textup{fatal}},f_{\textup{noise}})$ & MAP-B     \\
				Case 11  &  $\textbf{F}_{\text{eff}}=(f_{\textup{length}}+f_{\textup{height}},f_{\textup{distance}})$ & 
				$\textbf{F}_{\text{safe}}=(f_{\textup{fatal}},f_{\textup{noise}})$& MAP-B \\
				Case 12  &  $\textbf{F}_{\text{eff}}=(f_{\textup{fuel}},f_{\textup{distance}})$  & $\textbf{F}_{\text{safe}}=(f_{\textup{fatal}},f_{\textup{noise}})$& MAP-B  \\
				\bottomrule[1.5pt]
			\end{tabular}
			\label{table3}    
		}
	\end{table}
	\subsection{Experimental Results and Analysis}
	
    Table \ref{tab:sumHV_MPUAV} shows the experimental results for all algorithms across Case 1 - Case 12. Each cell displays the mean and variance of the \textit{sumHV} from these experiments. The results from the best-performing algorithm in each case are highlighted in bold. The Wilcoxon rank sum test results with a significance level of 0.05 are presented in the table's final row. The symbols ``$-$'', ``$+$'' and ``$\approx$'' indicate that the performance of the corresponding algorithm is respectively worse than, better than, and similar to that of the proposed MPIA. Regarding the \textit{sumHV} metric, a higher mean value indicates superior fairness and convergence of the algorithm's solutions.
	
    It should be noted that the \textit{sumHV} can sometimes obscure the performance degradation of a specific DM. Thus, to provide a more detailed comparison of the effects of different algorithms on the two DMs, we also present HV metrics computed by different algorithms from different DM perspectives (i.e., HV of the same solution set under different DM's objective set).

    Table \ref{tab:HV_MPUAV} presents the experimental results for the HV metric from the efficiency decision makers and safety decision makers. Each cell displays the mean and variance of the HV from these experiments. The results from the best-performing algorithm in each case are highlighted in bold. This metric measures the quality of the solution set from the perspective of a single DM.

\begin{figure}[htbp]
		\includegraphics[scale=0.3,trim=60 0 0 0,clip]{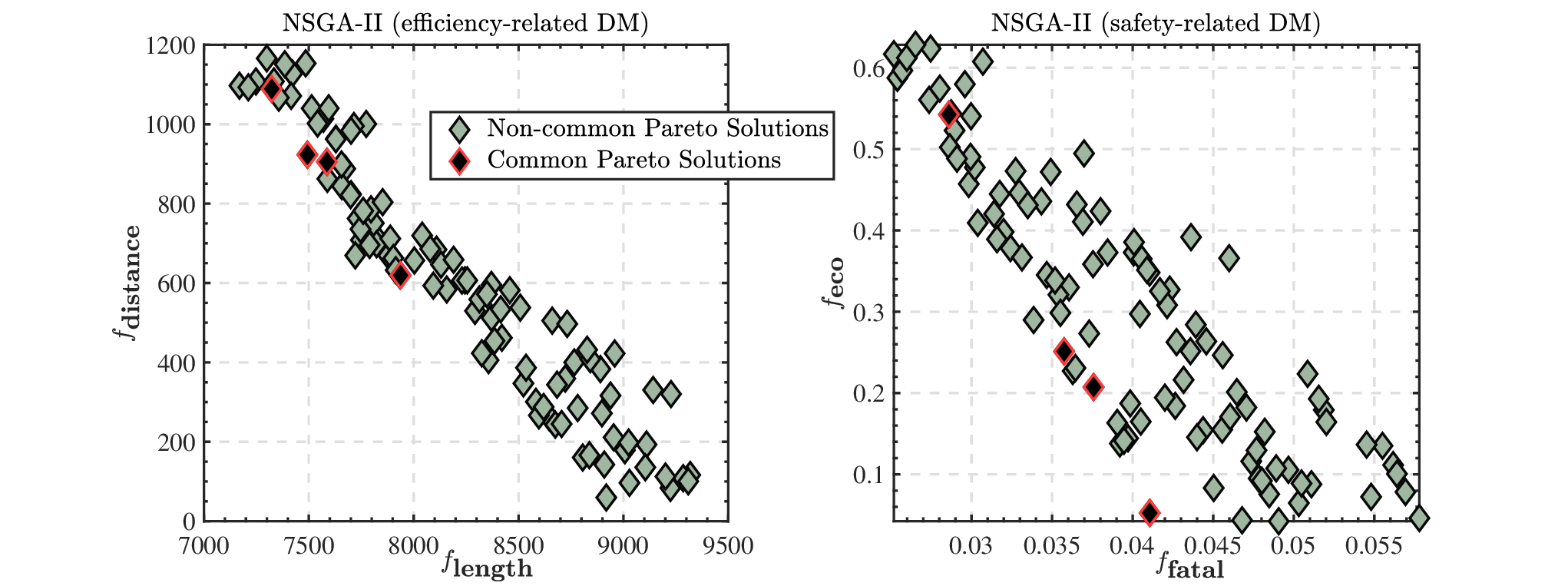}    
		\caption{Final population of NSGA-II in solving Case 1 problem}
		\label{fig:nsgaresult}
	\end{figure}

\begin{figure}[htbp]
		\includegraphics[scale=0.3,trim=60 0 0 0,clip]{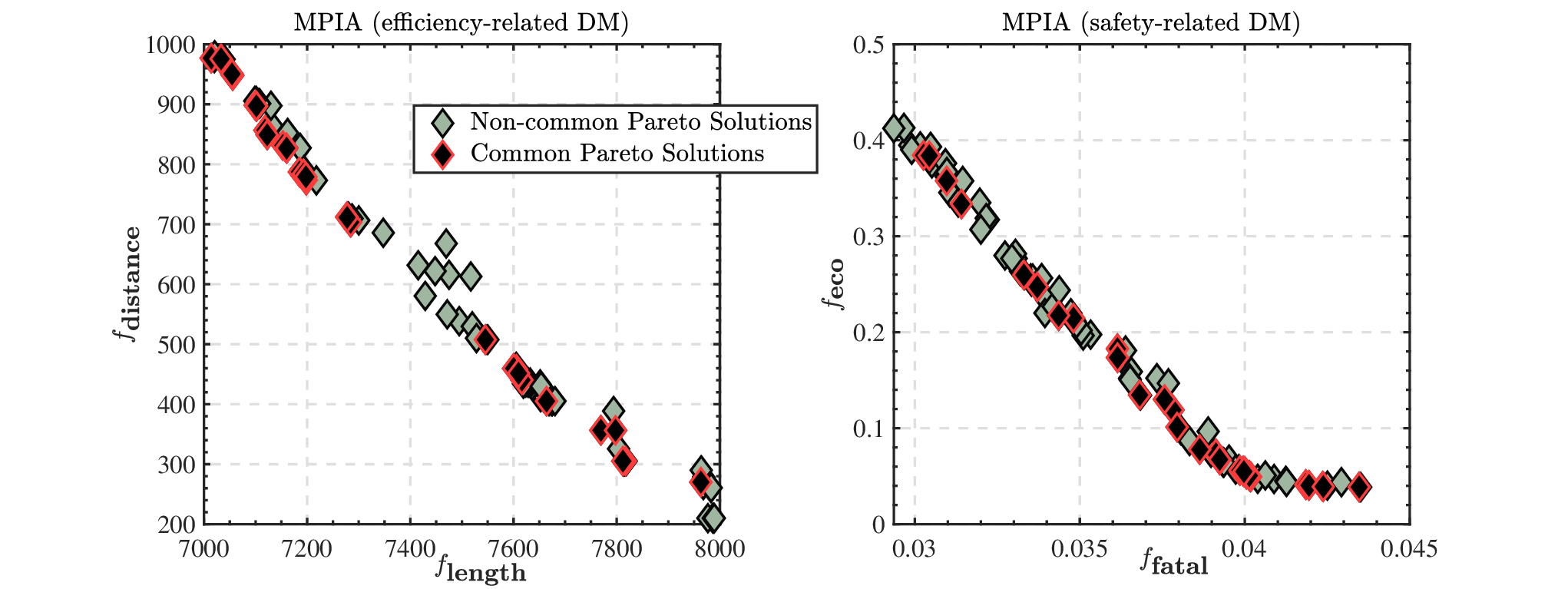}    
		\caption{Final population of MPIA in solving Case 1 problem}
		\label{fig:mpgiaresult}
	\end{figure}

    Fig. \ref{fig:nsgaresult} illustrates the final population of NSGA-II when tackling a biparty multiobjective UAV path optimization problem. Fig. \ref{fig:mpgiaresult} displays the final population of MPIA when solving the same problem. The first subplot illustrates the objective space from the perspective of the efficiency DM, while the second subplot illustrates the objective space from the perspective of the safety DM. In the appendix S--D \textit{``Visualization of MPIA Solutions for MPUAV-PP Problems''}, we provide more detailed visualization results of BPNNIA, MPIA, and MOEAs such as NSGA-II, and NSGA-III solving different cases. These figures underscore that an ordinary multiobjective optimization algorithm, such as NSGA-II and NSGA-III, fails to retain satisfactory proximity to the different DMs' PF when solving this problem. In stark contrast, MPIA solves the problem with most of its final population on the common PF of all decision makers. In another study \cite{chen2023MPUAV}, we analyzed the HV-based metric variations in MOEAs and MPMOEAs when solving MPMOPs.

    The results from table \ref{tab:sumHV_MPUAV} and table \ref{tab:HV_MPUAV} show that the solution results from multiparty multiobjective optimization algorithms such as MPNDS2, BPAIMA, and MPIA significantly outperform NSGA-II and NSGA-III in terms of the \textit{sumHV} or HV of the two DMs. Moreover, the proposed MPIA performs well in both measuring the quality of the solution set for all decision makers using the \textit{sumHV} indicator and measuring the quality of the solution set for a single DM using the HV indicator. It significantly outperforms other compared algorithms in most cases considering different objectives. At the same time, it also demonstrates that the overall performance improvement of MPIA comes from improving the solution set quality from the perspective of all decision makers without compromising the solution set quality from a single DM's perspective. This indicates that the strategy used by MPIA does not sacrifice the interests of any single decision-maker to improve the solution set quality from another decision-maker's perspective. By inter-party guided crossover and considering the adaptive activation strategy of individual MCM indicators under different decision-makers, it effectively balances the optimization goals of different decision-makers. In most cases, the MPIA algorithm can enhance overall performance by improving performance from the perspectives of all decision-makers.
	
\begin{figure}
		\includegraphics[scale=0.25,trim=0 0 0 100,clip]{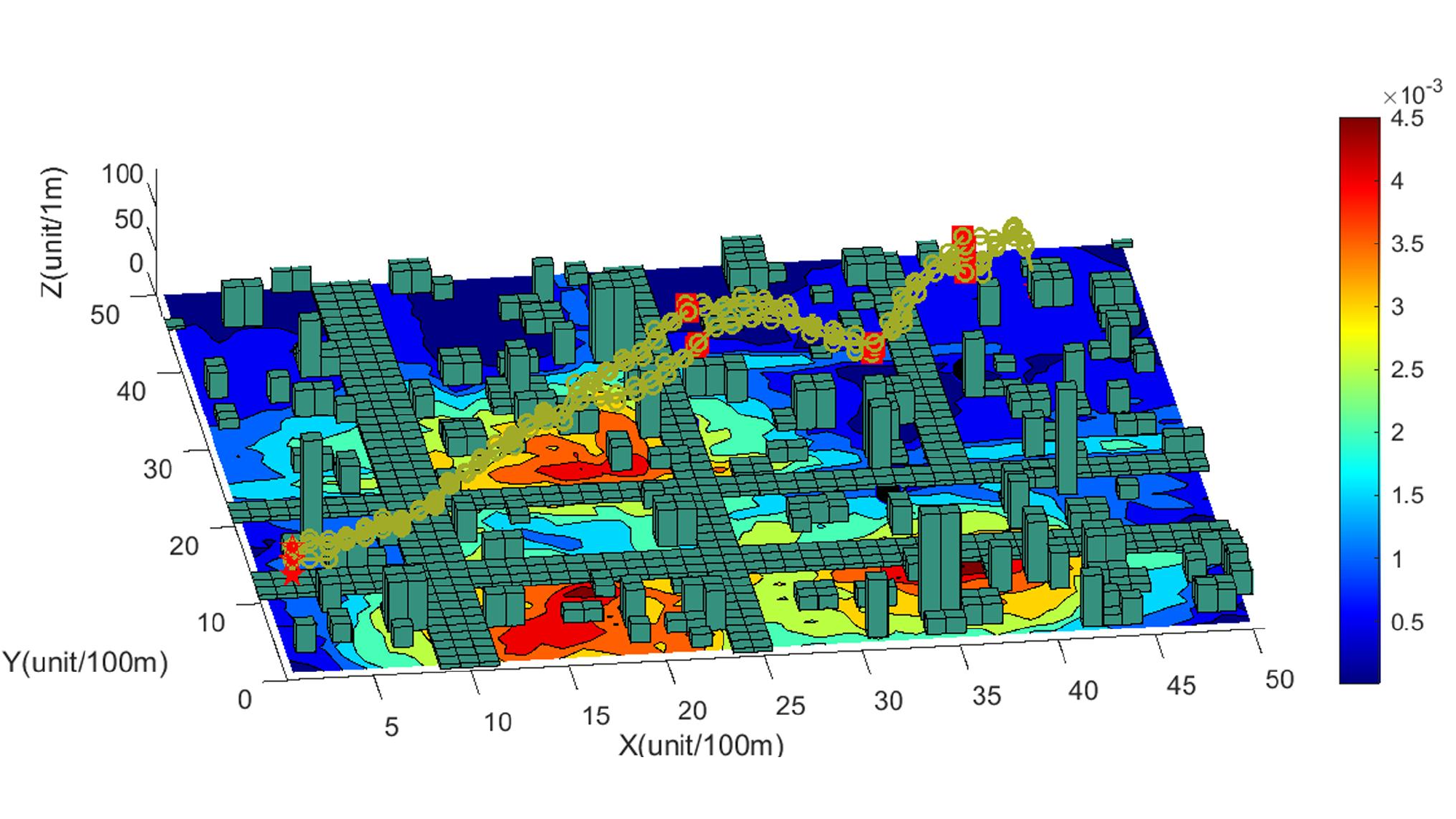}    
		\caption{A part of MPIA's solutions in solving BPUAV-PP Case 1}
		\label{fig:solutions}
	\end{figure}

    Fig. \ref{fig:solutions} visually presents the set of partial solutions obtained by MPIA when solving Case 1, corresponding to the collection of UAV paths. The background map includes building obstacles and a third-party risk heat map. From the visualization, the path set avoids high third-party risk areas, maintains proximity to the predefined hover points, exhibits smooth paths, and chooses nearly straight lines to minimize flight distance. For this problem, the MPIA algorithm effectively generates a feasible and high-quality solution set.
 
\begin{table*}[htbp]
    \centering
    \caption{\textit{sumHV} metric of different algorithms in BPUAV-PP Problem Case 1 - Case 12}
    \scalebox{1}{
        \begin{tabular}{l|cccccccc}
    \toprule[1.5pt]
    \multicolumn{1}{c}{\multirow{2}[4]{*}{Problems}} & \multicolumn{8}{c}{Algorithms} \\
\cmidrule{2-9}    \multicolumn{1}{c}{} & NSGA-II & NSGA-III & MPNDS & MPNDS2 & HEIA  & NNIA  & AIMA  & MPIA \\
    \midrule
    \midrule
    Case 1 & 0.136(0.094) & \multicolumn{1}{l}{0.359(0.114)} & 0.440(0.053) & 0.453(0.048) & 0.395(0.072) & 0.390(0.087) & 0.496(0.055) & \textbf{0.563(0.046)} \\
    Case 2 & 0.292(0.100) & \multicolumn{1}{l}{0.427(0.129)} & 0.451(0.066) & 0.461(0.061) & 0.413(0.085) & 0.396(0.092) & 0.509(0.122) & \textbf{0.565(0.072)} \\
    Case 3 & 0.292(0.106) & \multicolumn{1}{l}{0.391(0.124)} & 0.443(0.060) & 0.450(0.065) & 0.380(0.082) & 0.377(0.098) & 0.495(0.085) & \textbf{0.557(0.069)} \\
    Case 4 & 0.234(0.087) & \multicolumn{1}{l}{0.401(0.112)} & 0.408(0.056) & 0.447(0.065) & 0.383(0.073) & 0.358(0.088) & 0.480(0.075) & \textbf{0.540(0.064)} \\
    Case 5 & 0.229(0.093) & \multicolumn{1}{l}{0.456(0.082)} & 0.421(0.076) & 0.458(0.067) & 0.361(0.089) & 0.378(0.105) & 0.507(0.094) & \textbf{0.518(0.092)} \\
    Case 6 & 0.236(0.111) & \multicolumn{1}{l}{0.453(0.099)} & 0.421(0.063) & 0.434(0.064) & 0.371(0.075) & 0.349(0.091) & 0.485(0.100) & \textbf{0.550(0.057)} \\
    Case 7 & 0.292(0.116) & \multicolumn{1}{l}{0.366(0.111)} & 0.452(0.065) & 0.461(0.067) & 0.383(0.104) & 0.365(0.094) & 0.508(0.051) & \textbf{0.549(0.050)} \\
    Case 8 & 0.324(0.104) & \multicolumn{1}{l}{0.418(0.097)} & 0.467(0.083) & 0.494(0.077) & 0.375(0.110) & 0.390(0.119) & 0.536(0.070) & \textbf{0.576(0.067)} \\
    Case 9 & 0.279(0.110) & \multicolumn{1}{l}{0.371(0.110)} & 0.456(0.046) & 0.464(0.050) & 0.361(0.096) & 0.366(0.108) & 0.492(0.089) & \textbf{0.568(0.043)} \\
    Case 10 & 0.230(0.114) & \multicolumn{1}{l}{0.379(0.121)} & 0.423(0.075) & 0.448(0.066) & 0.338(0.120) & 0.336(0.101) & 0.452(0.089) & \textbf{0.509(0.067)} \\
    Case 11 & 0.249(0.131) & \multicolumn{1}{l}{0.436(0.085)} & 0.448(0.104) & 0.471(0.097) & 0.357(0.126) & 0.340(0.130) & 0.463(0.084) & \textbf{0.535(0.075)} \\
    Case 12 & 0.231(0.116) & 0.407(0.099) & 0.456(0.072) & 0.478(0.060) & 0.365(0.108) & 0.336(0.115) & 0.476(0.083) & \textbf{0.533(0.049)} \\
    \midrule
    \textbf{$(+/\approx/-)$} & 0/0/12 & 0/0/12 & 0/0/12 & 0/0/12 & 0/0/12 & 0/0/12 & 0/1/11 &  \\
    \bottomrule[1.5pt]
    \end{tabular}
    }
			\label{tab:sumHV_MPUAV}%
		\end{table*}%

\begin{table*}[htbp]
    
    \centering
    \caption{HV metric of different algorithms in Efficiency DM and Safety DM of BPUAV-PP Problem Case 1 - Case 12}
    \scalebox{0.95}{
            \begin{tabular}{llcccccccc}
    \toprule[1.5pt]
    \multirow{2}[4]{*}{Problems} & \multirow{2}[4]{*}{Decision Makers} & \multicolumn{8}{c}{Algorithms} \\
\cmidrule{3-10}          &       & NSGA-II & NSGA-III & MPNDS & MPNDS2 & HEIA  & NNIA  & AIMA  & MPIA \\
    \midrule
    \midrule
    \multirow{2}[1]{*}{Case 1} & Effciency & 0.081(0.023) & 0.083(0.027) & 0.108(0.020) & 0.110(0.017) & 0.097(0.015) & 0.103(0.018) & 0.122(0.016) & \textbf{0.134(0.020)} \\
          & Safe  & 0.192(0.060) & 0.275(0.082) & 0.332(0.061) & 0.343(0.054) & 0.298(0.055) & 0.287(0.043) & 0.374(0.046) & \textbf{0.428(0.048)} \\
    \multirow{2}[0]{*}{Case 2} & Effciency & 0.078(0.038) & 0.105(0.048) & 0.150(0.032) & 0.152(0.036) & 0.130(0.040) & 0.135(0.028) & 0.150(0.041) & \textbf{0.172(0.036)} \\
          & Safe  & 0.214(0.082) & 0.323(0.084) & 0.301(0.068) & 0.309(0.067) & 0.283(0.068) & 0.261(0.070) & 0.359(0.074) & \textbf{0.393(0.067)} \\
    \multirow{2}[0]{*}{Case 3} & Effciency & 0.076(0.024) & 0.089(0.026) & 0.118(0.016) & 0.119(0.022) & 0.104(0.024) & 0.112(0.023) & 0.126(0.024) & \textbf{0.141(0.026)} \\
          & Safe  & 0.215(0.080) & 0.302(0.075) & 0.325(0.056) & 0.331(0.052) & 0.277(0.061) & 0.264(0.075) & 0.369(0.058) & \textbf{0.416(0.050)} \\
    \multirow{2}[0]{*}{Case 4} & Effciency & 0.068(0.027) & 0.081(0.019) & 0.090(0.015) & 0.089(0.019) & 0.082(0.013) & 0.084(0.018) & 0.100(0.014) & \textbf{0.103(0.018)} \\
          & Safe  & 0.166(0.062) & 0.320(0.076) & 0.319(0.062) & 0.358(0.067) & 0.301(0.078) & 0.274(0.059) & 0.379(0.068) & \textbf{0.437(0.055)} \\
    \multirow{2}[0]{*}{Case 5} & Effciency & 0.074(0.024) & 0.096(0.035) & 0.127(0.038) & 0.119(0.031) & 0.095(0.028) & 0.122(0.028) & \textbf{0.141(0.036)} & 0.132(0.037) \\
          & Safe  & 0.155(0.056) & 0.360(0.084) & 0.294(0.062) & 0.339(0.067) & 0.266(0.086) & 0.256(0.077) & 0.367(0.070) & \textbf{0.386(0.072)} \\
    \multirow{2}[0]{*}{Case 6} & Effciency & 0.073(0.024) & 0.091(0.026) & 0.098(0.020) & 0.104(0.019) & 0.089(0.018) & 0.095(0.018) & \textbf{0.117(0.017)} & 0.110(0.022) \\
          & Safe  & 0.163(0.069) & 0.362(0.075) & 0.323(0.066) & 0.330(0.056) & 0.283(0.079) & 0.253(0.046) & 0.367(0.076) & \textbf{0.441(0.050)} \\
    \multirow{2}[0]{*}{Case 7} & Effciency & 0.074(0.025) & 0.084(0.033) & 0.105(0.027) & 0.108(0.027) & 0.089(0.025) & 0.094(0.030) & 0.122(0.023) & \textbf{0.129(0.018)} \\
          & Safe  & 0.218(0.077) & 0.283(0.097) & 0.347(0.050) & 0.353(0.055) & 0.294(0.070) & 0.271(0.067) & 0.386(0.055) & \textbf{0.420(0.042)} \\
    \multirow{2}[0]{*}{Case 8} & Effciency & 0.079(0.035) & 0.110(0.042) & 0.146(0.026) & 0.151(0.029) & 0.118(0.042) & 0.124(0.043) & 0.162(0.042) & \textbf{0.170(0.046)} \\
          & Safe  & 0.245(0.082) & 0.309(0.101) & 0.320(0.058) & 0.343(0.055) & 0.257(0.061) & 0.266(0.066) & 0.375(0.097) & \textbf{0.406(0.054)} \\
    \multirow{2}[0]{*}{Case 9} & Effciency & 0.074(0.033) & 0.092(0.029) & 0.122(0.031) & 0.123(0.026) & 0.099(0.031) & 0.106(0.041) & 0.132(0.030) & \textbf{0.146(0.030)} \\
          & Safe  & 0.204(0.085) & 0.279(0.107) & 0.334(0.053) & 0.340(0.063) & 0.262(0.073) & 0.260(0.073) & 0.360(0.076) & \textbf{0.422(0.061)} \\
    \multirow{2}[0]{*}{Case 10} & Effciency & 0.060(0.031) & 0.088(0.025) & 0.088(0.027) & 0.089(0.025) & 0.074(0.028) & 0.074(0.031) & 0.101(0.022) & \textbf{0.103(0.022)} \\
          & Safe  & 0.170(0.062) & 0.291(0.099) & 0.335(0.055) & 0.359(0.071) & 0.264(0.068) & 0.262(0.079) & 0.351(0.070) & \textbf{0.407(0.059)} \\
    \multirow{2}[0]{*}{Case 11} & Effciency & 0.069(0.041) & 0.112(0.039) & 0.128(0.041) & 0.123(0.042) & 0.100(0.043) & 0.099(0.045) & 0.140(0.031) & \textbf{0.142(0.037)} \\
          & Safe  & 0.179(0.062) & 0.324(0.059) & 0.320(0.074) & 0.349(0.073) & 0.258(0.078) & 0.241(0.086) & 0.323(0.096) & \textbf{0.393(0.091)} \\
    \multirow{2}[1]{*}{Case 12} & Effciency & 0.063(0.040) & 0.089(0.036) & 0.099(0.031) & 0.103(0.029) & 0.083(0.032) & 0.086(0.034) & 0.112(0.030) & \textbf{0.113(0.029)} \\
          & Safe  & 0.168(0.078) & 0.318(0.073) & 0.357(0.057) & 0.376(0.063) & 0.282(0.071) & 0.251(0.074) & 0.364(0.089) & \textbf{0.420(0.047)} \\
    \midrule
    \multicolumn{2}{c}{\textbf{Ranksum $(+/\approx/-)$}} & 0/0/24 & 0/0/24 & 0/5/19 & 0/3/21 & 0/0/24 & 0/2/22 & 0/11/13 &  \\
    \bottomrule[1.5pt]
    \end{tabular}%
			}
			\label{tab:HV_MPUAV}%
		\end{table*}%


		
		\section{Conclusion}
		\label{sec:con}
		This paper proposed a novel multiparty immune algorithm (MPIA) for multiparty multiobjective optimization. The algorithm uses the inter-party guided crossover strategy to enhance its searchability in solving MPMOPs. Meanwhile, an adaptive activation strategy based on the multiparty cover metric (MCM) is proposed to address the shortcomings of maintaining population diversity for different DMs in solving MPMOPs. Compared to state-of-the-art multiparty multiobjective evolutionary optimization algorithms (MPMOEAs), the algorithm MPIA proposed in this paper showed good performance in most experiments. In future work, we will further study the corresponding theoretical parts and address more else important real-world applications.
		\bibliographystyle{IEEEtran}
		\bibliography{ref}

@article{lin2018adaptive,
  title={An adaptive immune-inspired multi-objective algorithm with multiple differential evolution strategies},
  author={Lin, Qiuzhen and Ma, Yueping and Chen, Jianyong and Zhu, Qingling and Coello, Carlos A Coello and Wong, Ka-Chun and Chen, Fei},
  journal={Information Sciences},
  volume={430},
  pages={46--64},
  year={2018},
  publisher={Elsevier}
}

@article{singh2022multi,
  title={Multi-objective {NSGA-II} optimization framework for {UAV} path planning in an {UAV}-assisted {WSN}},
  author={Singh, Manish Kumar and Choudhary, Amit and Gulia, Sandeep and Verma, Anurag},
  journal={The Journal of Supercomputing},
  pages={1--35},
  year={2022},
  publisher={Springer}
}

@article{shen2022energy,
  title={Energy-Aware Dynamic Trajectory Planning for {UAV}-Enabled Data Collection in {mMTC} Networks},
  author={Shen, Lingfeng and Wang, Ning and Zhang, Di and Chen, Jun and Mu, Xiaomin and Wong, Kon Max},
  journal={IEEE Transactions on Green Communications and Networking},
  volume={6},
  number={4},
  pages={1957--1971},
  year={2022},
  publisher={IEEE}
}

@article{liu2022drl,
  title={{DRL-UTPS: DRL}-based Trajectory Planning for Unmanned Aerial Vehicles for Data Collection in Dynamic {IoT} Network},
  author={Liu, Run and Qu, Zhenzhe and Huang, Guosheng and Dong, Mianxiong and Wang, Tian and Zhang, Shaobo and Liu, Anfeng},
  journal={IEEE Transactions on Intelligent Vehicles},
  year={2022},
  publisher={IEEE}
}

@inproceedings{10.1007/978-3-030-78811-7_6,
  title = {A New Evolutionary Approach to Multiparty Multiobjective Optimization},
  booktitle = {Proceeding of Advances in Swarm Intelligence},
  author = {She, Zeneng and Luo, Wenjian and Chang, Yatong and Lin, Xin and Tan, Ying},
  editor = {Tan, Ying and Shi, Yuhui},
  year = {2021},
  pages = {58--69},
  publisher = {Springer International Publishing},
  address = {Cham},
  isbn = {978-3-030-78811-7}
}

@article{aitsaadiUAVPathPlanning2022a,
  title = {{UAV} Path Planning Using Optimization Approaches: A Survey},
  author = {Ait Saadi, Amylia and Soukane, Assia and Meraihi, Yassine and Benmessaoud Gabis, Asma and Mirjalili, Seyedali and {Ramdane-Cherif}, Amar},
  year = {2022},
  month = apr,
  journal = {Archives of Computational Methods in Engineering},
  issn = {1134-3060, 1886-1784},
  doi = {10.1007/s11831-022-09742-7}
}

@article{debFastElitistMultiobjective2002a,
  title = {A Fast and Elitist Multiobjective Genetic Algorithm: {NSGA-II}},
  author = {Deb, K. and Pratap, A. and Agarwal, S. and Meyarivan, T.},
  year = {2002},
  month = apr,
  journal = {IEEE Transactions on Evolutionary Computation},
  volume = {6},
  number = {2},
  pages = {182--197},
  issn = {1089778X},
  doi = {10.1109/4235.996017}
}

@inproceedings{ghambariEnhancedNSGAIIMultiobjective2020,
  title = {An Enhanced {NSGA-II} for Multiobjective {UAV} Path Planning in Urban Environments},
  booktitle = {Proceeding of 2020 IEEE 32nd International Conference on Tools with Artificial Intelligence (ICTAI)},
  author = {Ghambari, Soheila and Golabi, Mahmoud and Lepagnot, Julien and Brevilliers, Mathieu and Jourdan, Laetitia and Idoumghar, Lhassane},
  year = {2020},
  month = nov,
  pages = {106--111},
  publisher = {IEEE},
  address = {Baltimore, MD, USA},
  doi = {10.1109/ICTAI50040.2020.00027},
  isbn = {978-1-72819-228-4}
}

@article{gongMultiobjectiveImmuneAlgorithm2008,
  title = {Multiobjective Immune Algorithm with Nondominated Neighbor-Based Selection},
  author = {Gong, Maoguo and Jiao, Licheng and Du, Haifeng and Bo, Liefeng},
  year = {2008},
  month = jun,
  journal = {Evolutionary Computation},
  volume = {16},
  number = {2},
  pages = {225--255},
  issn = {1063-6560, 1530-9304},
  doi = {10.1162/evco.2008.16.2.225}
}

@article{linHybridEvolutionaryImmune2015,
  title = {A Hybrid Evolutionary Immune Algorithm for Multiobjective Optimization Problems},
  author = {Lin, Qiuzhen and Chen, Jianyong and Zhan, Zhi-Hui and Chen, Wei-Neng and Coello Coello, Carlos and Yin, Yilong and Lin, Chih-Min and Zhang, Jun},
  year = {2015},
  journal = {IEEE Transactions on Evolutionary Computation},
  pages = {711--729},
  issn = {1089-778X, 1089-778X, 1941-0026},
  doi = {10.1109/TEVC.2015.2512930}
}

@inproceedings{liuEvolutionaryApproachMultiparty2020,
  title = {Evolutionary Approach to Multiparty Multiobjective Optimization Problems with Common {P}areto Optimal Solutions},
  booktitle = {Proceeding of 2020 IEEE Congress on Evolutionary Computation (CEC)},
  author = {Liu, Wenjie and Luo, Wenjian and Lin, Xin and Li, Miqing and Yang, Shengxiang},
  year = {2020},
  month = jul,
  pages = {1--9},
  publisher = {IEEE},
  address = {Glasgow, United Kingdom},
  doi = {10.1109/CEC48606.2020.9185747},
  isbn = {978-1-72816-929-3}
}

@article{pangUAVPathOptimization2022a,
  title = {{UAV} Path Optimization with an Integrated Cost Assessment Model Considering Third-Party Risks in Metropolitan Environments},
  author = {Pang, Bizhao and Hu, Xinting and Dai, Wei and Low, Kin Huat},
  year = {2022},
  month = jun,
  journal = {Reliability Engineering \& System Safety},
  volume = {222},
  pages = {108399},
  issn = {09518320},
  doi = {10.1016/j.ress.2022.108399}
}

@article{pengDecompositionbasedConstrainedMultiobjective2022,
  title = {A Decomposition-Based Constrained Multi-Objective Evolutionary Algorithm with a Local Infeasibility Utilization Mechanism for {UAV} Path Planning},
  author = {Peng, Chaoda and Qiu, Shaojian},
  year = {2022},
  month = mar,
  journal = {Applied Soft Computing},
  volume = {118},
  pages = {108495},
  issn = {15684946},
  doi = {10.1016/j.asoc.2022.108495}
}

@article{tianPlatEMOMATLABPlatform2017,
	title={{PlatEMO}: A MATLAB platform for evolutionary multi-objective optimization [educational forum]},
	author={Tian, Ye and Cheng, Ran and Zhang, Xingyi and Jin, Yaochu},
	journal={IEEE Computational Intelligence Magazine},
	volume={12},
	number={4},
	pages={73--87},
	year={2017},
	publisher={IEEE}
}

@article{torijaEffectsHoveringUnmanned2020,
  title = {Effects of a Hovering Unmanned Aerial Vehicle on Urban Soundscapes Perception},
  author = {Torija, Antonio J. and Li, Zhengguang and Self, Rod H.},
  year = {2020},
  month = jan,
  journal = {Transportation Research Part D: Transport and Environment},
  volume = {78},
  pages = {102195},
  issn = {13619209},
  doi = {10.1016/j.trd.2019.11.024}
}

@article{zitzler1999multiobjective,
	title={Multiobjective evolutionary algorithms: a comparative case study and the strength {Pareto} approach},
	author={Zitzler, Eckart and Thiele, Lothar},
	journal={IEEE transactions on Evolutionary Computation},
	volume={3},
	number={4},
	pages={257--271},
	year={1999},
	publisher={IEEE}
}

@inproceedings{chang2022multiparty,
  title={Multiparty Multiobjective Optimization By {MOEA/D}},
  author={Chang, Yatong and Luo, Wenjian and Lin, Xin and She, Zeneng and Shi, Yuhui},
  booktitle={Proceeding of 2022 IEEE Congress on Evolutionary Computation (CEC)},
  pages={01--08},
  year={2022},
  organization={IEEE}
}

@article{lv2021digital,
	title={Digital twins in unmanned aerial vehicles for rapid medical resource delivery in epidemics},
	author={Lv, Zhihan and Chen, Dongliang and Feng, Hailin and Zhu, Hu and Lv, Haibin},
	journal={IEEE Transactions on Intelligent Transportation Systems},
	volume={23},
	number={12},
	pages={25106--25114},
	year={2021},
	publisher={IEEE}
}

@article{li2022reducing,
	title={Reducing negative transfer learning via clustering for dynamic multiobjective optimization},
	author={Li, Jianqiang and Sun, Tao and Lin, Qiuzhen and Jiang, Min and Tan, Kay Chen},
	journal={IEEE Transactions on Evolutionary Computation},
	volume={26},
	number={5},
	pages={1102--1116},
	year={2022},
	publisher={IEEE}
}

@article{jiang2020individual,
	title={Individual-based transfer learning for dynamic multiobjective optimization},
	author={Jiang, Min and Wang, Zhenzhong and Guo, Shihui and Gao, Xing and Tan, Kay Chen},
	journal={IEEE Transactions on Cybernetics},
	volume={51},
	number={10},
	pages={4968--4981},
	year={2020},
	publisher={IEEE}
}

@techreport{jiang2018benchmark,
	title={Benchmark functions for the {CEC}'2018 competition on dynamic multiobjective optimization},
	author={Jiang, Shouyong and Yang, Shengxiang and Yao, Xin and Tan, Kay Chen and Kaiser, Marcus and Krasnogor, Natalio},
	year={2018},
	month= {Jan},
	institution={Newcastle University},
	Number= {NCL/TR-CEC2018-DMOP}
}

@article{li2008multiobjective,
	title={Multiobjective optimization problems with complicated {Pareto} sets, {MOEA/D} and {NSGA-II}},
	author={Li, Hui and Zhang, Qingfu},
	journal={IEEE Transactions on Evolutionary Computation},
	volume={13},
	number={2},
	pages={284--302},
	year={2008},
	publisher={IEEE}
}

@article{chen2023MPUAV,
  title={Evolutionary Biparty Multiobjective UAV Path Planning: Problems and Empirical Comparisons},
  author={Chen, Kesheng and Luo, Wenjian and Lin, Xin and Song, Zhen and Chang, Yatong},
  journal={IEEE Transactions on Emerging Topics in Computational Intelligence},
  year={2024},
  publisher={IEEE}
}

@article{IMADE,
	title={An immune multi-objective optimization algorithm with differential evolution inspired recombination},
	author={Qi, Yutao and Hou, Zhanting and Yin, Minglei and Sun, Heli and Huang, Jianbin},
	journal={Applied Soft Computing},
	volume={29},
	pages={395--410},
	year={2015},
	publisher={Elsevier}
}

@article{ADE-MOIA,
	title={A novel hybrid multi-objective immune algorithm with adaptive differential evolution},
	author={Lin, Qiuzhen and Zhu, Qingling and Huang, Peizhi and Chen, Jianyong and Ming, Zhong and Yu, Jianping},
	journal={Computers \& Operations Research},
	volume={62},
	pages={95--111},
	year={2015},
	publisher={Elsevier}
}

@inproceedings{song2022multiobjective,
	title={On Multiobjective Knapsack Problems with Multiple Decision Makers},
	author={Song, Zhen and Luo, Wenjian and Lin, Xin and She, Zeneng and Zhang, Qingfu},
	booktitle={2022 IEEE Symposium Series on Computational Intelligence (SSCI)},
	pages={156--163},
	year={2022},
	organization={IEEE}
}

@article{she2022evolutionary,
	title={Multiparty distance minimization: Problems and an evolutionary approach},
	author={ She, Zeneng  and  Luo, Wenjian  and  Lin, Xin  and  Chang, Yatong  and  Shi, Yuhui },
	journal={Swarm and Evolutionary Computation},
	volume={83},
	year={2023},
}

@article{chang2023biparty,
	title={Biparty multiobjective optimal power flow: The problem definition and an evolutionary approach},
	author={Chang, Yatong and Luo, Wenjian and Lin, Xin and Song, Zhen and Coello, Carlos A Coello},
	journal={Applied Soft Computing},
	volume={146},
	pages={110688},
	year={2023},
	publisher={Elsevier}
}

@article{deb1995simulated,
	title={Simulated binary crossover for continuous search space},
	author={Deb, Kalyanmoy and Agrawal, Ram Bhushan and others},
	journal={Complex systems},
	volume={9},
	number={2},
	pages={115--148},
	year={1995},
	publisher={Citeseer}
}

@article{guo2022knowledge,
	title={A knowledge guided transfer strategy for evolutionary dynamic multiobjective optimization},
	author={Guo, Yinan and Chen, Guoyu and Jiang, Min and Gong, Dunwei and Liang, Jing},
	journal={IEEE Transactions on Evolutionary Computation},
	year={2022},
	publisher={IEEE}
}

@article{chen2023evolutionary,
	title={Evolutionary Dynamic Constrained Multiobjective Optimization: Test Suite and Algorithm},
	author={Chen, Guoyu and Guo, Yinan and Wang, Yong and Liang, Jing and Gong, Dunwei and Yang, Shengxiang},
	journal={IEEE Transactions on Evolutionary Computation},
	year={2023},
	publisher={IEEE}
}

@article{zuo2023process,
	title={Process Knowledge-guided Autonomous Evolutionary Optimization for Constrained Multiobjective Problems},
	author={Zuo, Mingcheng and Gong, Dunwei and Wang, Yan and Ye, Xianming and Zeng, Bo and Meng, Fanlin},
	journal={IEEE Transactions on Evolutionary Computation},
	year={2023},
	publisher={IEEE}
}

@article{huang2019differential,
	title={Differential evolution with a variable population size for deployment optimization in a UAV-assisted IoT data collection system},
	author={Huang, Pei-Qiu and Wang, Yong and Wang, Kezhi and Yang, Kun},
	journal={IEEE Transactions on Emerging Topics in Computational Intelligence},
	volume={4},
	number={3},
	pages={324--335},
	year={2019},
	publisher={IEEE}
}

@article{guo2023knee,
	title={A Knee-Guided Evolutionary Algorithm for Multi-Objective Air Traffic Flow Management},
	author={Guo, Tong and Mei, Yi and Tang, Ke and Du, Wenbo},
	journal={IEEE Transactions on Evolutionary Computation},
	year={2023},
	publisher={IEEE}
}

@article{yu2022correlation,
	title={A correlation-guided layered prediction approach for evolutionary dynamic multiobjective optimization},
	author={Yu, Kunjie and Zhang, Dezheng and Liang, Jing and Chen, Ke and Yue, Caitong and Qiao, Kangjia and Wang, Ling},
	journal={IEEE transactions on Evolutionary Computation},
	year={2022},
	publisher={IEEE}
}

@article{ardeh2021genetic,
	title={Genetic programming with knowledge transfer and guided search for uncertain capacitated arc routing problem},
	author={Ardeh, Mazhar Ansari and Mei, Yi and Zhang, Mengjie},
	journal={IEEE Transactions on Evolutionary Computation},
	volume={26},
	number={4},
	pages={765--779},
	year={2021},
	publisher={IEEE}
}

@article{li2021real,
	title={Real-time tracking algorithm for aerial vehicles using improved convolutional neural network and transfer learning},
	author={Li, Xiaofeng and Wei, Jin and Jiao, Hongshuang},
	journal={IEEE Transactions on Intelligent Transportation Systems},
	volume={23},
	number={3},
	pages={2296--2305},
	year={2021},
	publisher={IEEE}
}

@article{yang2022aoi,
	title={{AoI} optimization in the {UAV}-aided traffic monitoring network under attack: A Stackelberg game viewpoint},
	author={Yang, Yaoqi and Wang, Weizheng and Liu, Lingjun and Dev, Kapal and Qureshi, Nawab Muhammad Faseeh},
	journal={IEEE Transactions on Intelligent Transportation Systems},
	volume={24},
	number={1},
	pages={932--941},
	year={2022},
	publisher={IEEE}
}

@article{qin2008differential,
	title={Differential evolution algorithm with strategy adaptation for global numerical optimization},
	author={Qin, A Kai and Huang, Vicky Ling and Suganthan, Ponnuthurai N},
	journal={IEEE transactions on Evolutionary Computation},
	volume={13},
	number={2},
	pages={398--417},
	year={2008},
	publisher={IEEE}
}

@article{deb2013evolutionary,
  title={An evolutionary many-objective optimization algorithm using reference-point-based nondominated sorting approach, part I: solving problems with box constraints},
  author={Deb, Kalyanmoy and Jain, Himanshu},
  journal={IEEE transactions on evolutionary computation},
  volume={18},
  number={4},
  pages={577--601},
  year={2013},
  publisher={IEEE}
}

@article{qiao2023self,
	title={A self-adaptive evolutionary multi-task based constrained multi-objective evolutionary algorithm},
	author={Qiao, Kangjia and Liang, Jing and Yu, Kunjie and Wang, Minghui and Qu, Boyang and Yue, Caitong and Guo, Yinan},
	journal={IEEE Transactions on Emerging Topics in Computational Intelligence},
	volume={7},
	number={4},
	pages={1098--1112},
	year={2023},
	publisher={IEEE}
}

@article{wei2021review,
	title={A review on evolutionary multitask optimization: Trends and challenges},
	author={Wei, Tingyang and Wang, Shibin and Zhong, Jinghui and Liu, Dong and Zhang, Jun},
	journal={IEEE Transactions on Evolutionary Computation},
	volume={26},
	number={5},
	pages={941--960},
	year={2021},
	publisher={IEEE}
}

\section*{Appendx} 
The supplementary document mainly supplements some experimental results. Section \ref{mpmop} provides the visualization results of the final solution set obtained by MPIA in solving synthetic multiparty multiobjective optimization problems. Section \ref{bpuav} provides the detailed formulas, methods for calculating these objectives and maps used for setting up biparty multiobjective UAV path planning problems. Section \ref{as} provides the visualization and discussion on the impact of activation size. Section \ref{sec:mpuavpp} provides the visualization results of the final solution set obtained by NSGA-II, NAGA-III and our proposed algorithms called MPIA in solving biparty multiobjective UAV path planning problems. Section \ref{ablation} provides the ablation experiments about the proposed methods in main context. 
\subsection{Visualization Results in Solving Synthetic Multiparty Multiobjective Optimization Problems}
\label{mpmop}
\begin{figure}[htbp]
	\includegraphics[scale=0.3,trim=60 0 0 0,clip]{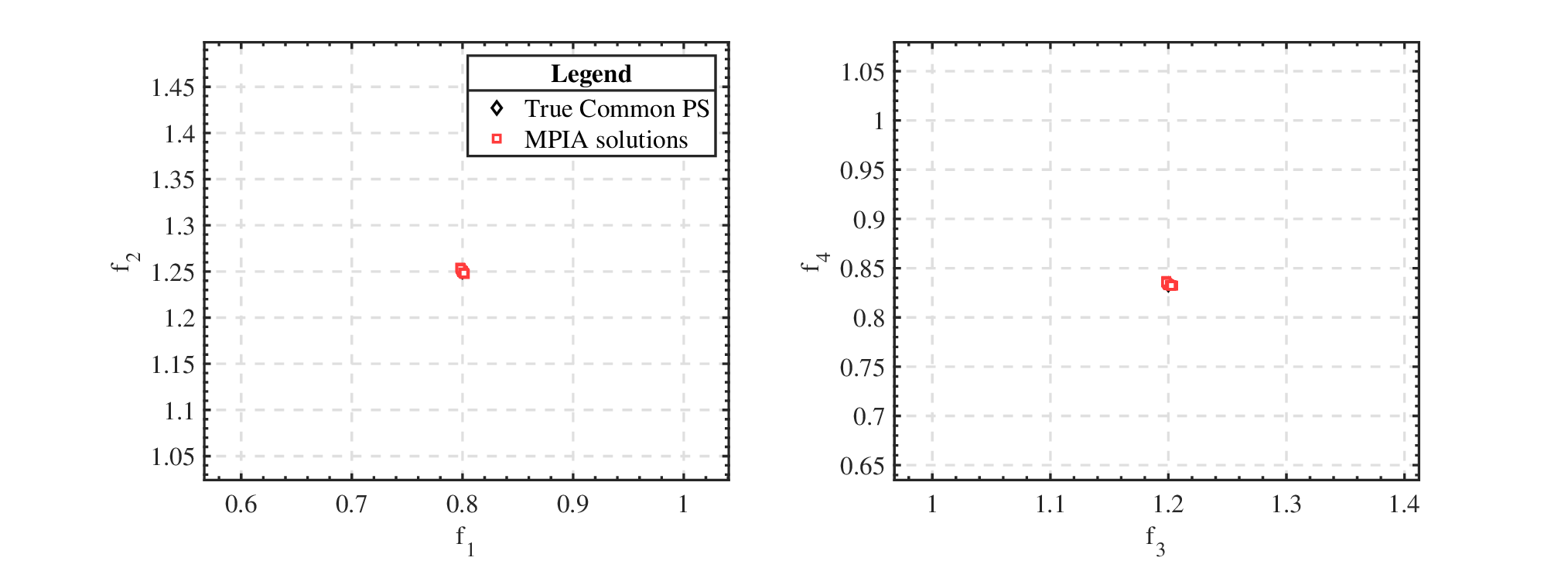}    
	\caption{Results of MPIA in solving MPMOP1}
	\label{fig:mpmop1}
\end{figure}

\begin{figure}[htbp]
	\includegraphics[scale=0.3,trim=60 0 0 0,clip]{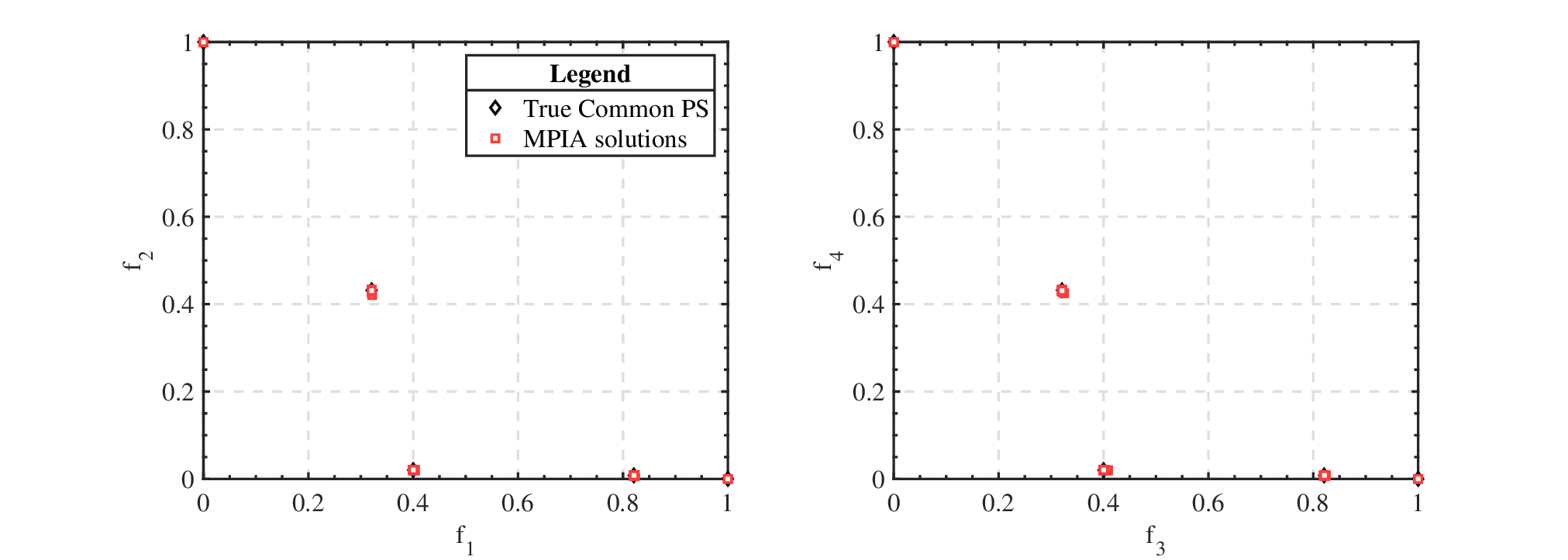}    
	\caption{Results of MPIA in solving MPMOP2}
	\label{fig:mpmop2}
\end{figure}

\begin{figure}[htbp]
	\centering
	\includegraphics[scale=0.3,trim=60 0 0 0,clip]{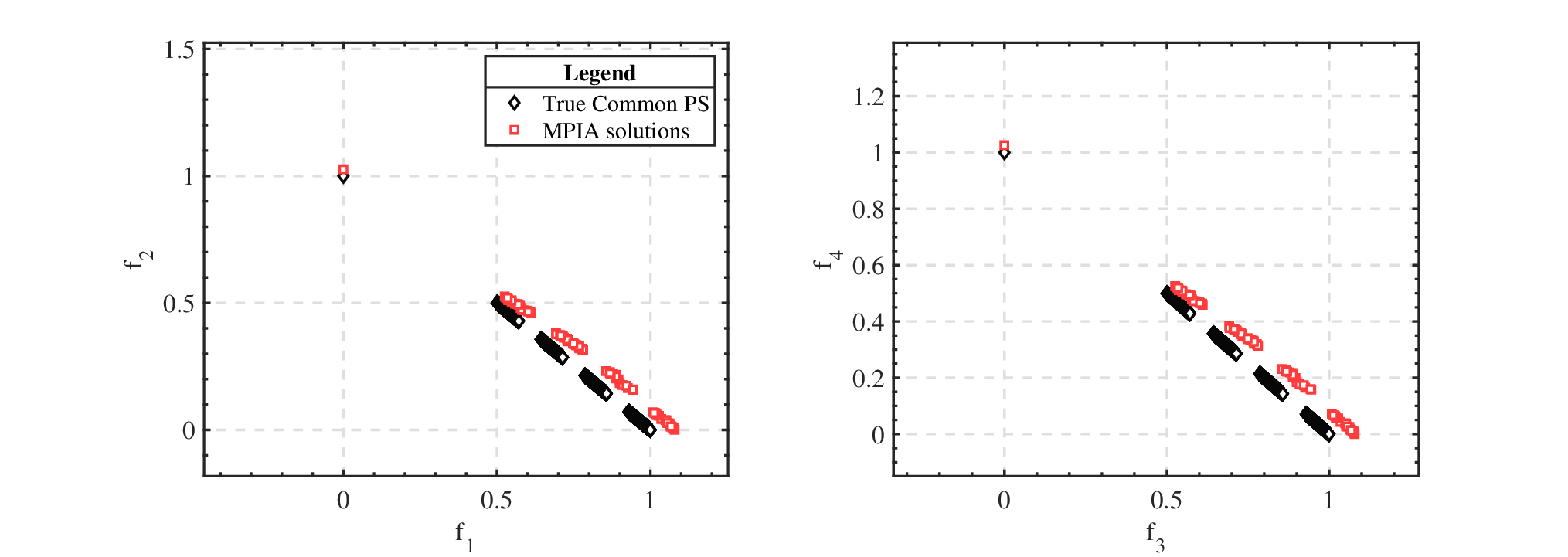}    
	\caption{Results of MPIA in solving MPMOP3}
	\label{fig:mpmop3}
\end{figure}

For MPMOP1 - MPMOP3, each problem involves two decision makers with two objectives. Figs. \ref{fig:mpmop1} to \ref{fig:mpmop3} show the results obtained using MPIA and the true Pareto front (PF). Each figure for a specific problem contains two subgraphs,  with each subgraph corresponding to the objective space from the respective DM perspective.

\begin{figure}[htbp]
	\centering
	\includegraphics[scale=0.3,trim=60 0 0 0,clip]{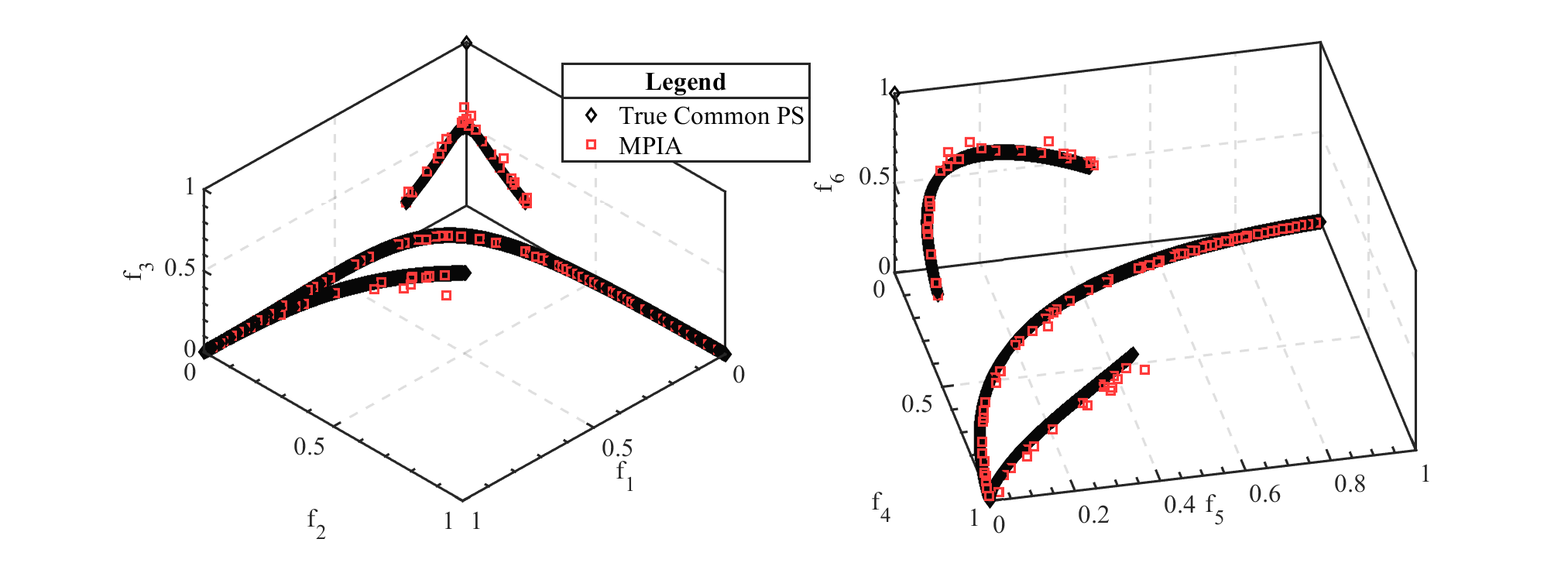}    
	\caption{Results of MPIA in solving MPMOP4}
	\label{fig:mpmop4}
\end{figure}

\begin{figure}[htbp]
	\centering
	\includegraphics[scale=0.3,trim=60 0 0 0,clip]{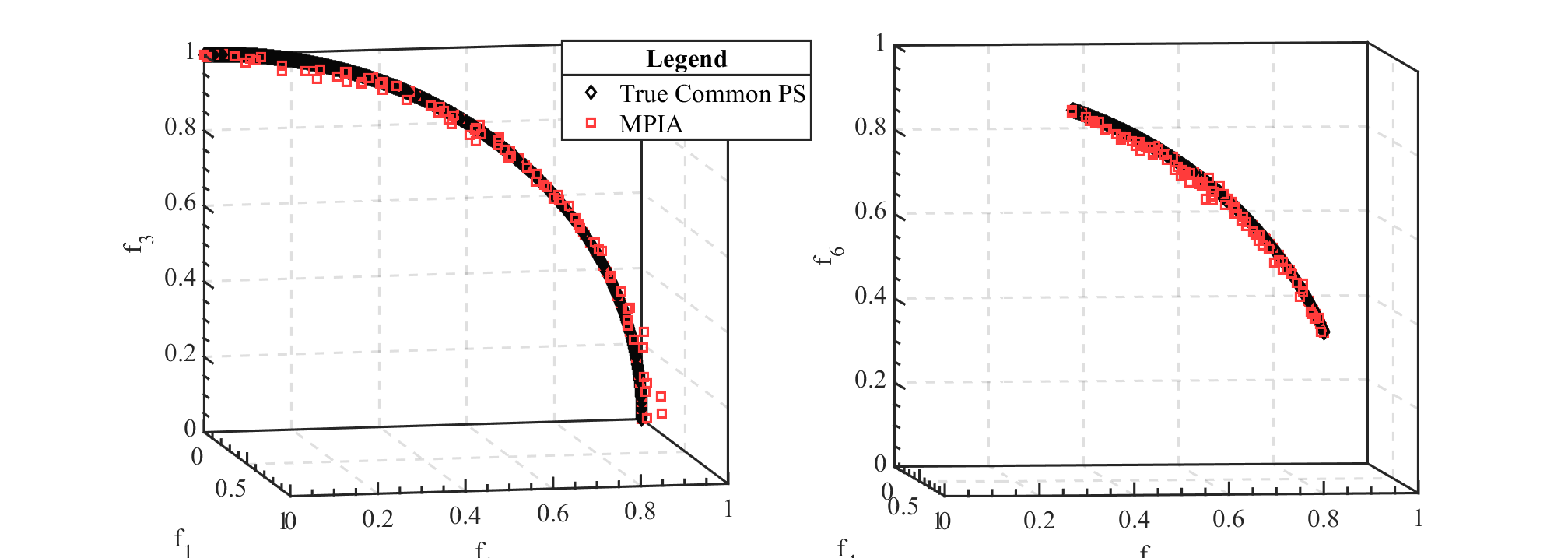}    
	\caption{Results of MPIA in solving MPMOP5}
	\label{fig:mpmop5}
\end{figure}

\begin{figure}[htbp]
	\centering
	\includegraphics[scale=0.3,trim=60 0 0 0,clip]{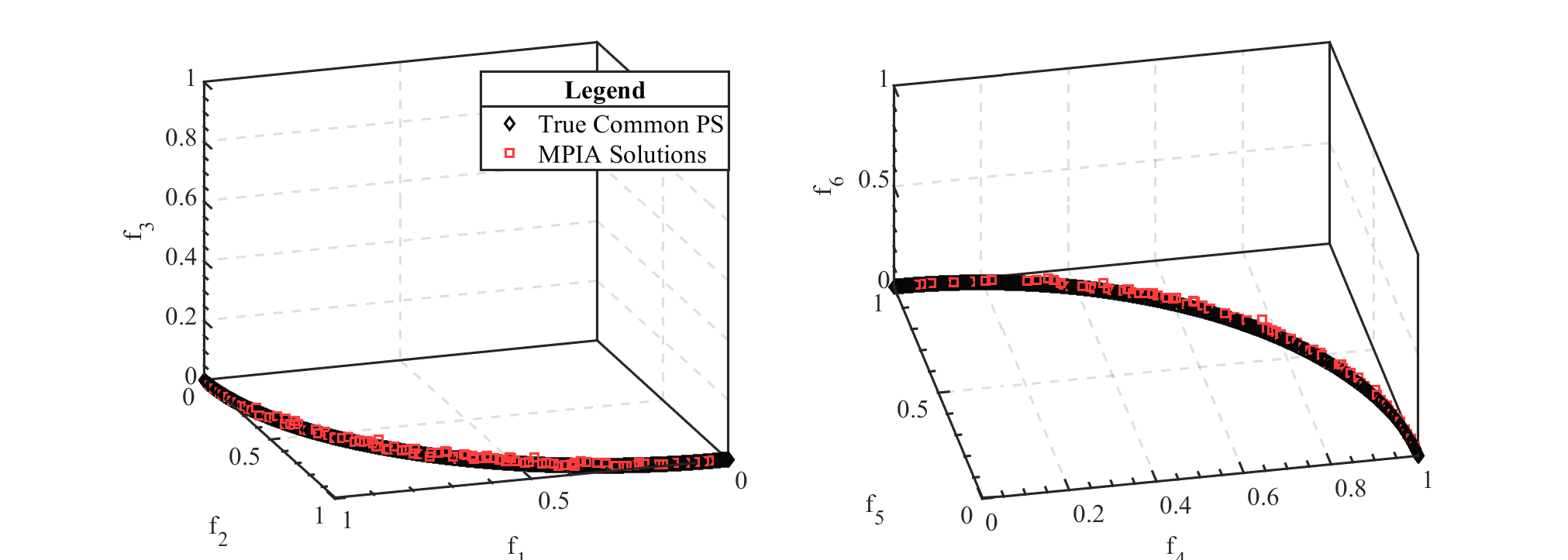}    
	\caption{Results of MPIA in solving MPMOP6}
	\label{fig:mpmop6}
\end{figure}

For MPMOP4 - MPMOP6, each problem involves two decision makers with three objectives. Figs. \ref{fig:mpmop4} to \ref{fig:mpmop6} show the results obtained using MPIA and the true Pareto front (PF). Each figure for a specific problem contains two subgraphs,  with each subgraph corresponding to the objective space from the respective DM perspective.


\begin{figure}[htbp]
	\centering
	\includegraphics[scale=0.3,trim=60 0 0 0,clip]{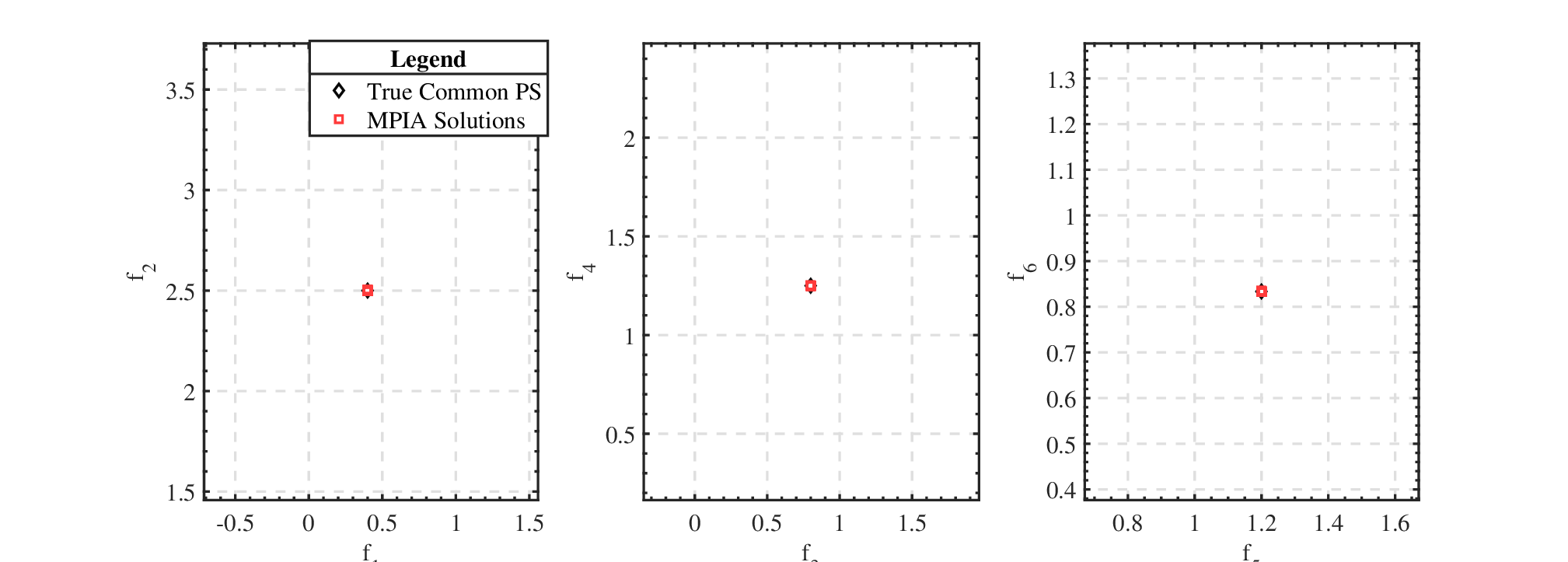}    
	\caption{Results of MPIA in solving MPMOP7}
	\label{fig:mpmop7}
\end{figure}

\begin{figure}[htbp]
	\centering
	\includegraphics[scale=0.3,trim=60 0 0 0,clip]{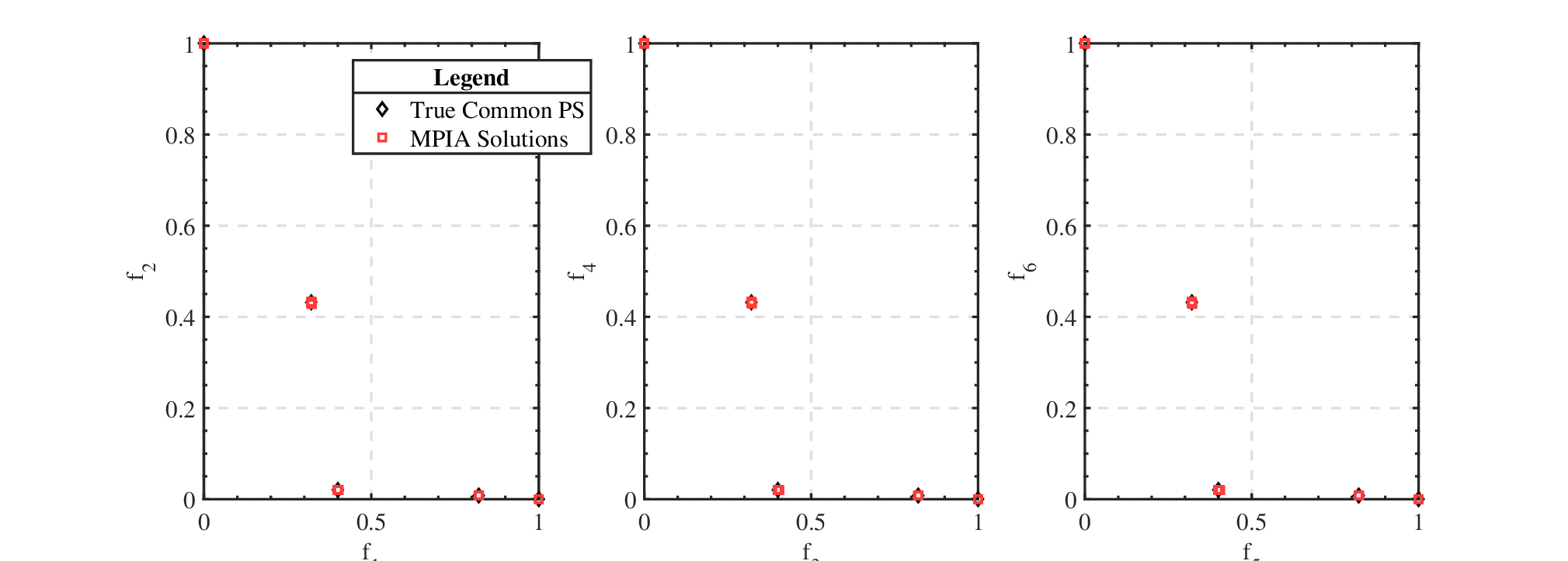}    
	\caption{Results of MPIA in solving MPMOP8}
	\label{fig:mpmop8}
\end{figure}

For MPMOP7 - MPMOP8, each problem involves three decision makers with two objectives. Figs. \ref{fig:mpmop7} to \ref{fig:mpmop8} show the results obtained using MPIA and the true Pareto front (PF). Each figure for a specific problem contains three subgraphs, with each subgraph corresponding to the objective space from the respective DM perspective.


\begin{figure}[htbp]
	\centering
	\includegraphics[scale=0.3,trim=60 0 0 0,clip]{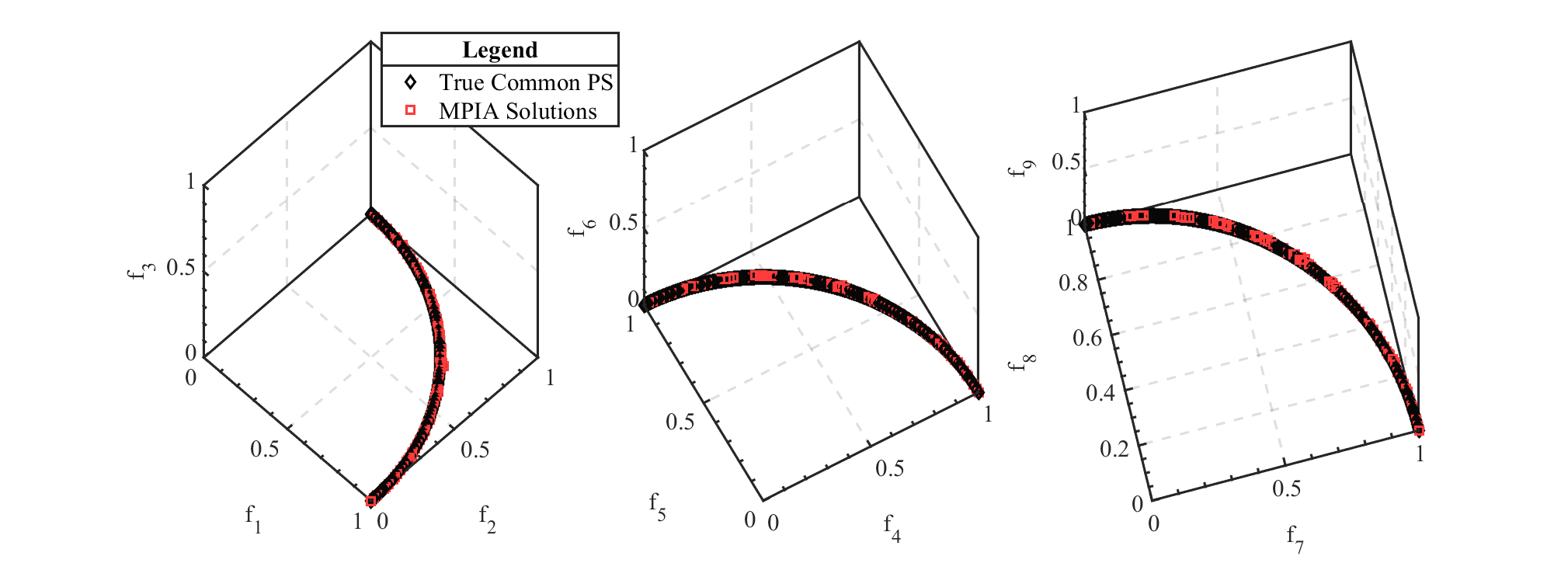}    
	\caption{Results of MPIA in solving MPMOP11}
	\label{fig:mpmop11}
\end{figure}

For MPMOP9 - MPMOP11, each problem involves three decision makers with three objectives. Figs. \ref{fig:mpmop9} to \ref{fig:mpmop11} show the results obtained using MPIA and the true Pareto front (PF). Each figure for a specific problem contains three subgraphs, with each subgraph corresponding to the objective space from the respective DM perspective.

\begin{figure}[htbp]
	\centering
	\includegraphics[scale=0.3,trim=60 0 0 0,clip]{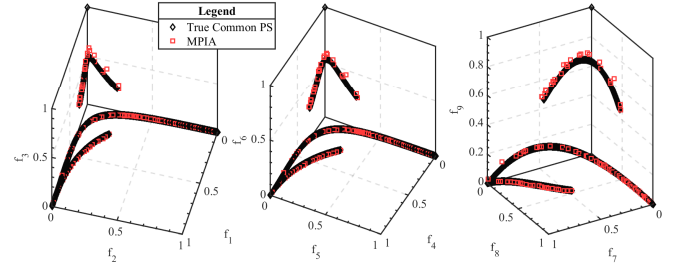}    
	\caption{Results of MPIA in solving MPMOP9}
	\label{fig:mpmop9}
\end{figure}
\begin{figure}[htbp]
	\centering
	\includegraphics[scale=0.3,trim=60 0 0 0,clip]{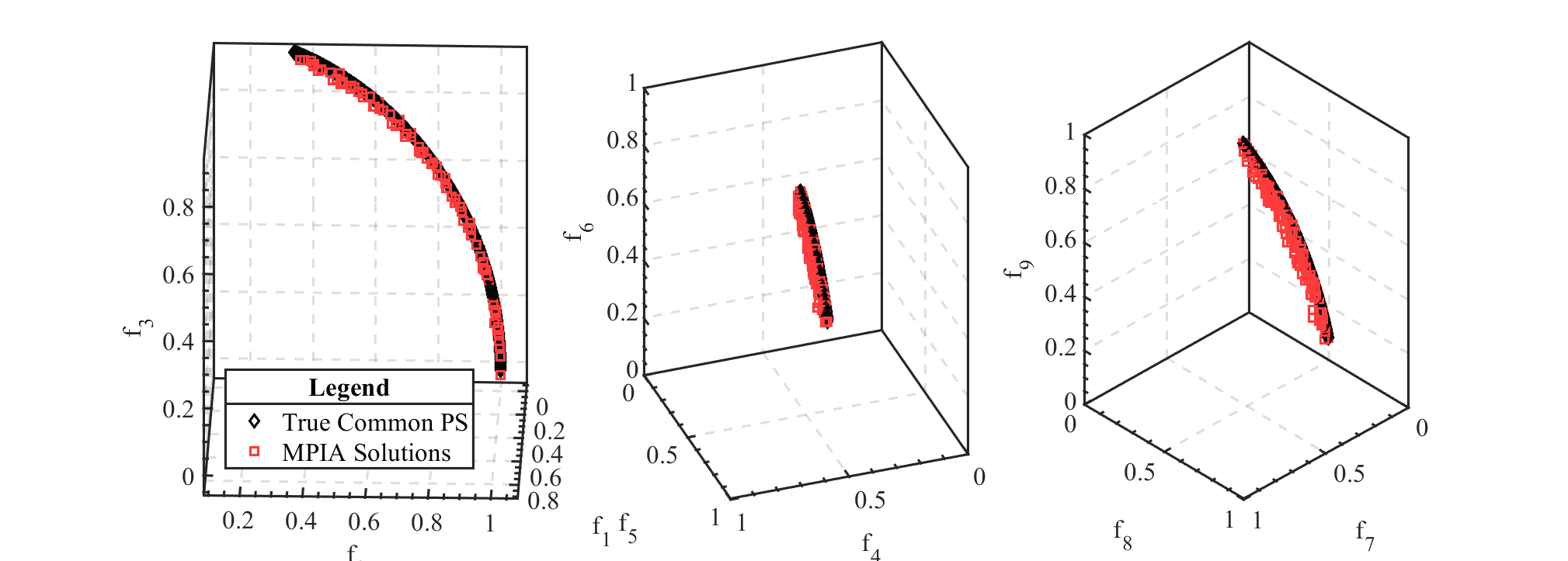}    
	\caption{Results of MPIA in solving MPMOP10}
	\label{fig:mpmop10}
\end{figure}

\subsection{Details about Biparty Multiobjective UAV Path Planning Problems}

\label{bpuav}
Here, we will discuss in detail the decision makers included in the problem, the corresponding objective functions considered, and the details of the maps used, and finally give the constructed problem.

For a given path consisting of $n$+1 trajectory points, where the $i$-th discrete trajectory point can be expressed as $\vec{p_i}=(x_i,y_i,z_i)$, the decision vector $\bold{x}$ corresponding to the given path can be expressed as $\bold{x} =(x_0,y_0,z_0,x_1,y_1,z_1,\dots,x_{n},y_{n},z_{n})$. Based on the path trajectory points of the UAV, we can easily calculate the objective values for efficiency decision maker and safety decision maker.

\subsubsection{Efficiency Decision Maker}
For the efficiency-related DM, maximizing the efficiency of the UAV flight is the core demand of the efficiency-related DM.  Here, path length, Flight energy consumption, path height changes and the mission hover point distances are considered\cite{chen2023MPUAV}.

(a) Path length \textup{\cite{pengDecompositionbasedConstrainedMultiobjective2022}}: In a UAV mission, the planned path length of the UAV directly affects the time of UAV operation. The $i$-th track segment can be expressed as $l_i=\vec{p}_{i+1}-\vec{p}_{i}$. The objective function is shown as follows.
\begin{equation}
	f_{\textup{length}} = \sum_{i=0}^{n-1}||l_i|| .
\end{equation}

(b) Flight energy consumption \textup{\cite{ghambariEnhancedNSGAIIMultiobjective2020}}: In the $i$-th track segment, the power consumption formula is given as follows.
\begin{equation}
	f_{\textup{fuel}}= \sum_{i=0}^{n-1} W^{\frac{3}{2}}\sqrt{\frac{G^3}{2\rho \zeta n}}\frac{||l_i||}{V}+WG\max(z_{i+1}-z_i,0),
\end{equation}
where parameter $W$ is the weight of the UAV and the battery, $\rho$ is the fluid density of air, $G$ is gravity, $n$ is the number of rotors, $||l_i||$ is the track length, $V$ is the flight speed, $\zeta $ is the area of the rotating blade disk. The atmospheric density as a function of altitude can be expressed as $\rho(h)=\rho_0e^{(-(Z_i+Z_{i+1})/(2*10.7))},$
where $\rho_0$ is the atmospheric density and is taken as 1.225$kg/m^3$.

(c) Path height changes \textup{\cite{aitsaadiUAVPathPlanning2022a}}: Fewer UAV altitude changes mean fewer climb and descent phases. As an objective function, the UAV altitude variation at all trajectory points is shown in (\ref{eqt:height_changes}).
\begin{equation}
	\label{eqt:height_changes}
	f_{\textup{height}} = \sum_{i=0}^{n-1} |z_{i+1}-z_{i}| .
\end{equation}

(d) Mission hover point distance:
In the UAV-assisted wireless sensor network, Refs.\cite {shen2022energy, liu2022drl, singh2022multi} proposed the application of UAVs to collect urban sensor data, which requires drone hover points (UHPs) to perform tasks in order for UAVs to collect sensor data during hover. UAV needs to minimize the sum of the distances to all preset UHPs. The objective function can be represented as follows.
\begin{equation}
	f_{\textup{distance}} = \sum_{k=1}^{n_k} \min_{i\in\{0,\dots,n\}} ||\vec{p_i}-p_k^{job}||,
\end{equation}
where $p_k^{job}=(x_k^{job},y_k^{job},z_k^{job})$ is the position of the $k$-th preset UHP, $n_k$ is the number of the preset UHP, and $\vec{p_i}$ is the $i$-th discrete trajectory points.
\subsubsection{Safety Decision Maker}
The core requirement of the safety-related DM is to minimize the negative impact of UAVs. Refs.\cite{pangUAVPathOptimization2022a,chen2023MPUAV} considered the third-party risks posed by UAV. In this paper, the main objectives of the safety-related DM are to minimize the risks to people, vehicles, property, and noise pollution.

(a) Fatal Risks\cite{pangUAVPathOptimization2022a,chen2023MPUAV}:
Fatal risks including the risk of people fatalities and the risk of vehicular fatalities are considered. The overall fatal risk is shown as follows.

\begin{equation}
	f_{\textup{fatal}} = \sum_{i=0}^{n}c_{r_p}(x_i,y_i,z_i)+\sum_{i=0}^{n}c_{r_v}(x_i,y_i,z_i).
\end{equation}

The risk cost associated with pedestrian fatalities, i.e., $c_{r_p}$, is expressed as follows.
\begin{equation}\label{crp}
	c_{r_p}(x_i,y_i,z_i)= P_{crash}S_{h}\sigma_{p}R^{p}_{f}(z_i) = \frac{P_{crash}S_{h}\sigma_{p}(x_i,y_i)}{1+\sqrt{\frac{\alpha}{\beta}}(\frac{\beta}{\frac{1}{2}mv(z_i)^2})^{\frac{1}{4S_{c}}}},
\end{equation}
where $P_{crash}$ mainly depends on the reliability of the UAV itself. The fatal risk index of people $R^{P}_{f}$ is related to the kinetic energy of the impact and the obscuration factor. $S_{h}$ is the size of the UAV crash impact area, $\sigma_{p}(x_i,y_i)$ is the population density within the administrative unit. $\alpha$ and $\beta$ are the energy with $S_{c}$ probability to death and the energy with $100\%$ to death, respectively, $m$ is the mass of the UAV, and $v$ is the velocity of the UAV when it hits the ground. The velocity $v$ is shown as follows.
\begin{equation}
	v(z_i)=\sqrt{\frac{2mg}{R_IS_{h}\rho(1-e^{(z_iR_IS_{h}\rho/m)})}},
\end{equation}
where $g$ is the gravitational constant. $R_I$ is the drag coefficient related to the UAV type and $\rho$ is the fluid density of air.

Similarly, the fatal risk index of vehicles can be expressed as $c_{r_v}(x_i,y_i,z_i) = P_{crash}S_{h}\sigma_{v}(x_i,y_i)R^{V}_{f}(z_i)$, where $\sigma_{v}(x_i,y_i)$ is the vehicles density within the administrative unit and $R^{V}_{f}(z_i)$ is related to the kinetic energy of the impact and the obscuration factor for vehicles.

(b) Property Risk\cite{pangUAVPathOptimization2022a,chen2023MPUAV}:
Property risk refers to the risk of collisions between buildings and other properties caused by high-altitude drone operations. The property risk index $f_{\textup{eco}}$ was established as follows.
\begin{equation}
	\psi(z_i;\mu,\sigma)=\frac{1}{z_i \sigma\sqrt{2\pi}}e^{-\frac{(\ln z_i-\mu)^2}{2\sigma^2}},
\end{equation}
\begin{equation}
	c_{r_p,d}(z_i)=\begin{cases}
		\psi(e^\mu) &{\text{if}}\ 0< z_i \le e^\mu \\
		\psi(z_i) &{\text{otherwise.}}
	\end{cases},
\end{equation}
\begin{equation}
	f_{\textup{eco}} = \sum_{i=0}^{n}c_{r_p,d}(z_i),
\end{equation}
where the function $\psi$ is a distribution function of the flight altitude. $\mu$ and $\sigma$ are the lognormal distribution parameters of the building height. $f_{\textup{eco}}$ is the final property risk objective value.

(c) Noise Pollution \textup{\cite{pangUAVPathOptimization2022a}}:
Noise impact is an important third-party impact for city safety-related government regulation department to consider. The noise pollution indicator can be expressed as the approximate value of spherical propagation, which can be expressed as follows.
\begin{equation}
	f_{\textup{noise}} = \sum_{i=0}^{n} \frac{k\sigma_{p}(x_i,y_i)L_h}{z_i^2+d^2},
\end{equation}
where $k$ is the conversion factor from sound intensity to sound level, $L_h$ is the noise produced by UAV, and $\sigma_{p}(x_i,y_i)$ is the density of people at track point $(x_i,y_i,z_i)$. $d$ is the distance between the UAV and the area of interest, and if the flight altitude exceeds a certain threshold, the noise impact will not be included in the calculation of pollution\cite{torijaEffectsHoveringUnmanned2020}. 
\subsubsection{Constraint}
This paper mainly considers the path constraints brought by the kinetic performance limitation of the UAV flight. Therefore, these constraints must be satisfied for all DMs.


Firstly, the UAV must fly at a limited altitude as shown follow.
\begin{equation}
	H_{min} \leq z_i \leq H_{max}.
\end{equation}

Let the projection of the $i$-th track segment in the plane be $l_i'$, and for a given track segment, the turning angle $\alpha_i$ between each track segment can be calculated as follows. Similarly, the slope angle formed by the $i$-th and the $(i+1)$-th segments can be defined as follows\cite {aitsaadiUAVPathPlanning2022a}.
\begin{equation}
	\label{eqt:alpha}
	\alpha_i = \arccos{\frac{l_i'*l_{i+1}'}{||l_i'||*||l_{i+1}'||}},\beta_i = \arctan{\frac{z_{i+1}-z_i}{||l_i'||}}.
\end{equation}

Due to the power performance of the UAV, the turning angle between the segments cannot be greater than the maximum turning angle, and the slope angle between the segments cannot be greater than the maximum slope angle, which is expressed as hard constraints form shown as follows.

\begin{equation}
	|\alpha_i| \leq \alpha_{max},|\beta_i| \leq \beta_{max},
\end{equation}
where  $\alpha_{max}$ is the maximum turning angle of the UAV, and $\beta_{max}$ is the maximum slope angle of the UAV.

\subsubsection{Problems Construction}
In the main text, the paper introduces the relevant objectives. Thus, the main focus of the discussion in the supplementary document is on constructing the corresponding cases and determining the relevant setting parameters.

When constructing different cases, we consider a set of objectives for different decision makers that consist of mutually conflicting objectives. This is because in practical situations, non-conflicting objectives can be combined into a single objective. 

Regarding the efficiency-related objectives of the decision makers, minimizing path length, energy consumption, and altitude changes usually fall into the same objective category and do not significantly conflict with each other. However, minimizing the total sum of the mission's hover point distance can sometimes lead to increased energy consumption or altitude changes.

Regarding the safety-related objectives of the decision makers, flying UAVs at higher altitudes is beneficial for minimizing noise impact and property risk, but it may not be ideal for minimizing fatal risks. Generally, minimizing noise impact and property risk are objectives that conflict with minimizing fatal risk.

In Ref. \cite{chen2023MPUAV}, six biparty biobjective optimization problems are formed by considering different decision makers choosing different conflicting objectives in Map-A. To further assess the performance of the algorithm, an additional Map-B is considered. Map-A and Map-B use the same distribution parameters, but they have different random seeds for generating random numbers. By incorporating different maps with different combinations of optimization objectives, the six cases in Ref. \cite{chen2023MPUAV} are expanded to twelve cases in this paper. The detailed objective combinations are displayed in Table \ref{table3}. These twelve cases are adopted to thoroughly evaluate the algorithms, and all of them are biparty biobjective optimization problems.

\begin{table}[]
	\centering
	\caption{Case design}
	\scalebox{0.85}{
		\begin{tabular}{llll}
			\toprule[1.5pt]
			Problems &  Efficiency DM Objectives   & Safety DM Objectives & Map Type      \\ 
			\midrule[1.5pt]
			Case 1   &  $\textbf{F}_{\text{eff}}=(f_{\textup{length}},f_{\textup{distance}})$  & $\textbf{F}_{\text{safe}}=(f_{\textup{fatal}},f_{\textup{eco}})$   & MAP-A	\\
			Case 2   &  $\textbf{F}_{\text{eff}}=(f_{\textup{length}}+f_{\textup{height}},f_{\textup{distance}})$  & $\textbf{F}_{\text{safe}}=(f_{\textup{fatal}},f_{\textup{eco}})$  & MAP-A    \\
			Case 3  &   $\textbf{F}_{\text{eff}}=(f_{\textup{fuel}},f_{\textup{distance}})$  & $\textbf{F}_{\text{safe}}=(f_{\textup{fatal}},f_{\textup{eco}})$  & MAP-A   \\
			Case 4  &   $\textbf{F}_{\text{eff}}=(f_{\textup{length}},f_{\textup{distance}})$  & $\textbf{F}_{\text{safe}}=(f_{\textup{fatal}},f_{\textup{noise}})$  & MAP-A    \\
			Case 5  &  $\textbf{F}_{\text{eff}}=(f_{\textup{length}}+f_{\textup{height}},f_{\textup{distance}})$ & 
			$\textbf{F}_{\text{safe}}=(f_{\textup{fatal}},f_{\textup{noise}})$& MAP-A \\
			Case 6  &  $\textbf{F}_{\text{eff}}=(f_{\textup{fuel}},f_{\textup{distance}})$  & $\textbf{F}_{\text{safe}}=(f_{\textup{fatal}},f_{\textup{noise}})$ & MAP-A \\
			
			Case 7   &  $\textbf{F}_{\text{eff}}=(f_{\textup{length}},f_{\textup{distance}})$  & $\textbf{F}_{\text{safe}}=(f_{\textup{fatal}},f_{\textup{eco}})$  & MAP-B 	\\
			Case 8   &  $\textbf{F}_{\text{eff}}=(f_{\textup{length}}+f_{\textup{height}},f_{\textup{distance}})$  & $\textbf{F}_{\text{safe}}=(f_{\textup{fatal}},f_{\textup{eco}})$   & MAP-B   \\
			Case 9  &   $\textbf{F}_{\text{eff}}=(f_{\textup{fuel}},f_{\textup{distance}})$  & $\textbf{F}_{\text{safe}}=(f_{\textup{fatal}},f_{\textup{eco}})$ & MAP-B    \\
			Case 10  &   $\textbf{F}_{\text{eff}}=(f_{\textup{length}},f_{\textup{distance}})$  & $\textbf{F}_{\text{safe}}=(f_{\textup{fatal}},f_{\textup{noise}})$ & MAP-B     \\
			Case 11  &  $\textbf{F}_{\text{eff}}=(f_{\textup{length}}+f_{\textup{height}},f_{\textup{distance}})$ & 
			$\textbf{F}_{\text{safe}}=(f_{\textup{fatal}},f_{\textup{noise}})$& MAB-B \\
			Case 12  &  $\textbf{F}_{\text{eff}}=(f_{\textup{fuel}},f_{\textup{distance}})$  & $\textbf{F}_{\text{safe}}=(f_{\textup{fatal}},f_{\textup{noise}})$& MAP-B  \\
			\bottomrule[1.5pt]
		\end{tabular}
		\label{table3}    
	}
\end{table}

Regarding map generation parameters, this paper adopts a lognormal distribution to describe the distribution of building heights. The parameter $\mu$ of the lognormal distribution is set to 3.04670, and the parameter $\sigma$ is set to 0.76023 \cite{pangUAVPathOptimization2022a}. The population distribution is closely related to the core metropolitan area, where the population is concentrated. Similar to Pang et al.'s work \cite{pangUAVPathOptimization2022a}, a radial basis model is used to simulate population density and generate the experimental data.

Regarding UAV-related parameters, these parameters associated with UAV fuel consumption are set to the same values as in Ref. \cite{ghambariEnhancedNSGAIIMultiobjective2020}, while the parameters related to third-party risks are set to the same values as in Ref. \cite{pangUAVPathOptimization2022a}. The other parameters are as follows: the maximum turning angle $\alpha_{max}$ is $\pi/3$, the maximum slope angle $\beta_{max}$ is $\pi/4$, the atmospheric density $\rho_{A}$ is $1.225 kg/m^3$, the flight speed $v$ is $10 m/s$, the rotating slope area $S_b$ is $0.1 m^2$, the number of rotating paddles $n$ is 4, and the UAV weight $m$ is $1.38 kg$ \cite{pangUAVPathOptimization2022a}. The mission starts at coordinates (1,1) and ends at (45, 45). The predefined UHPs (hover points) are located at (25, 30), (34, 20), and (40, 35).

\begin{figure}[H]  
	\centering
	\includegraphics[scale=0.52,trim= 50 280 105 310,clip]{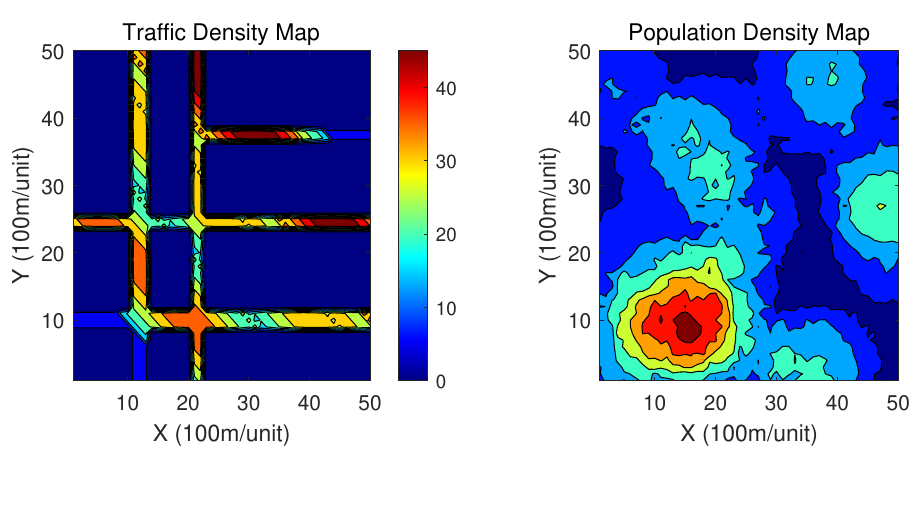}   
	\caption{Population Density and Traffic Density of MAP-A}
	\label{fig:mapa_dens}
\end{figure}

\begin{figure}[H]  
	\centering
	\includegraphics[scale=0.4,trim= 40 240 65 310,clip]{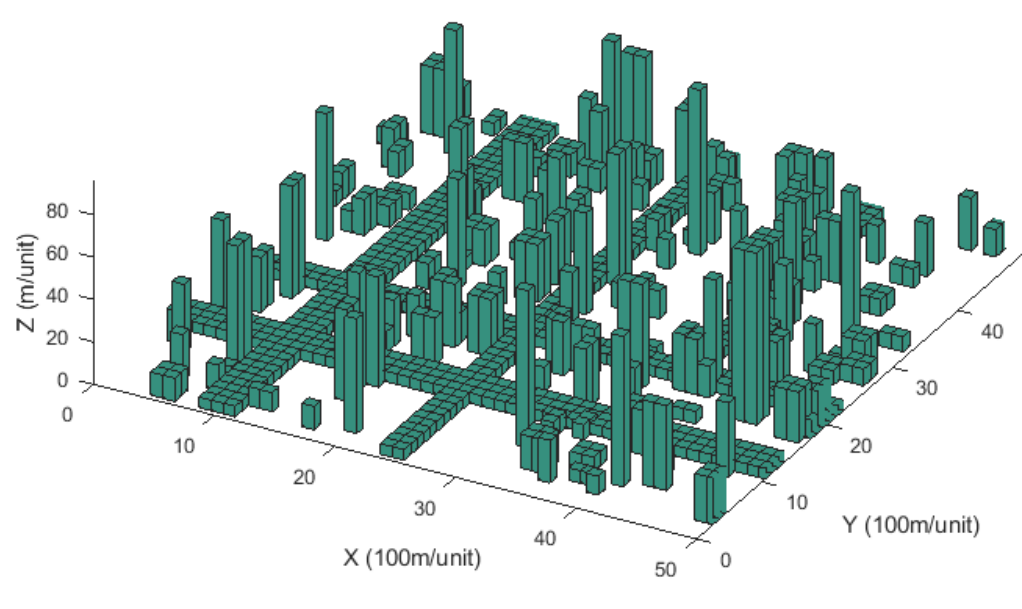}   
	\caption{Buildings of MAP-A}
	\label{fig:mapA_builds}
\end{figure}

\begin{figure}[H]  
	\centering
	\includegraphics[scale=0.50,trim= 50 280 105 310,clip]{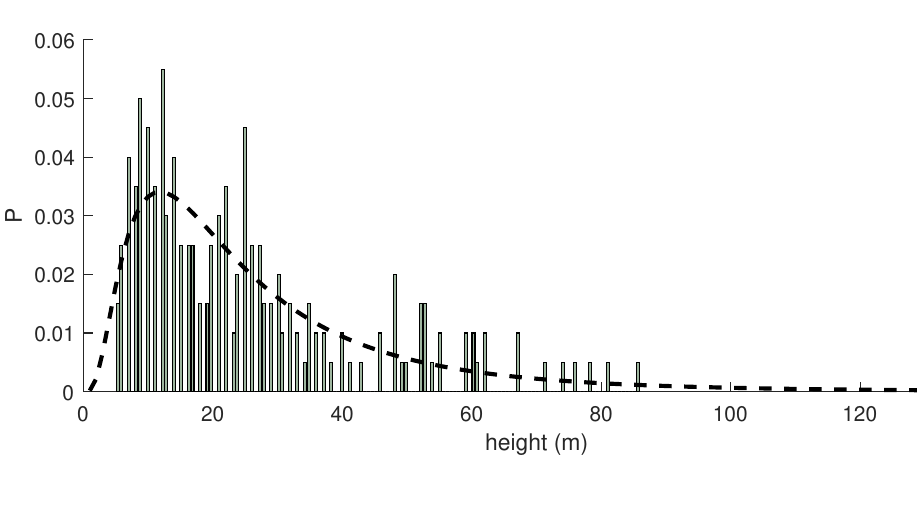}   
	\caption{Building height distribution of MAP-A}
	\label{fig:mapa_height}
\end{figure}

\begin{figure}[H]  
	\centering
	\includegraphics[scale=0.52,trim= 90 280 105 310,clip]{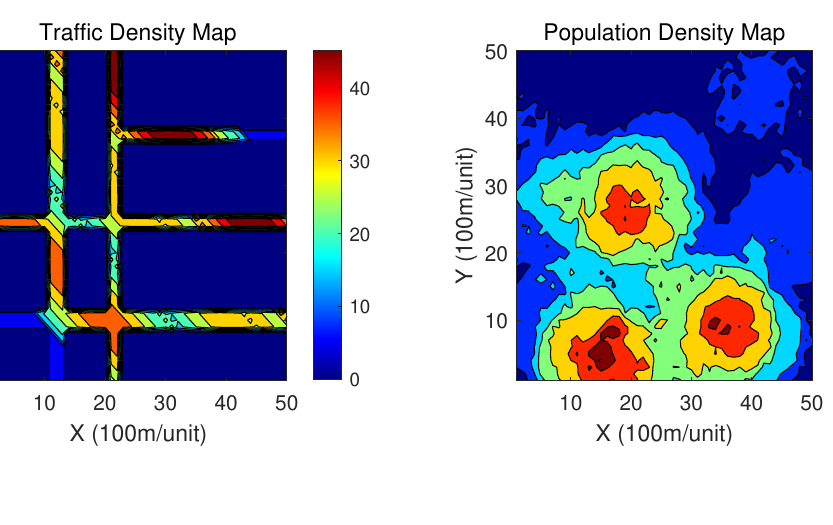}   
	\caption{Population Density and Traffic Density of MAP-B}
	\label{fig:mapB_dens}
\end{figure}

\begin{figure}[H]  
	\centering
	\includegraphics[scale=0.4,trim= 40 240 65 310,clip]{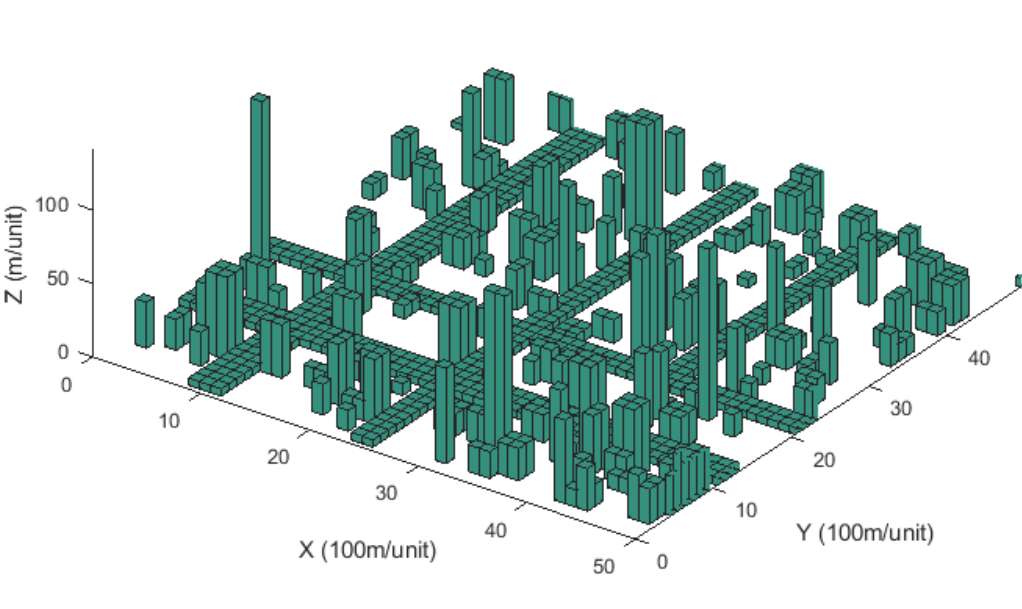}   
	\caption{Buildings of MAP-A}
	\label{fig:mapB_builds}
\end{figure}

\begin{figure}[H]  
	\centering
	\includegraphics[scale=0.5,trim= 50 280 105 310,clip]{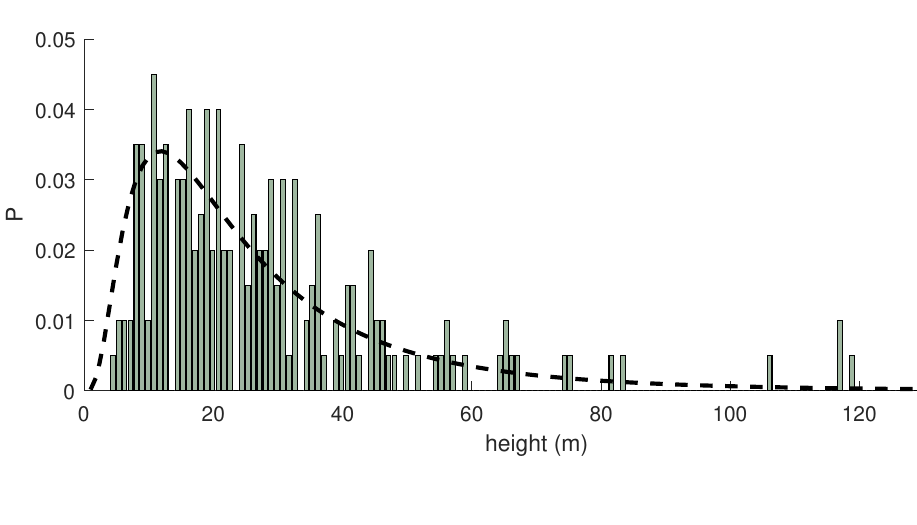}   
	\caption{Building height distribution of MAP-B}
	\label{fig:mapB_height}
\end{figure}

Fig.\ref{fig:mapa_dens} to Fig.\ref{fig:mapB_height} visualize the attributes related to the objective function calculations for MAP-A and MAP-B, including population density, vehicle density, and building height. Fig.\ref{fig:mapa_dens} and Fig.\ref{fig:mapB_dens} illustrate the continuous population density distribution and vehicle density distribution for Maps A and B, respectively, where the minimum unit on the X and Y axes is 100 meters, with each grid representing a 100m by 100m geometric plane. When calculating population density or traffic density for a grid, the population density value or vehicle density value at the geometric center of the grid is used as an approximation. Fig.\ref{fig:mapA_builds} and Fig.\ref{fig:mapB_builds} visually present the maximum building heights and the no-fly zone heights for highways, where the minimum unit on the X and Y axes is 100 meters, and each grid represents a 100m by 100m geometric plane. For simplification in calculations, all buildings are considered rectangular prisms with heights equal to their maximum heights. Fig.\ref{fig:mapa_height} and Fig.\ref{fig:mapB_height} display the frequency distribution and the corresponding distribution of building heights.

\subsection{Analysis and Visualization of the Impact of Activation Size}
\label{as}

To further illustrate the importance of an adaptive activation strategy, we present a visualization of the population distribution when solving the Case 1 problem and examine the distribution of activated individuals at different stages of the process for various activation sizes.

\begin{figure}[H]
	\centering  
	\includegraphics[scale=0.46,trim=20 0 0 0,clip]{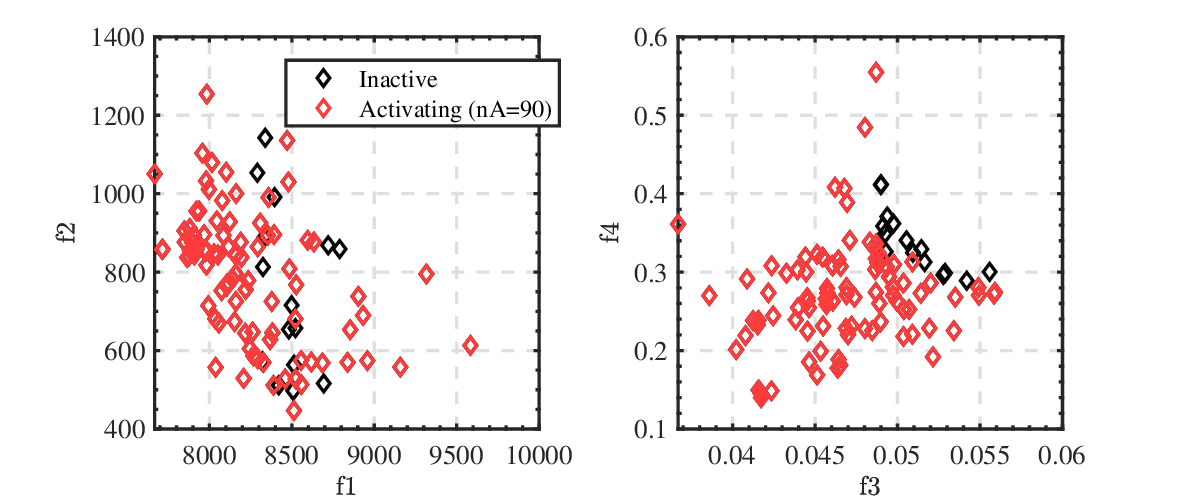}    
	\caption{The population distribution (pre-stage) in solving Case 1 using existing multiparty immune algorithms is shown when the number of activations is set to 90. The Case 1 problem is given in Table \ref{table3}.}
	\label{Asize90_P1}
\end{figure}

The experimental results indicate that the algorithm requires selecting an appropriate number of activated individuals. Figs. \ref{Asize20_P1} to \ref{Asize90_P1} depict the scenarios for different numbers of activations in the pre-stage. Specifically, Fig. \ref{Asize20_P1} corresponds to 20 activated individuals, Fig. \ref{Asize50_P1} corresponds to 50 activated individuals, and Fig. \ref{Asize90_P1} corresponds to 90 activated individuals. It is evident that the case shown in Fig. \ref{Asize20_P1} is insufficient to ensure diversity among the activated individuals on the Pareto fronts associated with different decision makers. On the other hand, Figs. \ref{Asize50_P1} and \ref{Asize90_P1} demonstrate satisfactory diversity. However, an excessive number of activated individuals as seen in Fig. \ref{Asize90_P1} can hinder the identification of individuals closer to the common PF, thereby negatively impacting the convergence of the population.

This highlights the importance of selecting the appropriate number of activated individuals based on the specific circumstances when employing multiparty immune algorithms.

\begin{figure}[h]
	\centering  
	\includegraphics[scale=0.46,trim=20 0 0 0,clip]{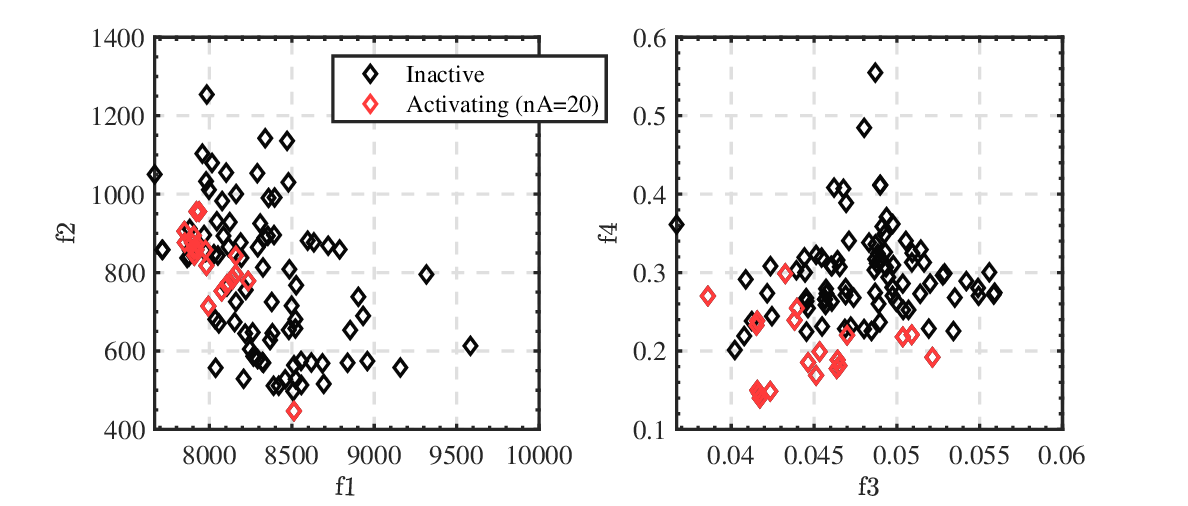}   
	\caption{The population distribution (pre-stage) in solving Case 1 using existing multiparty immune algorithms (such as BPNNIA, BPHEIA and BPAIMA) is shown when the number of activations is set to 20. The Case 1 problem is given in Table \ref{table3}.}
	\label{Asize20_P1}
\end{figure}

\begin{figure}[h]
	\centering  
	\includegraphics[scale=0.46,trim=20 0 0 0,clip]{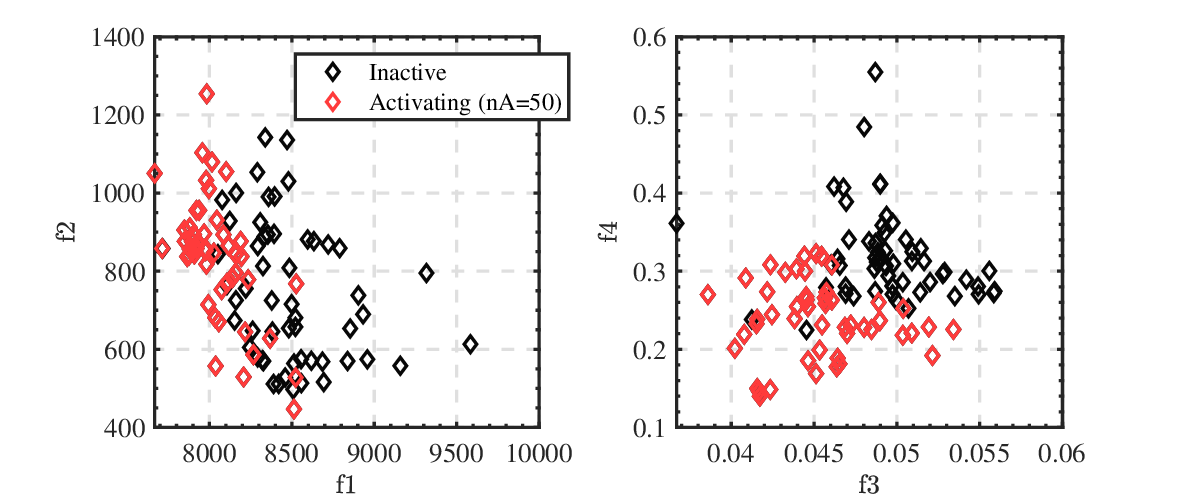}    
	\caption{The population distribution (pre-stage) in solving Case 1 using existing multiparty immune algorithms is shown when the number of activations is set to 50. The Case 1 problem is given in Table \ref{table3}.}
	\label{Asize50_P1}
\end{figure}

\begin{figure}[H]
	\centering  
	\includegraphics[scale=0.46,trim=20 0 0 0,clip]{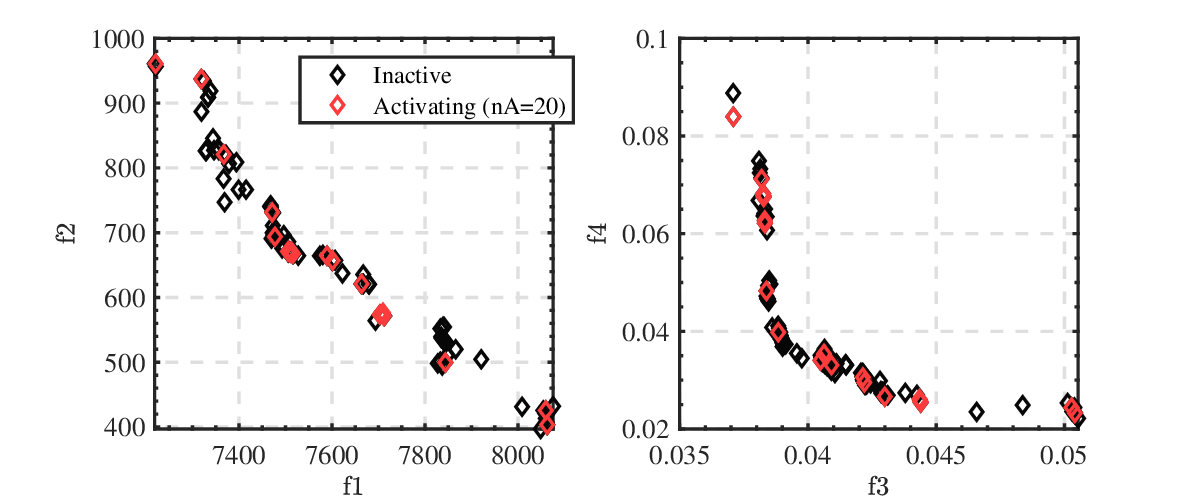}    
	\caption{The population distribution (mid-stage) in solving Case 1 using existing multiparty immune algorithms is shown when the number of activations is set to 20. The Case 1 problem is given in Table \ref{table3}.}
	\label{Asize20_P2}
\end{figure}

Additionally, other experimental results indicate that the selected activated individuals need to be adjusted according to different stages of the algorithm. For comparison, let's consider Fig. \ref{Asize20_P1} and Fig. \ref{Asize20_P2}, which illustrate the distribution of activated individuals when using 20 as the number of activated individuals in different stages of the process. In the case of Fig. \ref{Asize20_P2}, using only 20 activated individuals is enough to ensure the diversity of activating antibodies on the PF associated with different decision makers in the mid-stage of the process. However, as shown in Fig. \ref{Asize20_P1}, selecting only 20 activated individuals in the pre-stage of the process does not guarantee diversity across different decision maker perspectives.

\subsection{Visualization of MPIA Solutions for BPUAV-PP Problems}
\label{sec:mpuavpp}
\begin{figure*}[htbp]  
	\centering
	\includegraphics[scale=0.56,trim= 70 0 0 0,clip]{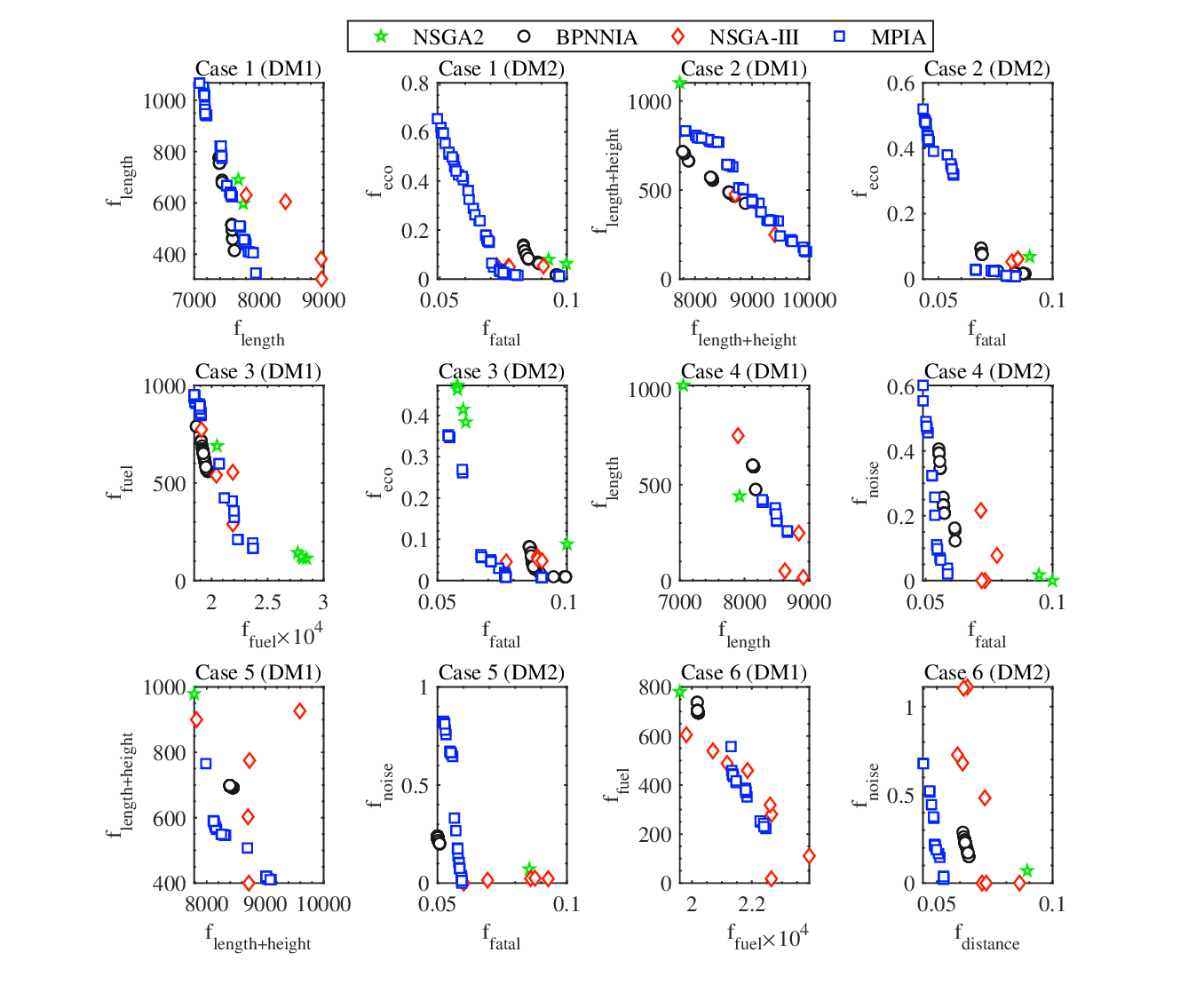}   
	\caption{Visualization of the results of solving Cases 1-6}
	\label{fig:solution1}
\end{figure*}

\begin{figure*}[htbp]  
	\centering
	\includegraphics[scale=0.56,trim= 70 0 0 0,clip]{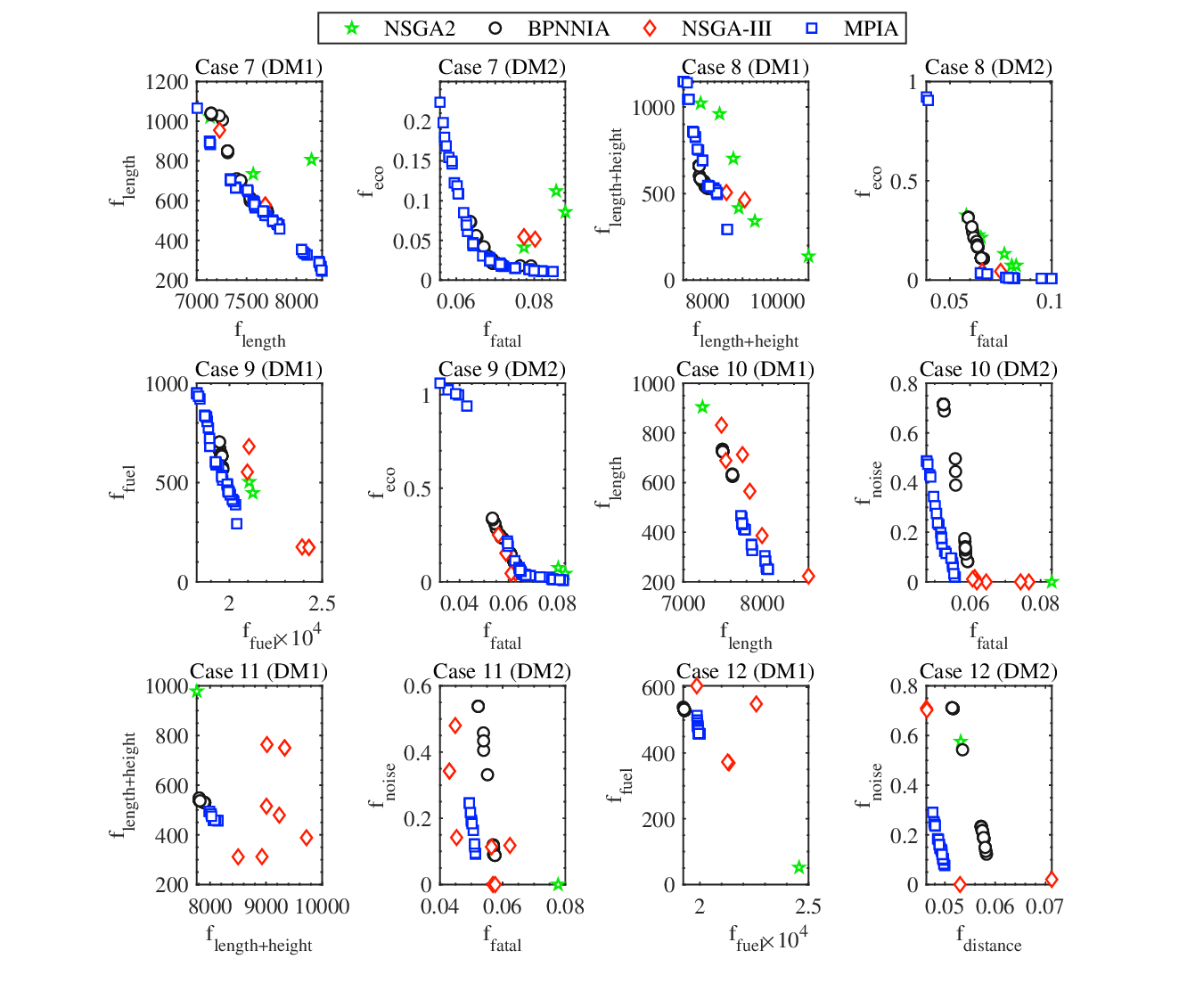}   
	\caption{Visualization of the results of solving Cases 7-12}
	\label{fig:solution2}
\end{figure*}

Fig. \ref{fig:solution1} shows the results of solving Cases 1-6 with the ordinary multiobjective optimization algorithm NSGA-II, NSGA-III; existing multiparty multiobjective immune algorithm BPNNIA and our proposed algorithm called MPIA. Fig. \ref{fig:solution2} shows the results of solving Cases 7-12 with these algorithms. For each algorithm, the optimal solutions are eventually obtained from among the final population based on the idea that the acceptable solutions are not Pareto-dominated from the perspective of at least one DM.

In the visualization results, we can observe that the final solution set of MPMOEAs contains significantly more individuals on the common Pareto front (PF) compared to MOEAs, indicating that MPMOEAs are more suitable for solving these problems. Additionally, MPIA effectively solves BPUAV-PP problems, producing a sufficiently diverse and convergent set of solutions on the common Pareto front.
\subsection{Ablation Experiments for Proposed Methods}
\label{ablation}
In the main text, we conducted comparative experiments between MPIA and typical multiobjective optimization algorithms against state-of-the-art multiparty multiobjective optimization algorithms. We designed several ablation experiments addressing the following questions to better validate the effectiveness of the two proposed strategies (crossover-guided operator and adaptive activation operator).
\begin{enumerate}
	\item [(1)] How does the algorithm perform without these strategies, using only advanced DE operators? We conducted an ablation experiment with an algorithm called MPIA-BASE to investigate this. To ensure fairness in the ablation experiments, MPIA-BASE maintains the same framework as MPIA, differing only in its activation strategy. It employs a fixed-size activation strategy used by existing multiparty multiobjective immune algorithms like BPNNIA, BPHEIA, and BPAIMA. In the part of the operator selection, the operator selection rules remain identical to that of MPIA. The only difference is that the component corresponding to the MPIA inter-party guided crossover replaces the conventional rand/1/bin DE operator-I ($CR=0.5, F=0.5$).
	\item [(2)] Does using the adaptive activation operator alone lead to performance improvement? To verify this, we designed an algorithm that combines MPIA-BASE with the adaptive activation operator, referred to as MPIA-A, for an ablation experiment. 
	\item [(3)] Does the use of the crossover-guided operator alone lead to performance improvement? To verify this, we created an algorithm that combines MPIA-BASE with the inter-party guided crossover, called MPIA-C, for another ablation experiment.
\end{enumerate}

In these experiments, we maintained consistency by using the same comparison algorithms from the main text, such as NSGA-II, NSGA-III, MPNDS2, and BPAIMA. Furthermore, to verify the combined effectiveness of both strategies across different problems, we retained the MPIA algorithm for comparative analysis. The parameter settings for the algorithms and experiments were kept consistent with those in the main text. Table \ref{tab:exp22} presents the performance of these algorithms on MPMOPs and Table \ref{tab:exp11} shows their performance on the BPUAV-PP problems. In the table, "(+)" indicates that the method with the corresponding improved strategy outperforms the baseline algorithm MPIA-BASE, while "(-)" indicates that it underperforms. The last row provides the number of instances where MPIA-A, MPIA-B, and MPIA achieved better or worse results compared to the MPIA-BASE algorithm.

From the results of the aforementioned experiments, by comparing the performance metrics of different algorithms solving the same problem, the following conclusions can be drawn.

\begin{table*}[h]
	\centering
	\caption{$sumHV$ Metric of Different Algorithms in BPUAV-PP Problem Case 1 - Case 12}
	\begin{tabular}{c|cccccc|ccc}
		\toprule[1.5pt]
		\multicolumn{1}{c}{\multirow{2}[3]{*}{Problems}} & \multicolumn{9}{c}{Algorithms} \\
		\cmidrule{2-10}    \multicolumn{1}{c}{} & NSGA-II & NSGA-III & MPNDS & MPNDS2 & BPAIMA & MPIA-BASE & MPIA-A & MPIA-B  & MPIA \\
		\midrule
		\midrule
		Case 1 & 2.98E-01 & 3.87E-01 & 4.65E-01 & 4.78E-01 & 5.24E-01 & 5.41E-01 & 5.86E-01(+) & 5.71E-01(+) & 5.91E-01(+) \\
		Case 2 & 3.03E-01 & 4.39E-01 & 4.66E-01 & 4.73E-01 & 5.26E-01 & 5.44E-01 & 5.85E-01(+) & 5.89E-01(+) & 5.85E-01(+) \\
		Case 3 & 3.25E-01 & 4.24E-01 & 4.74E-01 & 4.82E-01 & 5.32E-01 & 5.68E-01 & 5.92E-01(+) & 5.60E-01(+) & 5.90E-01(+) \\
		Case 4 & 2.41E-01 & 4.19E-01 & 4.30E-01 & 4.64E-01 & 5.04E-01 & 5.14E-01 & 5.74E-01(+) & 5.24E-01(+) & 5.57E-01(+) \\
		Case 5 & 2.33E-01 & 4.60E-01 & 4.27E-01 & 4.68E-01 & 5.16E-01 & 4.80E-01 & 5.35E-01(+) & 5.14E-01(+) & 5.36E-01(+) \\
		Case 6 & 2.42E-01 & 4.62E-01 & 4.32E-01 & 4.50E-01 & 4.97E-01 & 5.09E-01 & 5.62E-01(+) & 5.01E-01(-) & 5.60E-01(+) \\
		Case 7 & 3.18E-01 & 3.97E-01 & 4.78E-01 & 4.87E-01 & 5.38E-01 & 5.35E-01 & 5.85E-01(+) & 5.66E-01(+) & 5.82E-01(+) \\
		Case 8 & 3.46E-01 & 4.37E-01 & 4.85E-01 & 5.12E-01 & 5.56E-01 & 5.41E-01 & 5.81E-01(+) & 5.78E-01(+) & 6.01E-01(+) \\
		Case 9 & 3.10E-01 & 3.99E-01 & 4.82E-01 & 4.90E-01 & 5.22E-01 & 5.31E-01 & 5.87E-01(+) & 5.76E-01(+) & 5.97E-01(+) \\
		Case 10 & 2.32E-01 & 3.87E-01 & 4.28E-01 & 4.53E-01 & 4.58E-01 & 4.40E-01 & 5.04E-01(+) & 4.75E-01(+) & 5.13E-01(+) \\
		Case 11 & 2.52E-01 & 4.37E-01 & 4.51E-01 & 4.77E-01 & 4.69E-01 & 4.47E-01 & 5.50E-01(+) & 5.08E-01(+) & 5.36E-01(+) \\
		Case 12 & 2.34E-01 & 4.12E-01 & 4.59E-01 & 4.81E-01 & 4.79E-01 & 4.53E-01 & 5.14E-01(+) & 4.92E-01(+) & 5.34E-01(+) \\
		\midrule
		\textbf{($+/-$)} &  &  &  &  &  & baseline  & 12/0 & 11/1  & 12/0  \\
		\bottomrule[1.5pt]
	\end{tabular}%
	\label{tab:exp11}%
\end{table*}%

\begin{table*}[h]
	\setlength{\tabcolsep}{0.12cm}
	\centering
	\caption{MPIGD Metric of Different Algorithms in MPMOP 1 - MPMOP 12}
	\begin{tabular}{l|cccccccc|ccc}
		\toprule[1.5pt]
		\multicolumn{1}{c}{\multirow{2}[4]{*}{Problems}} & \multicolumn{11}{c}{Algorithms} \\
		\cmidrule{2-12}    \multicolumn{1}{c}{} & NSGA-II & NSGA-III & MPNDS & MPNDS2 & BPHEIA & BPNNIA & BPAIMA & MPIA-BASE & MPIA-A & MPIA-B & MPIA \\
		\midrule
		\midrule
		MPMOP1 & 9.26E-03 & Inf   & 1.59E-05 & 1.43E-05 & 4.15E-05 & 3.96E-05 & 6.19E-05 & 5.43E-05 & 3.07E-05(+) & 5.49E-05(-) & 2.89E-05(+) \\
		MPMOP2 & 3.90E-02 & 3.75E-01 & 1.28E-02 & 1.52E-02 & 3.59E-02 & 3.28E-02 & 3.08E-05 & 2.65E-05 & 2.32E-05(+) & 2.33E-05(+) & 2.37E-05(+) \\
		MPMOP3 & 2.25E-01 & 2.13E-01 & 2.27E-01 & 2.10E-01 & 1.82E-01 & 2.06E-01 & 7.86E-02 & 7.26E-02 & 7.28E-02(-) & 7.01E-02(+) & 6.84E-02(+) \\
		MPMOP4 & 1.40E+00 & 1.17E+00 & 4.88E-02 & 5.18E-02 & 5.85E-02 & 5.96E-02 & 3.03E-02 & 3.25E-02 & 3.18E-02(+) & 3.19E-02(+) & 3.25E-02(+) \\
		MPMOP5 & 3.56E-01 & Inf   & 4.00E-02 & 4.03E-02 & 4.31E-02 & 4.34E-02 & 3.58E-02 & 3.93E-02 & 3.73E-02(+) & 3.74E-02(+) & 3.49E-02(+) \\
		MPMOP6 & 1.99E+00 & 2.30E-01 & 1.50E-02 & 1.49E-02 & 1.46E-02 & 1.41E-02 & 1.81E-02 & 1.81E-02 & 1.89E-02(-) & 1.86E-02(-) & 1.85E-02(-) \\
		MPMOP7 & Inf   & Inf   & 5.42E-06 & 6.38E-06 & 6.52E-05 & 6.49E-05 & 2.80E-05 & 2.76E-05 & 1.74E-05(+) & 2.70E-05(+) & 1.88E-05(+) \\
		MPMOP8 & 1.56E-01 & 6.69E-01 & 1.54E-01 & 3.60E-03 & 7.54E-02 & 8.02E-02 & 1.71E-05 & 1.00E-05 & 1.05E-05(-) & 8.81E-06(+) & 7.83E-06(+) \\
		MPMOP9 & 2.00E+00 & 1.91E+00 & 7.04E-02 & 7.28E-02 & 8.16E-02 & 7.87E-02 & 4.41E-02 & 4.50E-02 & 4.45E-02(+) & 4.55E-02(-) & 4.40E-02(+) \\
		MPMOP10 & 6.68E-01 & Inf   & 3.88E+00 & 3.25E-02 & 3.54E-02 & 3.57E-02 & 3.58E-02 & 8.14E-01 & 3.65E-02(+) & 2.98E-01(+) & 3.62E-02(+) \\
		MPMOP11 & 3.58E+00 & 8.72E-01 & 1.70E-02 & 1.67E-02 & 1.77E-02 & 1.75E-02 & 1.92E-02 & 1.88E-02 & 1.94E-02(-) & 1.92E-02(-) & 1.91E-02(-) \\
		\midrule
		\ \ \textbf{($+/-$)} &  &  &  &  &  &  &  &baseline  & 7/4 & 7/4  & 9/2  \\
		\bottomrule[1.5pt]
	\end{tabular}%
	\label{tab:exp22}%
\end{table*}%
\begin{enumerate}
	\item [(1)] For the first issue, by observing the experimental results of MPIA-BASE and the similar algorithm BPAIMA, and comparing them with the more naive immune algorithm BPNNIA and BPAIMA, it can be seen that the advanced DE operator combination does indeed improve the performance of MPMOEAs in solving MPMOPs. However, this improvement is limited. Moreover, the performance of MPIA-BASE is similar to that of BPAIMA, which is consistent with the similar DE operators and strategies used by both algorithms. MPIA-BASE, which does not incorporate the two strategies proposed in this paper, does not show essential performance improvement compared to the BPAIMA algorithm.
	
	\item [(2)] For the second issue, by observing the experimental results of MPIA-BASE and MPIA-B, it can be seen that MPIA-B, which introduces a inter-party guided crossover, performs better than MPIA-BASE in solving most problems. This demonstrates that the strategy can effectively enhance the performance of MPMOEAs in solving MPMOPs.
	
	\item [(3)] For the third issue, by observing the experimental results of MPIA-BASE and MPIA-A, it can be seen that MPIA-A, which introduces an adaptive activation strategy, performs better than MPIA-BASE in solving most problems. This demonstrates that the strategy can effectively enhance the performance of MPMOEAs in solving MPMOPs.
\end{enumerate}
	\end{document}